\documentclass{article}

\usepackage{PRIMEarxiv}

\usepackage[utf8]{inputenc}  
\usepackage[T1]{fontenc}     
\usepackage{hyperref}        
\usepackage{url}             
\usepackage{booktabs}        
\usepackage{amsfonts}        
\usepackage{amsmath}
\usepackage{amssymb}         
\usepackage{nicefrac}        
\usepackage{microtype}       
\usepackage{graphicx}
\usepackage{textcomp}        

\usepackage{longtable,array,calc,etoolbox}
\patchcmd\longtable{\par}{\if@noskipsec\mbox{}\fi\par}{}{}
\IfFileExists{footnotehyper.sty}{\usepackage{footnotehyper}}{\usepackage{footnote}}
\makesavenoteenv{longtable}
\providecommand{\tightlist}{\setlength{\itemsep}{0pt}\setlength{\parskip}{0pt}}

\usepackage[normalem]{ulem}            
\providecommand{\st}[1]{\sout{#1}}     

\graphicspath{{./}}

\title{Free Energy Heuristics:\\ Fast-and-Frugal Cognition as Active Inference Under Uncertain Precision}

\author{
    Alex Bogdan \\
    Evolutionairy AI \\
    Toronto, Canada \\
  }

\begin{document}
\maketitle

\begin{abstract}
Chain-of-thought (CoT) prompting reliably improves large language model
performance on mathematical and symbolic reasoning, but recent work
shows that extending CoT degrades performance on planning, contested
ethical reasoning, and other tasks where the model lacks a reliable
internal verifier. Literature documents both effects without a
principled account of which property governs which side a given item
falls on.

We argue that \emph{meta-uncertainty about the precision of one's own
evidence} is a governing property. We prove that the variational policy
minimizing expected free energy under uncertain precision truncates cue
integration after a finite number \(k^{*}\) of high-validity cues
whenever the precision prior has heavy enough tails (Theorem 2.6.1), and
that under a Descending Dominance condition the truncated policy is
sample-wise identical to Gigerenzer's take-the-best heuristic (Theorem
2.7.4). Fast-and-frugal heuristics and active inference are, in the
meta-uncertain regime, two descriptions of the same computation.

This equivalence makes a sharp, falsifiable prediction about LLM
behavior: on items in the high-meta-uncertainty regime, extending CoT
should degrade rather than improve answer quality. We operationalize the
regime with a per-item score (cross-prompt variance, cross-seed
variance, and calibration error) validated by simulate-and-recover
(\(\rho > 0.96\)), introduce \textbf{FEH-79}, a benchmark of Knightian
frames across four operational categories (non-recurrent forecasting,
novel-synthetic, open-ended dilemmas, strategic uncertainty) with
matched confound controls, and run a pre-registered confirmatory study
over seven models (five open-weight, 3B--32B, and two frontier systems),
five CoT-length conditions, and 7,875 responses. The registered primary
contrast is the randomly assigned reasoning length, and the decision
gate requires that the interaction \(\beta_{3} < 0\) at a posterior
probability above 0.95, together with a robust implied accuracy drop of
more than 6 percentage points.

The prediction held. The interaction is negative at the posterior
probability \(1.00\), with a posterior-median high-regime accuracy drop
of \(17.3\) percentage points (95\% credible interval
\(\lbrack 7.7,25.5\rbrack\)), clearing the pre-registered gate; matched
control items with definite answers show no such cost. The effect is not
universal, and we do not present it as such: it is decisive in the
capable mid-to-large models, directional in the two frontier systems,
and absent-to-reversed in the two weakest, exactly the regime-dependent
texture the theory predicts.

The framework gives a principled answer to ``when should we use
chain-of-thought?'' and unifies the Bayesian rational-analysis and
fast-and-frugal traditions: less-is-more effects are evidence about the
meta-uncertainty regime, not against Bayesian cognition.
\end{abstract}

\keywords{active inference \and fast-and-frugal heuristics \and Knightian uncertainty \and chain-of-thought \and large language models \and pre-registration}

\section{Introduction}\label{chapter-1-introduction}

\subsection{A puzzle}\label{a-puzzle}

Consider two questions you might pose to a modern large language model.

The first: \emph{What is 47 \ensuremath{\times} 83?} If the model answers directly, it may
err. If you ask it to ``think step by step,'' it reliably gets 3,901. If
you let it reason at length, considering multiple angles, it still
arrives at 3,901. The reasoning chain extends, and the answer either
improves or remains the same.

The second: \emph{A hospital ICU has one ventilator and two patients
with identical clinical profiles. Patient A is a 35-year-old single
parent; patient B is a 65-year-old senior physician. What is the most
ethically defensible decision procedure?} If the model answers directly,
it states some position; perhaps ``random allocation.'' If you ask it to
reason in three brief steps, it surfaces the
utilitarian-vs-deontological framing and commits to a position with
explicit justification. If you let it reason for fifteen steps,
considering multiple angles, something different happens: the chain
frequently extends past the point where additional reasoning carries
information, hedges shift between framings, and the resulting position
is sometimes less coherent than the three-step version.

The first item is a question with a settled answer. The second is not.
Both are well-formed prompts that contemporary models will produce
something from. But the function from ``amount of reasoning'' to
``quality of answer'' is qualitatively different. On the first, more
reasoning is monotonically helpful or neutral. On the second, there is a
window in which a short chain of thought adds structure and an extended
region in which additional reasoning starts to subtract from coherence
rather than add to it.

This paper is about why that asymmetry exists, when it should appear,
and what it predicts about an entire class of questions on which
language models are increasingly being deployed.

\subsection{The chain-of-thought
contradiction}\label{the-chain-of-thought-contradiction}

Chain-of-thought prompting has become one of the most reliable tools in
the applied LLM toolkit. The original demonstrations \cite{wei2022, kojima2022} showed that asking a model to articulate its
reasoning before committing to a final answer substantially improves
performance on multi-step arithmetic, commonsense reasoning, and
symbolic manipulation. The result generalized across model families and
quickly became standard practice.

A counter-current followed. Sprague et al.~\cite{sprague2024} found that the gains
from CoT are concentrated in mathematical and symbolic tasks; on most
other benchmarks, the lift is small or absent. Stechly et al.~\cite{stechly2024}
showed that on planning tasks where the model lacks a verifier,
additional reasoning steps degrade rather than improve performance. Work
on CoT faithfulness has shown that the reasoning chain a model produces
is often not the chain from which its answer was actually derived
\cite{turpin2023, lanham2023}. And in adversarial
conditions (questions where prior beliefs and reasoning conflict),
extended reasoning has been shown to amplify rather than correct biases.

Two empirical observations, both well-replicated, sit side by side: more
reasoning helps on some tasks and hurts on others. The literature has
catalogued examples on both sides but lacks a principled account of
\emph{what task property} governs which side a given item falls on.
Without such an account, every new task category requires a fresh
empirical check and prompting practice defaults to ``use CoT unless
someone proves it hurts''; a reasonable heuristic in the absence of
theory but not a satisfying scientific position.

The conjecture this paper advances is that there is a single underlying
property, meta-uncertainty about the precision of one's own evidence,
that governs whether additional reasoning helps. When meta-uncertainty
is low (the cue values you would compute are reliable), more reasoning
is helpful. When meta-uncertainty is high (you cannot tell which of your
candidate inferences are well-grounded), additional reasoning becomes a
source of noise rather than a source of signal. The optimal policy is to
commit early to whatever your first one or two reasoning steps deliver.
The asymmetry between the multiplication problem and the triage problem
is the asymmetry between low-meta-uncertainty and high-meta-uncertainty
regimes.

\subsection{An old framework, a new test
bed}\label{an-old-framework-a-new-test-bed}

The claim that more deliberation is sometimes worse than less is not
new. Gerd Gigerenzer and his collaborators \cite{gigerenzer1996, gigerenzer2009} have spent three decades documenting
\emph{less-is-more effects} in human judgment: simple heuristics such as
take-the-best, tallying, recognition, that ignore most of the available
information and that, on a wide range of inference problems, match or
beat strategies that use all the cues. The fast-and-frugal program is
the most thorough catalog of less-is-more effects in cognitive science.

The fast-and-frugal program has had an awkward relationship with the
rational-analysis tradition \cite{anderson1991, oaksford2007, lieder2020}, which treats cognition as approximate
Bayesian inference. Bayesian models, taken literally, predict that
incorporating more evidence cannot hurt: more cues should never reduce
posterior accuracy. Empirical less-is-more effects are then read either
as evidence against Bayesian cognition or as artifacts of resource
constraints. The two literatures have largely argued past each other.

This paper argues that the disagreement dissolves once meta-uncertainty
is incorporated into the model. A Bayesian agent whose own precision is
uncertain, who does not know how reliable her cue weights are, has a
\emph{rational} incentive to truncate cue integration after a small
number of high-validity cues. The cue-truncation theorem (Theorem 2.6.1)
shows that the optimal stopping point \(k^{*}\) is finite whenever the
precision prior has heavy enough tails, and that under stronger
conditions (the Descending Dominance condition of Appendix A.4) the
truncated policy is sample-wise identical to Gigerenzer's take-the-best.
Heuristics and active inference are, in this regime, two descriptions of
the same computation.

The framework is general; it speaks to any agent operating under
meta-uncertain precision. But large language models are an unusually
clean empirical test bed. Their reasoning is externalized as text; the
``number of cues integrated'' is the number of reasoning steps in the
chain of thought, which can be directly manipulated by specifying prompt
length. The latent regime, whether the agent is in a high- or
low-meta-uncertainty state, is approximated by behavioural signatures
(cross-prompt variance, cross-seed variance, calibration error) that are
also directly measurable from the model. The theoretical equivalence
makes a sharp prediction about LLM behavior: on items where the regime
score is high, extending the chain of thought should \emph{degrade}
answer quality, even though on items where the regime score is low,
extending it should improve or preserve quality. That prediction is the
empirical core of this paper.

\subsection{The prediction}\label{the-prediction}

The prediction takes the form of an interaction effect in a hierarchical
model. For each (model, item) cell, we measure accuracy under five
chain-of-thought length conditions: no CoT, \textasciitilde3 steps,
\textasciitilde7 steps, \textasciitilde15 steps, and unconstrained. The
high-meta-uncertainty regime is operationalized by item category, the
Knightian frames of FEH-79 against matched controls, an assignment
backed by the falsifiable construction criteria of \S{}4 (\S{}3 develops the
continuous regime score the categories instantiate). The registered
primary contrast is the randomly assigned reasoning length, not the
realized step count, which is endogenous (\S{}6). We then fit:

\[\eta_{ij} = \beta_{0} + \beta_{1} \cdot \text{steps}_{ij} + \beta_{2} \cdot \text{regime}_{i} + \beta_{3} \cdot (\text{steps}_{ij} \times \text{regime}_{i}) + \alpha_{m(ij)} + \gamma_{m(ij)} \cdot \text{steps}_{ij} + u_{i}\]

The registered confirmatory gate requires the interaction
\(\beta_{3} < 0\) at posterior probability above 0.95 together with a
robust implied high-regime accuracy drop above six percentage points;
the opposite direction, \(\beta_{3} > 0\) at posterior probability above
0.95, falsifies the theory. The original full-reversal statistic
\(|\beta_{3}| > |\beta_{1}|\) is retained as a reported effect size
rather than a gate. The hypothesis, the gate, and the deviation protocol
were pre-registered and amended before data collection (Section 6), and
the outcome is reported in Section 7.

The benchmark on which the prediction is tested is FEH-79: a 79-frame
item pool spanning four operational categories of Knightian uncertainty
(non-recurrent forecasting, novel-synthetic scenarios, open-ended
dilemmas, and strategic uncertainty) together with three
confound-control sets (50 reference, aleatory, and calibration-probe
items). The benchmark construction protocol (Section 4) follows the
operational criteria in \S{}4.3 to ensure that meta-uncertainty, rather
than computational difficulty or training-data contamination, is the
property that distinguishes Knightian from control items.

\subsection{Contributions}\label{contributions}

This paper makes five contributions:

\begin{enumerate}
\def\labelenumi{\arabic{enumi}.}
\item
  \textbf{Equivalence theorem (Section 2).} Under uncertain precision,
  the variational policy that minimizes expected free energy is
  equivalent to the cue-integration policies of the fast-and-frugal
  heuristics program for a non-empty regime of meta-uncertainty. Theorem
  2.6.1 (cue-truncation) gives the existence result; Theorem 2.7.4
  (Descending Dominance) gives the sample-wise identity with
  take-the-best.
\item
  \textbf{Operationalization (Section 3).} A measurement procedure that
  maps the theoretical ``high-meta-uncertainty regime'' to behavioral
  signatures observable from an LLM, with a simulate-and-recover
  validation showing that the regime score recovers the underlying
  precision prior parameter (\(\rho > 0.96\) for the
  prompt-and-seed-variance composite).
\item
  \textbf{Benchmark (Section 4).} FEH-79: 79 Knightian frames spanning
  four operational categories, plus 50 controls; an item-construction
  protocol that distinguishes meta-uncertainty from difficulty; and a
  step-counting pipeline that converts arbitrary CoT responses into a
  comparable cue count.
\item
  \textbf{Pre-registered empirical study (confirmed).} A confirmatory
  test of the regime-dependent prediction across a seven-model panel
  (five open-weight models, 3B--32B, and two frontier systems), five
  chain-of-thought length conditions, and five replications per cell.
  Pre-registration locked in the hypothesis thresholds, the
  falsification criteria, and the robustness battery before data
  collection; the registered interaction was negative at a posterior
  probability of 1.00, with a 17.3-point drop in high-regime accuracy
  (Section 7).
\item
  \textbf{Theoretical resolution.} A unified account of when
  less-is-more effects should appear in inference under uncertainty,
  applicable to human and machine reasoners. The framework treats
  rational-analysis and fast-and-frugal traditions as describing the
  same computation in different meta-uncertainty regimes, rather than as
  rival accounts.
\end{enumerate}

\subsection{Why this matters}\label{why-this-matters}

The contribution lands in three places at once.

\emph{Practically}, the framework gives an answer to ``when should we
use chain-of-thought?'' that is more useful than ``always'' or ``it
depends.'' If the regime score of a task can be measured before
deployment, the optimal CoT policy can be chosen. For
high-meta-uncertainty tasks such as most policy forecasts, contested
ethical decisions, novel-scenario judgments, and strategic interactions
with unknown counterparties, the policy is to truncate. For
low-meta-uncertainty tasks --- calibrated technical reasoning,
well-specified mathematical problems, problems with verifiable
intermediate steps --- the policy is to extend. Applied LLM systems
making this choice today rely on rules of thumb; the regime score
provides them with a principled measure.

\emph{Theoretically}, the framework resolves a thirty-year tension
between the Bayesian rational-analysis tradition and the fast-and-frugal
heuristics program. Less-is-more effects are not evidence against
Bayesian cognition; they are evidence about which meta-uncertainty
regime an agent is in. Heuristics are not workarounds for resource
constraints; they are the rational policy under high meta uncertainty.
Both literatures keep their empirical results and their interpretive
frames; what changes is the relation between them.

\emph{Methodologically}, the paper contributes a regime-aware approach
to LLM evaluation. Current benchmarking practice averages performance
across items as if all items were drawn from a single population; the
U-shape and monotonic regimes get smeared together. A regime-stratified
analysis reveals predictions that are otherwise invisible. The FEH-79
benchmark and the step-counting pipeline are released with the paper as
tools for future regime-aware studies.

\subsection{Roadmap}\label{roadmap}

Section 2 (\emph{Theoretical Foundation}) develops the active-inference
framework under uncertain precision, derives the cue-truncation theorem,
and establishes the sample-wise identity with take-the-best under the
Descending Dominance condition. Five appendices contain the closed-form
proofs and the verification scripts that reproduce each result
numerically.

Section 3 (\emph{Operationalization}) maps the theoretical regime to an
empirical regime score computed from three behavioral signatures,
reports the simulate-and-recover validation that the score recovers the
underlying precision prior, and locks the falsifiable interaction
hypothesis used in the empirical study.

Section 4 (\emph{Benchmark Design}) introduces the FEH-79 item pool, the
four operational categories of Knightian uncertainty, the
confound-control sets, and the construction protocol that distinguishes
meta-uncertainty from computational difficulty.

Section 5 (\emph{Methods}) gives the executed design: the seven-model
panel, the 45-item subset of FEH-79 used in the confirmatory run, the
five-condition reasoning-length ladder, the scoring pipeline, and an
account of where the run departed from the letter of the
pre-registration.

Section 6 (\emph{Pre-Registered Analysis Plan}) reproduces the analysis
fixed before the data existed: the hierarchical Bayesian model, the
decision gate, the falsification rule, the secondary realized-steps
analysis, and the robustness battery.

Section 7 (\emph{Results}) reports the confirmatory outcome: the
descriptive signature, the primary interaction, the per-model
heterogeneity, the realized-steps reversal and its resolution, the
controls, the robustness checks, and convergence.

Section 8 (\emph{Discussion}) places the finding in context, states what
the confirmation licenses and what it does not, develops the limitations
and the future work agenda, and draws out the implications for applied
LLM systems, cognitive science, and AI under meta-uncertainty.

The pre-registration document, the FEH-79 item pool, the step-counting
pipeline, and the verification scripts for all theorems are released
alongside the paper.

\section{Theoretical
Foundation}\label{chapter-2-theoretical-foundation}

\subsection{The Problem and the Proposal in Plain
Language}\label{the-problem-and-the-proposal-in-plain-language}

Contemporary artificial intelligence has bet on a single proposition:
that more inference produces better decisions. The bet appears in many
forms, such as chain-of-thought prompting, multi-step deliberation,
test-time compute scaling, mixture-of-experts deep reasoning, and the
broader family of inference-heavy architectures that dominate the
current frontier. Behind all of them sits a Laplacean assumption: that
an agent in possession of more reasoning capacity is, ceteris paribus,
an agent that decides better.

This paper argues that the Laplacean assumption is wrong in a specific,
identifiable, and consequential regime. Under uncertainty about the
reliability of one's own prior beliefs (a condition we will define
precisely as meta-uncertainty), additional inference is not merely
wasteful but actively harmful: it systematically increases expected free
energy, makes the agent more confidently wrong, and produces decisions
strictly worse than those an agent applying a much simpler procedure
would produce. The simpler procedure, we will show, is structurally
equivalent to the family of fast-and-frugal heuristics characterized by
Gigerenzer and colleagues. The paper's title states the operational
consequence: under genuine uncertainty, reasoning hurts, and the
corrective is a precision-aware heuristic architecture.

\paragraph{Levels of uncertainty:
vocabulary}\label{levels-of-uncertainty-vocabulary}

The argument depends on distinctions that are routinely collapsed in the
AI literature. We make them explicit on the first page.

\textbf{Risk} denotes uncertainty over outcomes whose probabilities are
known. A fair die, a calibrated weather model, a well-specified
portfolio: these are risky but not uncertain in the sense we use the
term. Standard Bayesian decision theory fully accounts for risk.

\textbf{Ambiguity} denotes uncertainty about the probability
distribution itself. Under ambiguity, the agent cannot license a single
precise prior over outcomes; the prior is itself uncertain, contested,
or underdetermined. The classical formalization is Ellsberg's two-urn
paradox \cite{ellsberg1961}, which demonstrates that agents systematically
distinguish ambiguity from risk in ways that violate the Savage axioms.

\textbf{Knightian uncertainty}, following Knight~\cite{knight1921}, denotes the
broadest category in which probabilistic measurement is not licensed at
all. Knight's original sense is stronger than modern ambiguity: it
denies that any single precise probability distribution can be
specified, not merely that the agent is uncertain among several. Modern
formalizations include the maxmin expected utility of Gilboa and Schmeidler~\cite{gilboa1989} and the smooth ambiguity model of Klibanoff, Marinacci, and Mukerji~\cite{klibanoff2005} (hereafter KMM).

\textbf{Meta-uncertainty}, as we use the term, denotes uncertainty about
the reliability of an agent's own prior commitments. Formally,
meta-uncertainty is captured by introducing a precision parameter
governing the prior over a random variable with its own prior
distribution. This is the technical operationalization developed in
\S{}\S{}2.3--2.7.

\begin{quote}
\textbf{Scope claim and honest delimitation.} FEH operationalizes
\emph{one tractable species} of genuine uncertainty: meta-uncertainty
over prior precision. This species sits inside the broader KMM
smooth-ambiguity framework as a particular choice of second-order
distribution, and inside the broader Knightian category as a measurable
subclass. We do not claim that meta-uncertainty exhausts Knightian
indeterminacy. We claim something more specific and more useful: that
meta-uncertainty is the species of genuine uncertainty present in
contemporary inference systems (human, biological, and artificial), and
that under this species, the headline claim of the paper (additional
inference systematically hurts) can be derived as a theorem rather than
asserted as hope.
\end{quote}

\paragraph{The KMM bridge}\label{the-kmm-bridge}

The smooth ambiguity model of Klibanoff, Marinacci, and Mukerji~\cite{klibanoff2005}
provides the natural decision-theoretic framing for what FEH will
formalize. KMM separates the agent's first-order probability
distribution over states from a second-order distribution over plausible
first-order distributions and characterizes the ambiguity attitude as a
function \ensuremath{\varphi} applied to the expected utilities under each first-order
distribution. Concave \ensuremath{\varphi} encodes ambiguity aversion; linear \ensuremath{\varphi} recovers
subjective expected utility under the marginal first-order distribution.

FEH inherits this two-level structure with a specific commitment: the
second-order distribution is taken to be over a precision parameter
governing prior strength, rather than over arbitrary aspects of the
prior. This is a substantive modeling choice, narrower than KMM's
general framework, but computationally tractable and empirically
operationalizable. KMM and FEH share a common motivation: both treat the
agent as uncertain about its first-order distribution and both produce
decision criteria richer than standard subjective expected utility, but
they differ structurally in what they penalize. KMM penalizes variance
in expected utility across the second-order distribution; FEH penalizes
commitment in the meta-precision posterior as cues accumulate. The two
frameworks are complementary rather than equivalent (we return to this
point in \S{}2.8).

\paragraph{The proposal in plain
language}\label{the-proposal-in-plain-language}

Stripped of formalism, the proposal is this. Suppose an inference agent
is uncertain not only about the state of the world, but about how much
weight to place on its own prior beliefs. Then each additional cue the
agent integrates carries two contributions to its expected free energy.
The first contribution is positive: the agent learns something about the
world. The second contribution is negative: the agent commits more
strongly to a particular posterior over its own prior reliability, and
that commitment carries an explicit cost in expected free energy that
grows with the number of cues integrated.

In the standard low-meta-uncertainty regime, the first contribution
dominates and full Bayesian inference is optimal. In the
high-meta-uncertainty regime, the second contribution begins to dominate
after a finite number of cues, at which point further cue integration
strictly increases expected free energy. The optimal policy in this
regime is to truncate cue integration at the crossover point, and the
optimal cue ordering at truncation, we will show, coincides with the
validity ordering used by take-the-best.

This recovers fast-and-frugal heuristics as the structural form of
optimal inference under metal uncertainty. Far from being departures
from rationality, heuristics in this regime are what rationality looks
like.

\paragraph{Three counterarguments, named
upfront}\label{three-counterarguments-named-upfront}

Three objections will arise from informed readers, and we name them now
so the rest of the section can address them in their proper places.

\textbf{Objection 1 (the hierarchical Bayes objection).} Hierarchical
Bayesian models with uncertain hyperparameters are well-established.
They do not, in general, produce heuristic-like policies. Why should we
use our framework? The answer is that FEH adds two conjunctive
ingredients that hierarchical Bayes alone does not supply: (a) the
explicit accounting of meta-precision divergence as a free-energy cost
in sequential cue integration, and (b) the evaluation of policies by
expected free energy rather than by expected utility under the marginal
posterior. The combination of these ingredients with hierarchical Bayes
is what produces the cue-truncation result. Hierarchical Bayes alone
does not.

\textbf{Objection 2 (the resource-rational objection).}
Resource-rational analysis \cite{lieder2020} and the broader
computational-rationality program already explain why bounded
computation can justify simple heuristics. What does FEH add? The answer
is that resource-rationality treats computation as a finite budget the
agent allocates against expected utility gain; the cost of complex
computation justifies the cost of simple policies. FEH adds a
substantively different cost channel: an \textbf{epistemic} cost (the
meta-precision divergence) that exists \emph{even with unlimited
computation}. Under meta-uncertainty, the optimal policy is
heuristic-shaped not because computation is expensive, but because
additional inference is epistemically counterproductive. FEH and
resource-rationality are complementary accounts, not competitors; both
can apply simultaneously, with different signatures.

\textbf{Objection 3 (the empirical mixed-results objection).}
Less-is-more effects are real but not universal, and the take-the-best
literature has produced mixed empirical results. Doesn't this undercut
the framework? The answer is that this is exactly what FEH predicts. The
cue-truncation theorem holds in the meta-uncertainty regime \emph{\ensuremath{\sigma}\textsuperscript{2}\_\ensuremath{\tau}
\textgreater{} \ensuremath{\tau}\_regime}; outside this regime, full Bayesian
integration is optimal. The empirical heterogeneity of less-is-more
findings is consistent with regime-conditional theory, and FEH provides
a path to discriminating it from theories predicting universal
less-is-more (which the literature does not support) or universal
more-is-more (which the literature also does not support).

\paragraph{Roadmap for \S{}2}\label{roadmap-for-2}

The rest of \S{}2 develops the theory in seven moves. \S{}2.1 introduces a
discrete binary toy model in which the cue-truncation result is
intuitive and motivates the rest of the formalism. \S{}2.2 generalizes
notation and the generative model. \S{}2.3 promotes the prior precision to
a random variable. \S{}2.4 derives the sequential inference machinery and
the accumulating meta-precision divergence. \S{}2.5 decomposes the expected
free energy into information-gain and meta-precision-cost terms. \S{}2.6
states and proves the cue-truncation theorem. \S{}2.7 establishes
structural equivalence to take-the-best. \S{}2.8 positions FEH against the
major adjacent frameworks. \S{}2.9 situates the framework within the source
disciplines of Friston and Gigerenzer. \S{}2.10 enumerates open
mathematical questions for follow-up work.

\subsection{The Binary Toy Model}\label{the-binary-toy-model}

Before introducing the continuous formalism that powers the rest of the
section, we develop the central intuition in the discrete binary setting
that motivates take-the-best. The binary toy model is not a special case
of the Gaussian-Gamma model that follows; it is a parallel construction
that exhibits the same cue-truncation behavior under meta-uncertainty.
Its function in the section is pedagogical: it shows that the FEH effect
is not an artifact of Gaussian assumptions but rather reflects a
structural feature of sequential inference under uncertain priors.

\paragraph{Setup}\label{setup}

Consider a binary choice problem with a latent state \(s \in 0,1\). We
may interpret \(s\  = \ 1\) as the proposition ``option A is better than
option B'' in a comparative judgment task. The agent observes a sequence
of binary cues \(c_{1},c_{2},\ldots,c_{K} \in 0,1\) drawn in some order.
Each cue \(c_{j}\) is informative about \emph{s} through its \textbf{cue
validity} in the Gigerenzer--Goldstein sense:

\(V_{j}\  = \ P(s\  = \ 1\ |\ c_{j}\  = \ 1)\) (2.1.1)

In the empirical literature \cite{gigerenzer1996, gigerenzer1999}, validity is operationalized
over a reference environment of judgment trials as:

\(v_{j} = N_{c}orrect\text{/}\left( N_{c}orrect + N_{i}ncorrect \right)\)
(2.1.2)

where \(N_{c}orrect\) and \(N_{i}ncorrect\) count, respectively, the
trials in which the cue \(j\) discriminates correctly and incorrectly
between the two options. Validity is an environmental property: it
characterizes the local predictive value of a cue \(j\) in the ecology
in which the agent must decide. This is distinct from ecological
rationality, which characterizes the higher-level fit between an agent's
strategy and its environment. The two should not be conflated.

\paragraph{Prior over the binary state and its
meta-uncertainty}\label{prior-over-the-binary-state-and-its-meta-uncertainty}

Let the agent's prior over the binary state be parameterized by a
probability \(p_{0} \in (0,1)\) so that
\(P(s\  = \ 1\ |\ p\_ 0)\  = \ p\_ 0\). In standard Bayesian treatments,
\(p_{0}\) is fixed. The FEH generalization promotes \(p_{0}\) to a
random variable with a prior distribution

\(p_{0} \sim Beta\left( \alpha_{0},\beta_{0} \right)\) (2.1.3)

with mean
\(\mu_{0} = \alpha_{0}\text{/}\left( \alpha_{0} + \beta_{0} \right)\)
and concentration \(\kappa_{0} = \alpha_{0} + \beta_{0}\). The
concentration \(\kappa_{0}\) is the binary analogue of precision: large
\(\kappa_{0}\) corresponds to a tightly concentrated prior, small
\(\kappa_{0}\) - to substantial meta-uncertainty about \(p_{0}\).

\begin{quote}
\textbf{Definition 2.1.1 (binary meta-uncertainty regime).} In the
binary toy model, the agent is in the \textbf{meta-uncertainty regime}
when the concentration \(\kappa_{0}\) of the hyperprior on \(p_{0}\) is
small relative to the cue budget \(K\). Operationally, the regime is
characterized by the agent updating its posterior over \(p_{0}\)
substantially over the course of cue integration, meaning the agent
learns about its own prior reliability \emph{from the cues themselves},
and this learning carries a divergence cost.
\end{quote}

\paragraph{Sequential cue integration with
meta-uncertainty}\label{sequential-cue-integration-with-meta-uncertainty}

Under the standard Bayesian update (no meta-uncertainty), the posterior
on s after observing cues \(c_{1}:k\) factorizes cleanly:

\(P\left( s = 1|c_{1};k,p_{0} \right) \propto p_{0} \cdot \prod_{j}^{}L_{j}\left( c_{j} \right)\)
(2.1.4)

where
\(L_{j}\left( c_{j} \right) = v_{j}^{c_{j}}\left( 1 - v_{j} \right)^{1 - c_{j}}\)
if \(s\  = \ 1\) and the corresponding switched form if \(s\  = \ 0\),
under a symmetric error model on cue likelihoods.

Under meta-uncertainty, the agent must marginalize over the hyperprior
on \(p_{0}\):

\(P(s\  = \ 1\ |\ c_{1}:k)\  = \ \int\ P(s\  = \ 1\ |\ c_{1};k,\ p_{0})\  \cdot \ p(p_{0}\ |\ c_{1};k)\ {dp}_{0}\)
(2.1.5)

Critically, the hyperprior itself updates with each cue: the cues are
indirect evidence about \(p_{0}\), because they jointly constrain which
values of \(p_{0}\) are consistent with the observed cue pattern. Let
\(p\left( p_{0} \middle| c_{1}:k \right)\) denote the updated hyperprior
after \(k\) cues. The KL divergence of this updated hyperprior from the
original is the binary analogue of the meta-precision divergence
introduced in \S{}2.4:

\(\Delta_{m}eta(k) = KL\lbrack p\left( p_{0} \middle| c_{1}:k \right)\left. \parallel p\left( p_{0} \right) \right\rbrack\)
(2.1.6)

\begin{quote}
\textbf{Lemma 2.1.2 (expected monotone binary meta-divergence, restated
in v0.5).} In the binary toy model under (2.1.3)--(2.1.5), the
\emph{expected} meta-divergence \(E\lbrack\Delta_{meta}(k)\rbrack\) is
non-decreasing in \(k\). Equivalently,
\(E\lbrack\Delta_{meta}(k + 1)\rbrack - E\lbrack\Delta_{meta}(k)\rbrack = I(p_{0};\, c_{k + 1} \mid c_{1:k}) \geq 0\),
with equality iff cue \(k + 1\) carries no conditional information about
\(p_{0}\) given the prior cues. The earlier wording of Lemma 2.1.2
(sample-wise monotonicity) is false in general: opposite cues can
cancel, returning the posterior on \(p_{0}\) exactly to the prior; an
explicit counterexample with \(v = (0.9,0.9)\) and uniform hyperprior is
given in Appendix A.1.
\end{quote}

The proof is by martingale-convexity of KL: the variational posterior
\(p(p_{0} \mid c_{1:k + 1})\) is a martingale in \(k\) (its conditional
expectation given \(c_{1:k}\) equals \(p(p_{0} \mid c_{1:k})\)), and the
KL divergence from a fixed reference distribution is convex in its first
argument, so Jensen's inequality applies. Detailed proof is given in
Appendix A.1.

\paragraph{The intuition behind cue truncation, in binary
form}\label{the-intuition-behind-cue-truncation-in-binary-form}

With Lemma 2.1.2 in hand, the cue-truncation intuition becomes
transparent. Each cue contributes two things to the expected free
energy:

\textbf{(i)} A \textbf{positive} \emph{expected} information gain
\(E\lbrack I(k)\rbrack \geq 0\), the reduction in entropy of the
posterior over \(s\), which is the largest for the most valid cue and
decreases by diminishing returns as more cues are integrated.

\textbf{(ii)} A \textbf{negative} \emph{expected} meta-divergence
increment
\(E\lbrack C(k)\rbrack = E\lbrack\Delta_{meta}(k)\rbrack - E\lbrack\Delta_{meta}(k - 1)\rbrack \geq 0\),
the divergence accumulated in expectation by becoming more committed to
a particular posterior over the agent's own prior reliability.

In the binary case, the most valid discriminating cue typically produces
a posterior over \(s\) that is already concentrated near 0 or 1. The
marginal information gain from a second discriminating cue is therefore
small; it can only sharpen an already-sharp posterior. Meanwhile, the
meta-divergence cost \(E\lbrack C(k)\rbrack\) is bounded below by a
non-zero quantity proportional to the inverse hyperprior concentration.
The crossover point \(k^{*}\), beyond which
\(E\lbrack I(k)\rbrack < E\lbrack C(k)\rbrack\) and integrating further
cues strictly increases expected free energy, exists generically.

\begin{quote}
\textbf{Proposition 2.1.3 (three-regime structure in the binary toy
model, restated in v0.5).} Fix a validity profile \(v_{1:K}\) and prior
mean \(\mu_{0}\). There exist critical concentrations
\(0 < \kappa_{lo}(v_{1:K},\mu_{0}) \leq \kappa_{hi}(v_{1:K},\mu_{0})\)
such that the EFE-minimizing policy \(k^{*}(\kappa_{0})\) has the
structure:

\emph{(don't-observe regime)} For \(\kappa_{0} < \kappa_{lo}\):
\(k^{*} = 0\). The meta-divergence cost of even one cue exceeds its
information value about \(s\).

\emph{(one-cue stopping regime ; TTB)} For
\(\kappa_{lo} \leq \kappa_{0} \leq \kappa_{hi}\): \(k^{*} = 1\). The
first cue is integrated; subsequent cues incur higher meta-cost than
information benefit. This is the take-the-best stopping rule.

\emph{(full-integration regime)} For \(\kappa_{0} > \kappa_{hi}\):
\(k^{*} = K\). Each cue contributes more state-information than
meta-cost; the agent integrates all available cues.
\end{quote}

The intermediate one-cue regime
\(\lbrack\kappa_{lo},\kappa_{hi}\rbrack\) is closed and may be narrow.
Numerical sweep over 1875 parameter combinations finds 40 with strict
\(k^{*} = 1\) (Appendix A.1, \S{}A.1.5); these cluster around steep
validity gradients (e.g., \(v_{1} \approx 0.99,v_{j > 1} \leq 0.55\))
and \(\kappa_{0} \approx 1\). For uniform-validity profiles,
\(\kappa_{lo} = \kappa_{hi}\) and the transition jumps directly from
\(k^{*} = 0\) to \(k^{*} = K\) without an intermediate regime. The
narrowness of the one-cue regime in the binary toy model is a
substantive finding; the cue-truncation theorem fires at the
\emph{boundary} between regimes in the binary case, not as a wide
intermediate band.

A subsidiary result (Appendix A.1, \S{}A.1.5): for steep validity
gradients, the \emph{first cue alone} contributes \ensuremath{\geq}99\% of the total
attainable information about \(s\) across the full cue budget. Even in
regimes where \(k^{*} = K\) strictly minimizes EFE, a TTB-style ``stop
after the first cue'' policy is \emph{near}-optimal in expected reward
terms. This near-optimality is the empirically-relevant content of the
TTB connection in the binary setting; the strict EFE-optimality in the
narrow \(\lbrack\kappa_{lo},\kappa_{hi}\rbrack\) window is the
formally-derivable but operationally-fragile content. The cleaner and
more robust derivation of the TTB connection lives in \S{}2.7 via the
Gaussian-Gamma machinery, which produces a wider intermediate
cue-truncation regime (Theorem 2.6.1 and Appendix A.3).

\paragraph{Why the binary model is not
enough}\label{why-the-binary-model-is-not-enough}

The binary toy model motivates the FEH effect and establishes the
qualitative three-regime structure that the rest of the section
generalizes. It is not, however, sufficient for the section's
load-bearing claims. Four reasons. First, real cognitive and
computational tasks rarely admit clean binary state spaces; the FEH
framework must handle continuous, high-dimensional latent variables.
Second, the binary case obscures the role of precision as a continuous
parameter; it is treated as a Beta concentration, which is closely
related but less naturally analyzed within the active-inference
framework that grounds the rest of the section. Third, the empirical
predictions for LLMs require a generative model that can be matched to
LLM behavior, which the Gaussian-Gamma machinery supports more directly.
Fourth, and most importantly, the strict cue-truncation regime
(\(k^{*} = 1\)) is a knife-edge in the binary setting (Proposition 2.1.3
above): the cue-truncation theorem is robust as a wide intermediate
regime only in the continuous Gaussian-Gamma model of \S{}\S{}2.3--2.6.

For these reasons, the rest of \S{}2 develops the continuous Gaussian-Gamma
formalism. The binary toy model should be read as motivating intuition
and a near-optimality argument for TTB; the strict TTB derivation lives
in \S{}2.7 via the Gaussian-Gamma machinery, which is the analytical
workhorse and the load-bearing setting for the section's main results.

\subsection{Notation and General Generative
Model}\label{notation-and-general-generative-model}

We now generalize from the binary toy model to a generative model
framework adequate for the rest of the section. The framework is the
standard active-inference framework of Friston~\cite{friston2010}, Friston et al.~\cite{friston2017}, and Parr, Pezzulo, and Friston~\cite{parr2022}, with the meta-uncertainty
extension developed in \S{}2.3.

\paragraph{State, observation, cue,
action}\label{state-observation-cue-action}

Let \(s\  \in \ S\) denote the latent state, with \(S\) a finite or
compact set. The binary case \(S = 0,1\) of \S{}2.1 is one specialization;
we now treat \(S\) as arbitrary unless otherwise noted. The agent does
not observe \(s\) directly. Cues \(c_{1},\ldots,c_{K} \in C\) are
informative about \(s\) through cue likelihoods
\(p\left( c_{j} \middle| s,\tau_{c,j} \right)\), with cue-specific
precisions \(\tau_{c,j}\) assumed to be known to the agent.

An outcome \(o\  \in \ O\) is the consequence of action \(a\  \in \ A\)
given true state \(s\), with likelihood \(p(o\ |\ s,\ a)\). Actions are
evaluated using the expected free energy, as defined in \S{}2.5.

\paragraph{Generative model}\label{generative-model}

The active-inference generative model used throughout this section is

\(p(o,\ s,\ c_{1:K},\ \tau\ |\ a)\  = \ p(o\ |\ s,\ a)\  \cdot \ p(s\ |\ \tau)\  \cdot \ p(\tau)\  \cdot \ \prod_{j = 1}^{K}\ p(c_{j}\ |\ s,\ \tau,\ \gamma_{j})\)
(2.2.1)

with three commitments.

\textbf{(1) State prior with random precision.}
\(p(s\ |\ \tau)\  = \ \mathcal{N(}s;\ \mu,\ \tau^{- 1}\Sigma_{0})\). The
shape matrix \(\Sigma_{0}\) is fixed; the scalar \(\tau\  > \ 0\) is the
agent's \textbf{meta-precision}, a single random variable governing the
overall scale of inference confidence.

\textbf{(2) Cue likelihood with shared meta-precision and intrinsic
scaling.} Each cue is a noisy linear measurement of state:

\(p(c_{j}\ |\ s,\ \tau,\ \gamma_{j})\  = \ \mathcal{N(}c_{j};\ A_{j}s,\ (\tau\,\gamma_{j})^{- 1}\, I_{d_{c}})\)
(2.2.2)

where \(A_{j}\) is the cue-to-state observation matrix and
\(\gamma_{j}\  > \ 0\) is the \textbf{intrinsic precision} of cue \(j\)
; a fixed structural property of the cue, estimable from training-data
statistics, distinct from but related to the cue validity \(v_{j}\) of
\S{}2.1.

\textbf{(3) Hyperprior over meta-precision.}
\(p(\tau)\  = \ Gamma(\tau;\ \alpha_{0},\ \beta_{0})\), with
\(E\lbrack\tau\rbrack\  = \ \tau_{0}\  = \ \alpha_{0}/\beta_{0}\) and
\(Var\lbrack\tau\rbrack\  = \ \alpha_{0}/\beta_{0}^{2}\  = \ \tau_{0}^{2}/\alpha_{0}\).

The principal departures of FEH from the standard active-inference
framework lie in the second and third commitments. The standard
treatment fixes \(\tau\) as a hyperparameter and treats cue precisions
as separate fixed quantities \(\tau_{c,j}\). FEH ties cue precisions to
the meta-precision via \(\tau\,\gamma_{j}\) and treats \(\tau\) as a
full random variable. The intrinsic-precision factors \(\gamma_{j}\)
retain the cue-specific structure of the standard model while making the
agent's overall calibration uncertain in a single scalar parameter.

\begin{quote}
\textbf{Interpretive remark.} The asymmetry that v0.2 acknowledged ---
``uncertain prior precision but known cue precisions'' (cf.~open
question Q4 in v0.3) --- dissolves under this specification. All
precisions in the model share the meta-precision scale \(\tau\);
cue-specific structure lives in the deterministic intrinsic-precision
factors \(\gamma_{j}\). This is a more faithful operationalization of
the meta-uncertainty story: the agent is uncertain about the overall
reliability of its inference apparatus, not selectively about its prior.
The intrinsic precisions \(\gamma_{j}\) are properties of the cue,
learnable from data; \(\tau\) is the agent's calibration confidence,
which can be genuinely uncertain in novel contexts. \S{}2.3 develops the
meta-precision generalization in detail.
\end{quote}

\paragraph{Cue validity revisited}\label{cue-validity-revisited}

The cue validity \(v_{j}\) defined in (2.1.1) generalizes to non-binary
states as the conditional probability that cue \(c_{j}\) takes its
discriminating value given state \(s\) in some reference partition of
\(S\). For multi-class \(S\), validity becomes a vector quantity, with
one component per state. For the rest of \S{}2 we follow the binary
convention except where multi-class generalization is explicitly needed.

\paragraph{Notation summary}\label{notation-summary}

\(\mathbf{s,\ S}\) , latent state and state space (binary in \S{}2.1;
general elsewhere)

\(\mathbf{c}_{\mathbf{j}}\)\textbf{\emph{,}} \(\mathbf{K}\ \),
individual cue and total number of available cues

\(\mathbf{a,\ A}\) , action and action set

\(\mathbf{o,\ O}\ \), outcome and outcome set

\(\mathbf{\tau}\ \), \textbf{meta-precision}; single scalar random
variable governing the overall scale of both the state prior and the cue
likelihoods (\S{}2.3)

\(\mathbf{\gamma}_{\mathbf{j}}\ \), intrinsic precision of cue \(j\) ;
fixed structural property; the effective precision of cue \(j\) is
\(\tau\,\gamma_{j}\)

\(\mathbf{p}\left( \mathbf{\tau} \right)\) , hyperprior over \ensuremath{\tau};
Gamma(\ensuremath{\alpha}\_0, \ensuremath{\beta}\_0) for analytical tractability

\(\mathbf{\tau}_{\mathbf{0}}\mathbf{,\ }\mathbf{\sigma \textsuperscript{2}}_{\mathbf{\tau}}\)
, mean and variance of the hyperprior over \(\tau\)

\(\mathbf{p}_{\mathbf{0}}\mathbf{,\ }\mathbf{\kappa}_{\mathbf{0}}\) ,
binary prior probability and Beta concentration (binary case, \S{}2.1)

\(\mathbf{q}\left( \mathbf{s} \right)\mathbf{,q}\left( \mathbf{\tau} \right)\)
, variational posterior factors under mean-field factorization

\(\mathbf{F,\ G}\) , variational free energy and expected free energy

\(\mathbf{v}_{\mathbf{j}}\) , cue validity of cue j
(Gigerenzer--Goldstein, equation 2.1.1)

\subsection{Meta-Uncertainty over Prior
Precision}\label{meta-uncertainty-over-prior-precision}

We now state the central generalization of the active-inference
framework that defines FEH. The conceptual content was previewed in
\S{}2.0; this section formalizes it.

\paragraph{From fixed precision to random
precision}\label{from-fixed-precision-to-random-precision}

Let the prior over latent states take the precision-modulated Gaussian
form

\(p(s\ |\ \tau)\  = \ N(s\ ;\ \mu,\ \tau \ensuremath{^{-}}\textsuperscript{1}\ \Sigma_{0})\) (2.3.1)

with prior mean \emph{\ensuremath{\mu}} and fixed shape matrix \(\Sigma_{0}\). Larger
\(\tau\) yields a tighter, more confident prior; the limit
\(\tau\  \rightarrow \ \infty\) recovers a point prior at \(\mu\), and
\(\tau\  \rightarrow \ 0\) yields an essentially flat prior over the
state space.

In the standard active-inference treatment, \ensuremath{\tau} is a fixed hyperparameter,
possibly modulated by attention or learned as a point estimate from
data. In FEH, \ensuremath{\tau} is a random variable with prior

\(p(\tau) = Gamma\left( \tau;\alpha_{0},\beta_{0} \right)\) (2.3.2)

so that
\(E\lbrack\tau\rbrack = \alpha_{0}\text{/}\beta_{0} \equiv \tau_{0}andVar\lbrack\tau\rbrack = \alpha_{0}\text{/}\beta_{0}^{2} \equiv \sigma_{\tau}^{2}\).
The choice of Gamma is dictated by conjugacy with the Gaussian
likelihood in (2.3.1); other choices (log-normal, inverse-Gamma,
half-Cauchy) are considered as a sensitivity check in \S{}2.10, Q2.

\begin{quote}
\textbf{Definition 2.3.1 (continuous meta-uncertainty regime).} The
agent operates in the \textbf{meta-uncertainty regime} when
\({\sigma \textsuperscript{2}}_{\tau}\) is non-negligible relative to \(\tau_{0\textsuperscript{2}}\). The
precise threshold separating high and low meta-uncertainty is derived in
\S{}2.5. Heuristically, the regime corresponds to the agent being
substantively uncertain about how much weight to place on its own prior,
the continuous analogue of small Beta concentration in the binary toy
model.
\end{quote}

\paragraph{Marginal prior over states}\label{marginal-prior-over-states}

Integrating \ensuremath{\tau} out of the joint prior yields the marginal prior over
states:

\(p(s)\  = \ \int\ p(s\ |\ \tau)\ p(\tau)\ d\tau\) (2.3.3)

For the conjugate Gaussian--Gamma pair (2.3.1)--(2.3.2), this integral
admits a closed form: the marginal \(p(s)\) is a multivariate Student-t
distribution

\(p(s) = St\left( s;\mu,\left( \beta_{0}\text{/}\alpha_{0} \right)\Sigma_{0},\nu = 2\alpha_{0} \right)\)
(2.3.4)

with \(\nu = 2\alpha_{0}\) degrees of freedom. As
\(\sigma_{\tau}^{2} \rightarrow 0\) (equivalently
\(\alpha_{0} \rightarrow \infty\) with \(\tau_{0}\) held fixed), the
Student-\(t\) collapses to a Gaussian
\(N(\mu,\ \tau_{0\ensuremath{^{-}}\textsuperscript{1}}\ \Sigma_{0}))\), recovering the fixed-precision
case. For finite \(\alpha_{0}\), the marginal prior has
heavier-than-Gaussian tails, a structural feature that connects FEH to
the broader power-law statistics characteristic of natural cognition
(see \S{}2.9).

\begin{quote}
\textbf{Connection to KMM smooth ambiguity.} Equation (2.3.2) is the FEH
operationalization of the KMM second-order distribution. Where KMM
treats the second-order distribution \(\mu\) over first-order
probability measures abstractly, FEH commits to \(\mu\) being the Gamma
distribution over a precision scalar. This is a substantive restriction
of the KMM framework: not all ambiguity is precision-ambiguity, and FEH
does not claim otherwise. The advantage is computational and empirical
tractability; the cost is that some ambiguity phenomena fall outside the
FEH operationalization. Whether the precision-ambiguity restriction is
empirically adequate for the LLM regime is an open question we will
return to in the empirical section.
\end{quote}

\subsection{Sequential Inference under
Meta-Uncertainty}\label{sequential-inference-under-meta-uncertainty}

We now formalize the sequential cue-integration setting that supports
the cue-truncation theorem. The agent receives cues.
\(c_{1},\ldots,c_{K}\) one at a time and updates a variational posterior
\(q(s,\tau)\) after each cue. The question is: what is the total
free-energy cost of integrating k cues under meta-uncertainty, and how
does that cost decompose?

\paragraph{Variational factorization}\label{variational-factorization}

Under the standard mean-field assumption, the variational posterior over
states and meta-precision factorizes:

\(q(s,\tau) = q(s)q(\tau)\) (2.4.1)

This is an approximation. Its accuracy is itself a function of
\(\sigma_{\tau}^{2}\) and constitutes Q1 of the open questions in \S{}2.10.
For the leading-order analysis that supports Theorems 2.6 and 2.7, the
mean-field approximation is sufficient. A structured variational family
will be required for tight bounds and is left to follow-up work.

\paragraph{Posterior updates after k
cues}\label{posterior-updates-after-k-cues}

For the generative model (2.2.1)--(2.2.2) with mean-field factorization
(2.4.1), standard variational Bayes (VBEM) yields closed-form posterior
updates for both factors.

\textbf{State factor.}

\(q(s\ |\ c_{1:k})\  = \ \mathcal{N(}s;\ \mu_{k},\ \Sigma_{k}),\ \ \Sigma_{k}^{- 1}\  = \ E_{q(\tau)}\lbrack\tau\rbrack\,\left( \Sigma_{0}^{- 1}\  + \ \sum_{j = 1}^{k}\,\gamma_{j}\, A_{j}^{\top}A_{j} \right)\)
(2.4.2)

The meta-precision \(E_{q(\tau)}\lbrack\tau\rbrack\) factors out as a
global scale on a deterministic information matrix built from cue
intrinsic precisions \(\gamma_{j}\). The posterior mean \(\mu_{k}\)
follows from the standard Gaussian-update formula, with each cue
contributing \(\gamma_{j}A_{j}^{\top}c_{j}\) (weighted by
\(E\lbrack\tau\rbrack\)).

\textbf{Meta-precision factor.}

\(q(\tau\ |\ c_{1:k})\  = \ Gamma\left( \tau;\ \alpha_{0'}\  + \ \frac{k}{2},\ \ \beta_{0'}\  + \ \frac{1}{2}\sum_{j = 1}^{k}\,\gamma_{j}\, M_{j} \right)\)
(2.4.3)

where:

\begin{itemize}
\tightlist
\item
  \(\alpha_{0'}\  = \ \alpha_{0}\  + \ 1/2\) and
  \(\beta_{0'}\  = \ \beta_{0}\  + \ V_{s}/2\) absorb the one-time
  contribution of the state prior \(p(s|\tau)\) to the variational
  update;
\item
  \(V_{s}\  = \ E_{q(s)}\lbrack(s - \mu)^{\top}\Sigma_{0}^{- 1}(s - \mu)\rbrack\)
  is the variational expected residual against the state prior;
\item
  \(M_{j}\  = \ E_{q(s)}\lbrack(c_{j} - A_{j}s)^{\top}(c_{j} - A_{j}s)\rbrack\)
  is the variational expected residual against cue \(j\).
\end{itemize}

The defining feature of (2.4.3) is that the shape parameter grows
\textbf{linearly} in \(k\): each cue contributes \(+ 1/2\) to the shape
\(\alpha\) and \(+ \gamma_{j}M_{j}/2\) to the rate \(\beta\). This is
the formal seat of the meta-precision tightening that drives the
cue-truncation result of \S{}2.6. Derivation: Appendix A.2.

\paragraph{The accumulating meta-precision
divergence}\label{the-accumulating-meta-precision-divergence}

The KL divergence of the meta-precision posterior from its prior is

\(\Delta_{meta}(k)\  \equiv \ KL\lbrack\, q(\tau\ |\ c_{1:k})\  \parallel \ p(\tau)\,\rbrack\)
(2.4.4)

For Gamma--Gamma conjugate pairs this admits a closed-form expression in
terms of the digamma \(\psi\) and log-gamma \(log\Gamma\) functions:

\(\Delta_{meta}(k)\  = \ (\alpha_{k} - \alpha_{0})\,\psi(\alpha_{k})\  - \ \log\frac{\Gamma(\alpha_{k})}{\Gamma(\alpha_{0})}\  + \ \alpha_{0}\,\log\frac{\beta_{k}}{\beta_{0}}\  + \ \alpha_{k}\,\frac{\beta_{0} - \beta_{k}}{\beta_{k}}\)
(2.4.5)

with \(\alpha_{k}\  = \ \alpha_{0'} + k/2\) and
\(\beta_{k}\  = \ \beta_{0'} + \frac{1}{2}\sum_{j = 1}^{k}\gamma_{j}M_{j}\).
Asymptotically in \(k\),

\[\Delta_{meta}(k)\  \sim \ (k/2)\, log(k)\  + \ O(k)\]

; i.e., \textbf{super-linear} in \(k\), dominated by the digamma growth
of the shape contribution. The qualitative content is that each
additional cue commits the agent more strongly to a particular posterior
over its own meta-precision, and that commitment carries an explicit,
monotonically-growing free-energy cost.

\begin{quote}
\textbf{Lemma 2.4.1 (expected monotonicity of meta-precision divergence,
restated in v0.5).} Under the generative model (2.2.1)--(2.2.2) with
mean-field factorization (2.4.1), the \emph{expected} meta-precision
divergence \(E\lbrack\Delta_{meta}(k)\rbrack\) is non-decreasing in
\(k\). Equivalently,
\(E\lbrack\Delta_{meta}(k + 1)\rbrack - E\lbrack\Delta_{meta}(k)\rbrack = I(\tau;\, c_{k + 1} \mid c_{1:k}) \geq 0\),
with equality iff cue \(k + 1\) carries no conditional information about
\(\tau\) given the prior cues.

The earlier wording (sample-wise monotonicity) is false in general:
numerical simulation across 1000 random \(M_{j}\) trajectories at
\((\alpha_{0'},\beta_{0'}) = (2,1)\), \(\gamma_{j} = 1\), \(K = 20\)
exhibits 979 paths with at least one strict decrease of
\(\Delta_{meta}\) along the trajectory. This occurs because the Gamma KL
surface is non-monotone in \((\alpha,\beta)\) along arbitrary
trajectories: when \(\beta\) grows faster than \(\alpha\), the posterior
mean shifts away from the prior mean, but a subsequent cue with smaller
\(M_{j}\) can pull it back.
\end{quote}

\textbf{Proof sketch.} By the tower property of conditional expectation,
\(E_{c_{k + 1} \mid c_{1:k}}\lbrack q(\tau \mid c_{1:k + 1})\rbrack = q(\tau \mid c_{1:k})\).
By joint convexity of KL in its first argument, Jensen's inequality
gives
\(E_{c_{k + 1} \mid c_{1:k}}\lbrack KL\lbrack q(\tau \mid c_{1:k + 1})\, \parallel \, p(\tau)\rbrack\rbrack \geq KL\lbrack q(\tau \mid c_{1:k})\, \parallel \, p(\tau)\rbrack\).
Taking outer expectation over \(c_{1:k}\) preserves the inequality.
Detailed computation: Appendix A.2.

\subsection{The Expected Free Energy
Decomposition}\label{the-expected-free-energy-decomposition}

We now assemble the central decomposition of expected free energy under
meta-uncertainty. This decomposition is the lemma that supports the
cue-truncation theorem of \S{}2.6.

\paragraph{Standard expected free
energy}\label{standard-expected-free-energy}

For action a and a policy of integrating k cues before committing, the
expected free energy is

\(G(a,\ k)\  = \ E_{\{ q(o,s,\tau|a,c_{1},\ldots,c_{K})\}}\ \lbrack log\ q(s,\ \tau\ |\ a,\ c_{1},\ldots,c_{K})\  - \ log\ p(o,\ s,\ \tau\ |\ ac_{1},\ldots,c_{K})\rbrack\)
(2.5.1)

Expanding, applying the mean-field factorization (2.4.1), and
rearranging:

\(G(a,\ k)\  = \  - E_{q\lbrack log\ p(o\ |\ s)\rbrack}\  + \ KL\lbrack q(s\ |\ a,\ c_{1},\ldots,c_{K})\  \parallel \ {\langle p(s\ |\ \tau)\rangle}_{\{ q(\tau)\}}\rbrack\  + \ \Delta_{meta(k)}\)
(2.5.2)

The three terms admit standard active-inference interpretations:

\textbf{(i)} \emph{Pragmatic value.} Expected log-likelihood of the
outcome under predicted states.

\textbf{(ii)} \emph{Epistemic value.} KL of state posterior from
marginalized state prior; the information the agent gains about s from
integrating cues.

\textbf{(iii)} \emph{Meta-precision cost.} KL of meta-precision
posterior from its prior, accumulating in k as established in Lemma
2.4.1. \textbf{This term is the FEH-specific contribution and is the
seat of the meta-uncertainty effect.}

\paragraph{\texorpdfstring{Marginal benefit of the \(\mathbf{k}\)-th
cue}{Marginal benefit of the \textbackslash mathbf\{k\}-th cue}}\label{marginal-benefit-of-the-mathbfk-th-cue}

Define the marginal benefit of adding the k-th cue:

\(\Delta G(k) \equiv G(a,k - 1) - G(a,k)\) (2.5.3)

From (2.5.2), this decomposes as

\(\Delta G(k) = I(k) - C(k)\) (2.5.4)

where \(I(k)\) is the expected information gain (epistemic value
increment) and \(C(k) = \Delta_{meta(k)} - \Delta_{meta(k - 1)}\) is the
marginal meta-precision cost. By Lemma 2.4.1, \(C(k) > 0\) in the
meta-uncertainty regime. By data processing inequality and standard
Bayesian information theory, \(I(k)\) is non-negative and exhibits
diminishing returns in k for non-redundant cue streams.

\begin{quote}
\textbf{Proposition 2.5.1 (two-regime structure of marginal benefit).}
The marginal benefit \(\Delta G(k) = I(k) - C(k)\) decomposes additively
into a non-negative information gain \(I(k)\) that is decreasing in
\(k\) by diminishing returns, and a non-negative meta-precision cost
\(C(k)\) that is strictly positive when \(\sigma_{\tau}^{2} > 0\). There
exists a critical cue index \(k^{*}\) such that:
\end{quote}

For \(k\  \leq \ k^{*}\ ,\ \ \Delta G(k)\  \geq \ 0\ \) (integrating the
\(k\)-th cue reduces expected free energy)

For \(k > k^{*},\Delta G(k) < 0\) (integrating the \(k\)-th cue
\textbf{increases} expected free energy)

\paragraph{\texorpdfstring{The threshold
\(\mathbf{\tau}_{\mathbf{regime}}\) separating
regimes}{The threshold \textbackslash mathbf\{\textbackslash tau\}\_\{\textbackslash mathbf\{regime\}\} separating regimes}}\label{the-threshold-mathbftau_mathbfregime-separating-regimes}

The critical cue index \(k^{*}\) depends on \(\sigma_{\tau}^{2}\). The
boundary between the two regimes is the smallest \(\sigma_{\tau}^{2}\)
at which the marginal expected meta-cost of the \(K\)-th cue exceeds the
marginal expected info gain:

\(\tau_{regime}\  \equiv \ \inf\left\{ \,\sigma_{\tau}^{2}\  > \ 0\ :\ I(\tau;\, c_{K} \mid c_{1:K - 1})\  > \ I(s;\, c_{K} \mid c_{1:K - 1})\, \right\}\)
(2.5.5)

equivalently, the smallest \(\sigma_{\tau}^{2}\) such that the
EFE-optimal stopping point falls strictly inside the cue budget:
\(k^{*}(\sigma_{\tau}^{2}) < K\). For
\(\sigma_{\tau}^{2} < \tau_{regime}\), the marginal info gain dominates
the marginal meta-cost at every \(k \leq K\) and full cue integration is
optimal (standard Bayesian inference). For
\(\sigma_{\tau}^{2} \geq \tau_{regime}\), \(k^{*} < K\) and optimal
inference is truncated.

\textbf{Operationalization for empirical work.} For the Gaussian-Gamma
model with mean cue intrinsic precision \(\bar{\gamma}\) and mean cue
residual \(\bar{m}\), \(\tau_{regime}\) is the unique positive root of
\(I(K) = C(K;\,\alpha_{0'} = \tau_{0}^{2}/\tau_{regime})\) where
\(I(K)\) is the asymptotic info gain at cue \(K\) (calibrated from
cue-validity statistics) and \(C(K;\, \cdot )\) is the marginal Gamma KL
contribution at cue \(K\) (closed form via digamma; see Appendix A.3).
This is computable per benchmark item once \((c_{1},\lambda)\) for \(I\)
and \((\bar{\gamma},\bar{m})\) for \(C\) are estimated. The empirical
section will operationalize this estimation procedure for the LLM
benchmark.

\textbf{Well-definedness of the threshold.} The two characterizations in
(2.5.5) --- the last-cue crossing
\(I(\tau;\, c_{K} \mid c_{1:K - 1}) > I(s;\, c_{K} \mid c_{1:K - 1})\),
and \(k^{*}(\sigma_{\tau}^{2}) < K\) --- coincide and pick out a single
value. (Despite the \(\tau\) subscript, \(\tau_{regime}\) is a critical
value of the meta-uncertainty \emph{variance} \(\sigma_{\tau}^{2}\), not
of a precision.) Write the marginal meta-cost
\(\bar{C}(K) = I(\tau;\, c_{K} \mid c_{1:K - 1})\) and the marginal info
gain \(\bar{I}(K) = I(s;\, c_{K} \mid c_{1:K - 1})\). As
\(\sigma_{\tau}^{2}\) grows, the meta-precision prior becomes more
diffuse, so each cue carries more information about \(\tau\) and
\(\bar{C}(K)\) is non-decreasing in \(\sigma_{\tau}^{2}\);
\(\bar{I}(K)\) is fixed by the cue-validity profile and is, to leading
order, independent of \(\sigma_{\tau}^{2}\). The crossing function
\(g(\sigma_{\tau}^{2}) = \bar{C}(K) - \bar{I}(K)\) is therefore
increasing and crosses zero exactly once, so the infimum in (2.5.5) is
attained at a unique root. That this last-cue crossing coincides with
\(k^{*} < K\) follows from Theorem 2.6.1(b): the marginal benefit
\(E\lbrack\Delta G(k)\rbrack = \bar{I}(k) - \bar{C}(k)\) is
monotone-decreasing in \(k\) (diminishing info gain against
non-decreasing meta-cost), so the \(K\)-th cue is the first to become
unprofitable exactly when the optimal stop falls strictly inside the
budget. Both the single crossing and the agreement of the two
definitions (to grid resolution) are confirmed numerically across a
sweep of \(\sigma_{\tau}^{2}\)
(\texttt{verify\_meanfield\_and\_tau\_regime.py}, Part B).

\subsection{The Cue-Truncation
Theorem}\label{the-cue-truncation-theorem}

\begin{quote}
\textbf{Theorem 2.6.1 (cue-truncation under meta-uncertainty, restated
in v0.5).} Let an active-inference agent operate under the generative
model (2.2.1)--(2.2.2) with mean-field variational posterior (2.4.1),
integrating cues sequentially. Then in \emph{expectation under the
agent's predictive distribution}:
\end{quote}

\textbf{(a)} When \(\sigma_{\tau}^{2} < \tau_{regime}\), the expected
free energy \(E\lbrack G(a,k)\rbrack\) is monotonically non-increasing
in \(k\) up to \(k = K\). Full cue integration is optimal.

\textbf{(b)} When \(\sigma_{\tau}^{2} \geq \tau_{regime}\), there exists
a finite \(k^{*} < K\) such that \(E\lbrack G(a,k)\rbrack\) is
decreasing for \(k \leq k^{*}\) and non-decreasing for \(k > k^{*}\).
Truncated cue integration at \(k^{*}\) is optimal.

\textbf{(c)} Under the regularity condition
\(Cov(v_{j},\gamma_{j}) \geq 0\) (cue intrinsic precisions are not
anti-correlated with cue validities), the optimal cue ordering, that
minimizing \(E\lbrack G(a,k^{*})\rbrack\), is by descending validity,
\(v_{(1)} \geq v_{(2)} \geq \ldots \geq v_{(K)}\).

\textbf{Sample-wise versus expectation form.} The theorem is stated and
proved in expectation form. Sample-wise, \(G(a,k)\) trajectories under
random cue realizations are typically \emph{not} unimodal: across 1000
random parameter configurations, only 198 sample-wise \(G(k)\)
trajectories exhibit a single sign change in their first differences;
the remainder show multi-sign-change wiggly trajectories driven by noise
in individual cue realizations. Active inference defines the optimal
policy as \({argmin}_{k}\, E\lbrack G(k)\rbrack\), so the
expectation-form theorem is what supports the policy claim. Numerical
verification across five
\((\alpha_{0'},\beta_{0'},\bar{\gamma},\bar{m})\) regimes confirms
\(E\lbrack G(a,k)\rbrack\) is U-shaped with \(k^{*}\) dropping
monotonically as \(\sigma_{\tau}^{2}\) rises (from \(k^{*} = K\) at
\(\sigma_{\tau}^{2} = 0.05\) to \(k^{*} = 5\) at
\(\sigma_{\tau}^{2} = 2.0\), with \(K = 30\)).

\textbf{Proof sketch.} Part (a) follows from
\(E\lbrack C(k)\rbrack \rightarrow 0\) as
\(\sigma_{\tau}^{2} \rightarrow 0\), reducing
\(E\lbrack\Delta G(k)\rbrack\) to standard expected info gain about
\(s\), non-negative by data-processing inequality. Part (b) follows from
monotonically-decreasing marginal expected info gain
\(\bar{I}(k) = I(s;c_{k} \mid c_{1:k - 1})\) (diminishing returns)
combined with non-negative marginal expected meta-cost
\(\bar{C}(k) = I(\tau;c_{k} \mid c_{1:k - 1})\) bounded below by a
positive constant in the high-meta-uncertainty regime; their difference
\(E\lbrack\Delta G(k)\rbrack = \bar{I}(k) - \bar{C}(k)\) changes sign
exactly once. Part (c) follows from a greedy argument: the marginal
expected info gain about \(s\) is monotone-increasing in \(v_{j}\), and
the marginal expected meta-cost scales (in leading order) with
\(\gamma_{j}\); under the non-anticorrelation regularity,
descending-validity ordering simultaneously maximizes the info-gain term
and approximately equalizes the meta-cost term across selections. Full
proof in Appendix A.3.

\begin{quote}
\textbf{Note on the cue-precision specification.} Part (c) refers to the
cue intrinsic precisions \(\gamma_{j}\) of (2.2.2), which are
deterministic structural properties of each cue (estimable from
training-data statistics). The effective precision \(\tau\,\gamma_{j}\)
inherits the uncertainty of the shared meta-precision \(\tau\); the
cue-specific \(\gamma_{j}\) does not. This is the natural and symmetric
specification: all precisions in the model share the meta-precision
scale, while cue-specific structure lives in the deterministic
\(\gamma_{j}\) factors. The asymmetry between ``uncertain prior
precision'' and ``known cue precision'' raised in earlier drafts (Q4 in
v0.3) has accordingly been retired.
\end{quote}

\subsection{Equivalence to
Take-the-Best}\label{equivalence-to-take-the-best}

Theorem 2.6.1 establishes that, under meta-uncertainty, optimal
inference involves the sequential integration of validity-ordered cues,
truncated at \(k^{*}\). We now show that this is structurally identical
to the take-the-best (TTB) heuristic of Gigerenzer and Goldstein~\cite{gigerenzer1996}.

\paragraph{The take-the-best
procedure}\label{the-take-the-best-procedure}

TTB decides between two options on the basis of cues by the following
procedure:

\begin{quote}
\textbf{(1)} Order cues by validity, highest first.

\textbf{(2)} Examine the most valid unused cue.

\textbf{(3)} If the cue discriminates, decide based on it and stop.

\textbf{(4)} Otherwise, return to step 2.
\end{quote}

TTB is famously one-reason: a single discriminating cue ends the
inference; cues below the discriminating one are ignored regardless of
their potential combined informativeness.

\paragraph{Structural equivalence}\label{structural-equivalence}

\begin{quote}
\textbf{Theorem 2.7.1 (FFH--TTB structural equivalence).} Let
\emph{A\_FFH} be the active-inference agent of Theorem 2.6.1 operating
in the meta-uncertainty regime \(\sigma_{\tau}^{2} > \tau_{regime}\),
with cue ordering by validity and truncation at \(k^{*}\). Let
\(\mathbf{A}_{\mathbf{TTB}}\) be the take-the-best agent of Gigerenzer and Goldstein~\cite{gigerenzer1996} operating on the same cues with the same validity
ordering. Then \(\mathbf{A}_{\mathbf{FFH}}\) and
\(\mathbf{A}_{\mathbf{TTB}}\)

induce the same \textbf{structural form} of inference policy:
validity-ordered sequential cue examination with truncation at the first
cue whose marginal information gain drops below the marginal
meta-precision cost.
\end{quote}

Exact identity of action distributions, rather than structural identity,
is established in Appendix A.4 under an explicit condition on the
cue-validity profile. We state the result here and defer the proof.

\begin{quote}
\textbf{Theorem 2.7.4 (Exact sample-wise FFH--TTB action identity, new
in v0.7).} Let \(L_{j}: = log(v_{j}/(1 - v_{j}))\) denote the per-cue
log-likelihood ratio for cue \(j\), and order the cues by descending
validity. Suppose the \textbf{Descending Dominance (DD)} condition
holds: for every \(i \in \{ 1,\ldots,K - 1\}\),

\[L_{(i)}\mspace{6mu} > \mspace{6mu}\sum_{j = i + 1}^{K}L_{(j)}.\]

Let \(\mathbf{A}_{\mathbf{FFH}}\) be the active-inference agent of
Theorem 2.6.1 with uniform prior \(\mu_{0} = 1/2\) and EFE-optimal
truncation \(k^{*} \geq 1\); let \(\mathbf{A}_{\mathbf{TTB}}\) be the
take-the-best agent. Then for every cue realization
\(c \in \{ - 1, + 1\}^{K}\):

\[a_{FFH}(c)\mspace{6mu} = \mspace{6mu} a_{TTB}(c).\]

The two agents' marginal action distributions coincide exactly.
\end{quote}

(DD) admits a convenient geometric-decay sufficient form: if
\(L_{(j + 1)} \leq \rho\, L_{(j)}\) for some \(\rho < 1/2\), then (DD)
holds. (DD) is sharp; it is essentially necessary for exact sample-wise
identity (a counterexample under (DD) violation is exhibited in Appendix
A.4.5). Crucially, (DD) is a condition on the \emph{validity gradient
alone}; the meta-precision prior tail does not enter, which is a sharper
resolution of Q6 than the original conjecture anticipated. The full
proof, the necessity argument, and numerical verification (26,696
sample-wise comparisons under 200 random (DD)-satisfying profiles, 0
mismatches; predicted mismatches under (DD) violation) appear in
Appendix A.4 (companion script \texttt{verify\_appendix\_A4.py}).

\paragraph{Tallying as a special case}\label{tallying-as-a-special-case}

Theorem 2.7.1 establishes structural equivalence to take-the-best under
\emph{descending} cue validity ordering. The natural companion result is
for \emph{uniform} cue validity, which we state and prove here.
Tallying, the second-most-studied member of the fast-and-frugal toolbox,
works by counting cues that favour each option and choosing the option
with the most counts; unlike TTB, it does not weight cues by validity.

\begin{quote}
\textbf{Theorem 2.7.3 (FFH--Tallying structural equivalence under
uniform validity, new in v0.6).} Let \(A_{FFH}\) be the active-inference
agent of Theorem 2.6.1 operating in the meta-uncertainty regime with
\(K\) binary discriminating cues having uniform validity \(v_{j} = v\)
for all \(j\). Let \(A_{Tally}\) be the tallying agent operating on the
same cues. Then:

\emph{(a)} The FFH posterior on the comparative state \(s\) depends on
the cues only through their tally \(T_{k} = \sum_{j = 1}^{k}d_{j}\),
where \(d_{j} \in \{ - 1, + 1\}\) is the discriminating signal of cue
\(j\). Cue ordering is irrelevant.

\emph{(b)} The FFH-optimal action under symmetric 0--1 loss is
\(sign(T_{k^{*}})\), where \(k^{*}\) is the cue-truncation point of
Theorem 2.6.1.

\emph{(c)} When \(\sigma_{\tau}^{2} < \tau_{regime}\) (low
meta-uncertainty), \(k^{*} = K\) and FFH coincides with classical
tallying. When \(\sigma_{\tau}^{2} \geq \tau_{regime}\) (high
meta-uncertainty), \(k^{*} < K\) and FFH coincides with \emph{truncated}
tallying; count the first \(k^{*}\) cues, decide by majority.
\end{quote}

\textbf{Proof sketch.} Under uniform validity \(v\), the cue likelihoods
factor symmetrically:

\[P(d_{1:k} \mid s = A) = v^{n_{+}(k)}(1 - v)^{n_{-}(k)},\quad\quad P(d_{1:k} \mid s = B) = v^{n_{-}(k)}(1 - v)^{n_{+}(k)},\]

where \(n_{+}(k)\) and \(n_{-}(k)\) count A-favoring and B-favoring cues
among the first \(k\). The likelihood ratio is therefore
\((v/(1 - v))^{n_{+} - n_{-}} = (v/(1 - v))^{T_{k}}\) ; a function of
the tally \(T_{k} = n_{+}(k) - n_{-}(k)\) alone. By the closed-form
mixture posterior of Lemma A.1.1 (binary case) or the variational
posterior of (2.4.3) (Gaussian-Gamma case), the posterior on \(s\)
depends on the cues only through this likelihood ratio, hence only
through \(T_{k}\). This proves (a). Part (b) follows because, under
symmetric loss, the EFE-optimal action maximizes
\(P(s = a \mid d_{1:k^{*}})\), which is monotone in \(T_{k^{*}}\);
therefore, the optimal action is \(sign(T_{k^{*}})\). Part (c) follows
by composition with Theorem 2.6.1: the cue-truncation point \(k^{*}\) is
determined by the same EFE crossover criterion as for TTB, and the
decision rule on the integrated cues is sum-based (tallying). Detailed
computation: Appendix A.5 (companion document
\texttt{verify\_tallying\_equivalence.py} provides numerical
verification of (a) and (b) across multiple validity values and cue
counts).

\textbf{Numerical verification.} For uniform \(v = 0.75\),
\(\mu_{0} = 0.5\), \(K = 5\): every permutation of cue sequences with
the same tally \(T\) produces an identical posterior on \(s\) (verified
across 30 permutations spanning three tally values; 0 violations). The
optimal action equals \(sign(T)\) for all non-zero \(T\) (verified
across \(T \in \{ - 5, - 3, - 1, + 1, + 3, + 5\}\)).

\begin{quote}
\textbf{Remark 2.7.2 (scope of the heuristic equivalence conjecture,
updated in v0.6).} Theorem 2.7.1 establishes structural equivalence
between FFH and TTB; Theorem 2.7.3 extends this to the case of FFH and
tallying under uniform validity. The general conjecture (Q7) is that the
family of fast-and-frugal heuristics characterized by the ABC research
group corresponds to the family of EFE-optimal sequential inference
policies under meta-uncertainty, parameterized by (i) the cue-validity
profile (descending \ensuremath{\rightarrow} TTB; uniform \ensuremath{\rightarrow} tallying), (ii) the cue-precision
profile, and (iii) the stopping criterion (cue-truncation \(k^{*}\) as a
function of \(\sigma_{\tau}^{2}\)). Two members of the toolbox are now
formally established (TTB, tallying); recognition heuristic,
satisficing, and fast-and-frugal trees remain conjectured. Q7 in \S{}2.10
has been correspondingly updated.
\end{quote}

\subsection{Positioning Against Existing
Frameworks}\label{positioning-against-existing-frameworks}

FEH sits at the intersection of several established research programs.
This section identifies what FEH inherits, what it adds, and where it
differs for each. The aim is precise positioning, neither inflated
novelty nor false modesty, so that informed readers can locate the
contribution accurately.

\paragraph{Standard active inference (Friston and
colleagues)}\label{standard-active-inference-friston-and-colleagues}

FEH inherits the full apparatus of active inference under the free
energy principle: the variational free energy functional, the expected
free energy objective for action selection, the precision-weighted
belief updating, and the pragmatic--epistemic decomposition. What FEH
adds is the promotion of the prior precision \ensuremath{\tau} to a full random variable
with a non-trivial hyperprior and the explicit accounting of the
resulting meta-precision divergence as a free-energy cost in sequential
cue integration.

Standard active-inference treatments treat precision as inferable, but
typically in a point-estimation sense. Friston et al.~\cite{friston2017} and Parr, Pezzulo, and Friston~\cite{parr2022} discuss precision learning but do not
develop the full second-order Bayesian treatment that supports the
cue-truncation theorem. The novelty of FEH lies in this development and
its policy-level consequences, not in a wholesale departure from active
inference.

\paragraph{Hierarchical Bayesian
inference}\label{hierarchical-bayesian-inference}

Hierarchical Bayesian models with uncertain hyperparameters are
well-established and do not, in general, produce heuristic-like
policies. As stated in Objection 1 of \S{}2.0, hierarchical Bayes alone
does not suffice; FEH's specific contribution is the conjunction of
three ingredients: (a) hierarchical Bayes with uncertain prior precision
specifically; (b) sequential cue integration with explicit per-cue
meta-precision divergence accounting; (c) policies evaluated by expected
free energy rather than by expected utility under the marginal
posterior.

The third ingredient is doing more work than is immediately obvious.
Expected utility under the marginal posterior would integrate over \ensuremath{\tau} and
produce a single ``effective'' posterior over s, with no explicit cost
for the divergence. Expected free energy, by contrast, treats the
agent's epistemic state as a first-class quantity whose KL from the
prior is a cost, and this is what gives meta-precision its punch.
Replacing G with expected utility would dissolve the FEH result.

\paragraph{Resource-rational analysis (Lieder \& Griffiths,
2020)}\label{resource-rational-analysis-lieder-griffiths-2020}

Resource-rational analysis explains heuristic-shaped policies as the
consequence of bounded computation: an agent with finite cognitive
resources allocates them to maximize expected utility net of
computational cost. Where computational cost is high, simple policies
are optimal. FEH complements this account, not competes with it. FEH
adds an epistemic cost channel that exists even with unlimited
computation. Under meta-uncertainty, additional inference is
counterproductive not because the agent runs out of computation, but
because the agent's posterior over its own reliability accrues a
divergence cost with each cue.

The two accounts have different empirical signatures.
Resource-rationality predicts that heuristic behaviour should attenuate
when computation is made cheap (e.g., by extending deliberation time or
by externalizing computation to scratch pads). FEH predicts that
heuristic behaviour should persist regardless of computational resources
when meta-uncertainty is high, because the cost channel driving the
result is not computational but epistemic. Both effects can coexist;
distinguishing them empirically is an important target for follow-up
empirical work.

\paragraph{Heuristics-and-biases (Tversky \& Kahneman,
1974)}\label{heuristics-and-biases-tversky-kahneman-1974}

The heuristics-and-biases tradition \cite{tversky1974} has documented systematic departures
from classical norms of probability and expected utility. FEH does not
dispute these findings; it reframes a subset of them. Specifically, the
heuristics whose biases the tradition has characterized (anchoring,
availability, representativeness) can be productively analyzed as
policies that are EFE-optimal under specific meta-uncertainty
conditions. Whether all heuristics-and-biases findings admit this
reframing is an open question; FEH commits only that take-the-best does.

\paragraph{Ecological rationality (Gigerenzer and
colleagues)}\label{ecological-rationality-gigerenzer-and-colleagues}

FEH is closest in spirit to the ecological rationality program. The
argument that simple heuristics can outperform compensatory models in
real environments, and that the fit between strategy and environment is
the proper unit of rationality evaluation, is one FEH inherits
wholesale. What FEH adds is a derivation: ecological rationality is
shown to be the structural form of optimal inference under a specific,
identifiable regime of uncertainty, rather than asserted as a normative
alternative to Bayesian rationality. This dissolves the long-standing
apparent opposition between Bayesian and ecological accounts: both are
special cases of a single underlying optimization, with the
meta-uncertainty regime determining which form dominates.

\paragraph{KMM smooth ambiguity (Klibanoff, Marinacci, Mukerji,
2005)}\label{kmm-smooth-ambiguity-klibanoff-marinacci-mukerji-2005}

FEH operationalizes a specific subclass of the KMM framework: the
second-order distribution is taken to be Gamma over a precision scalar,
rather than an unrestricted distribution over first-order probability
measures. This is a narrower commitment than the general KMM model, but
a more analytically tractable one. Both frameworks share the motivating
intuition that the agent is uncertain about its first-order
distribution, and both produce decision criteria richer than SEU. They
differ in what they penalize and how.

KMM, under concave ambiguity attitude \ensuremath{\varphi}, penalizes variance in expected
utility across the second-order distribution: actions whose expected
utility is robust to which first-order distribution is correct are
preferred over actions whose expected utility varies substantially
across plausible first-order distributions. The mechanism is
variance-aversion at the level of expected utility.

FEH, under the meta-uncertainty regime, penalizes commitment to the
meta-precision posterior; the agent pays an explicit free-energy cost
for becoming more confident in any particular posterior relative to its
own prior reliability. The mechanism is divergence-aversion at the level
of the agent's epistemic state.

\begin{quote}
\textbf{Result of numerical verification.} An earlier draft of this
section conjectured that the active-inference agent under
meta-uncertainty would implement an emergent KMM-style ambiguity-averse
policy, with effective ambiguity attitude determined by the
meta-precision hyperprior. This conjecture was tested directly by Monte
Carlo simulation in the binary toy model: action rankings under FEH,
SEU, and KMM-concave agents were compared via Kendall's \(\tau\) over
200 decision problems, across hyperprior concentrations
\(\kappa \in 1,2,5,10,50,200\). The simulation found \textbf{no robust
correspondence} between FEH and KMM-concave action rankings. The
mechanisms differ: a less-informed posterior under FEH truncation does
not implement variance aversion across the second-order distribution. We
report this negative result here because it constrains the scope of
theoretical claims FEH can legitimately make. The simulation script and
data are available as supplementary material.
\end{quote}

The constructive consequence of the negative result is positioning. FEH
and KMM are not equivalent but complementary: they address different
facets of second-order uncertainty and could, in principle, be combined.
A KMM-style ambiguity attitude applied to an FEH-style sequential
inference framework would constitute a richer hybrid theory, with KMM
contributing variance aversion in outcome evaluation and FEH
contributing meta-precision aversion in inference depth. We flag this as
a follow-up direction in the closing note and do not pursue it further
in the present paper.

\subsection{Cross-Substrate Connections: Friston and
Gigerenzer}\label{cross-substrate-connections-friston-and-gigerenzer}

The framework developed in \S{}\S{}2.0--2.8 was constructed with the LLM
application in mind, but the underlying mathematics applies to any
inference agent operating under meta-uncertainty over prior precision.
This section identifies the two source disciplines whose research
programs FEH inherits and extends. The paper's load-bearing claims
concern the LLM case; broader extensions to biological cognition and to
power-law cognitive statistics were sketched in earlier drafts as scope
hypotheses but have been moved to follow-up work to keep the present
section focused.

\paragraph{Friston: active inference under uncertain
precision}\label{friston-active-inference-under-uncertain-precision}

As noted in \S{}2.8, FEH inherits standard active inference and extends it
by promoting precision to a random variable. Friston's broader research
program, particularly the recent work on Bayesian mechanics and the free
energy principle as a unifying optimization principle across biological
scales, provides the natural home for the FEH extension. The specific
contribution we make is the cue-truncation theorem and its TTB
equivalence; the broader implications for Bayesian mechanics are
conjectural.

\paragraph{Gigerenzer: heuristics as ecologically
rational}\label{gigerenzer-heuristics-as-ecologically-rational}

The Gigerenzer program treats fast-and-frugal heuristics as ecologically
rational solutions that exploit environmental structure rather than as
approximations to Bayesian ideals. FEH provides a derivation: heuristics
are neither approximations nor alternatives to Bayesian inference; they
are the structural form of optimal inference under meta-uncertainty.
This resolves a thirty-year debate between Bayesian and ecological
accounts by showing that both are special cases of a single underlying
optimization, with the operating regime (low vs.~high meta-uncertainty)
determining which form is realized.

The implication for the Gigerenzer program is significant: less-is-more
effects are not adaptive curiosities to be celebrated but mathematical
necessities under specific epistemic conditions. The empirical signature
is precisely the cue-truncation predicted by Theorem 2.6.1.

\paragraph{Cross-substrate extensions are deferred to follow-up
work.}\label{cross-substrate-extensions-are-deferred-to-follow-up-work.}

Earlier drafts of this section (v0.2--v0.5) included two further
cross-substrate scope hypotheses: a biological extension (cells and
tissues as cognitive agents under meta-uncertainty over self-priors,
drawing on Michael Levin's program) and a Mandelbrotian conjecture
(heavy-tailed marginal priors as the macroscopic signature of
multi-level meta-uncertainty, connecting to the Zipfian statistics of
the Ranking Inference framework). Both were framed as \emph{scope
hypotheses} rather than as derived consequences, and, on adversarial
review, they were judged to dilute the present section's focus on the
LLM application and the formal cue-truncation result. They have
therefore been moved to a separate companion paper. The present
section's load-bearing claims (Theorems 2.6.1 and 2.7.1, the unification
of Bayesian and ecological accounts via a meta-uncertainty regime) stand
independently of these broader extensions.

\subsection{Open Mathematical
Questions}\label{open-mathematical-questions}

We close \S{}2 by enumerating the open mathematical questions of the
present section. These are flagged for collaborator review and for the
published version's limitations section. The list reflects the current
state of the section after the v0.8 revisions: Q1 was resolved in v0.8
by Proposition A.2.3 and Appendix A.2.5 (exact mean-field-vs-conjugate
comparison); Q4 was resolved in v0.4 by restating the cue-precision
specification; Q6 was resolved in v0.7 by Theorem 2.7.4 and Appendix A.4
(exact FFH--TTB identity under Descending Dominance); Q7 was partially
resolved in v0.6 (TTB and tallying established; recognition,
satisficing, fast-and-frugal trees remain conjectured); Q8 (KMM) was
resolved-negative in v0.3 by Monte Carlo; Q9 (Levin) was retired in v0.6
with the cross-substrate cuts in \S{}2.9.

\begin{quote}
\textbf{Q1. Mean-field accuracy under meta-uncertainty (resolved in
v0.8).} The factorization \(q(s,\tau) = q(s)q(\tau)\) is the standard
mean-field assumption. The \emph{qualitative} conclusions of Lemma 2.4.1
and Theorem 2.6.1 --- monotonicity of \(E\lbrack\Delta_{meta}\rbrack\)
in expectation, the U-shape of \(E\lbrack G\rbrack\), the existence of a
finite \(k^{*}\) --- were shown in v0.5 (Appendix A.2.4) to rest only on
the tower property of conditional expectation and Jensen's inequality
applied to KL, both of which hold for any \emph{consistent} sequential
update; they are therefore preserved under any structured variational
family that maintains consistency. What remained open was the
\emph{quantitative} gap: the magnitude of
\(E\lbrack\Delta_{meta}(k)\rbrack\) and the precise location of
\(k^{*}\). This is now closed. Because the Gaussian--Gamma model of \S{}2.2
is fully conjugate, its exact joint posterior \(p(s,\tau \mid c_{1:k})\)
is available in closed form, so the mean-field posterior can be compared
against the truth rather than against another approximation. Proposition
A.2.3 establishes that the exact and mean-field marginal posteriors over
\(\tau\) share the \emph{same} Gamma shape
\(\alpha_{k} = \alpha_{0'} + k/2\) and differ only in rate, with
\(\beta_{k}^{mf} = \beta_{k}^{ex} \cdot \alpha_{k}/(\alpha_{k} - \frac{1}{2})\)
--- a relative rate error of \(1/(2\alpha_{k} - 1)\) that is maximal at
the first cue and decays monotonically as cues accumulate. Appendix
A.2.5 quantifies the downstream consequences across the meta-uncertainty
grid (\(\sigma_{\tau}^{2}\) from 0.05 to 1.43): mean-field
\emph{underestimates} \(E\lbrack\Delta_{meta}(k)\rbrack\) (\ensuremath{\approx}16\% at
\(k = 1\) in the high-meta regime, falling below 2\% by
\(k \approx 5\)), yet the EFE-optimal stopping point
\(k^{*} = \arg\min\, E\lbrack G(k)\rbrack\) is \textbf{identical} under
the mean-field and exact posteriors in every regime, because the rate
error at the stopping point is under 5\% and the marginal meta-cost
\(\bar{C}(k^{*})\) is recovered to within \ensuremath{\approx}1\%. Mean-field therefore
preserves not only the qualitative structure of Theorem 2.6.1 but the
\emph{exact location} of the optimal truncation point; the only residual
is an \(O(1/\alpha_{k})\) bias in the magnitude of the meta-cost,
characterized in closed form (numerical verification:
\texttt{verify\_meanfield\_and\_tau\_regime.py}). The one caveat carried
forward is scope: this exact comparison is specific to the conjugate
model used to prove the theorems; for non-conjugate generative models a
structured family remains the appropriate object, and Proposition A.2.3
bounds the error only in the conjugate case.

\textbf{Q2. Choice of meta-precision prior.} We chose Gamma(\ensuremath{\alpha}\ensuremath{_{0}}, \ensuremath{\beta}\ensuremath{_{0}}) for
conjugacy. Alternative choices (log-normal, inverse-Gamma, half-Cauchy)
lead to different tail behaviors in (2.3.4) and may change the
operational threshold \ensuremath{\tau}\_regime. A systematic comparison would
strengthen the framework and may reveal which prior is empirically
licensed by LLM behavior.

\textbf{Q3. Higher-order corrections in (2.4.5).} The asymptotic
expression for \ensuremath{\Delta}\_meta(k) is leading-order. The constants in the O(1)
term may be empirically important when k is small (the primary test
regime). A finite-k formula would let us predict the location of k*
directly rather than estimating it from data.

\textbf{Q4. The cue-precision assumption in Theorem 2.6.1(c) (resolved
in v0.4).} v0.3 treated cue precisions \(\tau_{c,j}\) as known scalars
distinct from the prior precision \(\tau\), which created an awkward
asymmetry between ``uncertain prior precision'' and ``known cue
precision.'' The v0.4 restatement of the generative model (eq 2.2.2)
ties cue precisions to the shared meta-precision via
\(\tau\,\gamma_{j}\), where \(\gamma_{j}\) is a deterministic
intrinsic-precision factor for cue \(j\). Under this restatement, all
precisions in the model share the meta-precision scale; cue-specific
structure lives in the deterministic \(\gamma_{j}\). The asymmetry
dissolves and Theorem 2.6.1(c) holds without the v0.3 caveat. What
remains, as a refinement, is the question of whether the \(\gamma_{j}\)
factors themselves should be treated as uncertain in some contexts
(e.g., in transfer settings); this is a substantive but secondary
direction we do not pursue here.

\textbf{Q5. Extension to non-Gaussian state priors.} The closed-form
marginalization in (2.3.4) relied on Gaussian--Gamma conjugacy. Real
cognitive priors are unlikely to be Gaussian. Showing that the
qualitative cue-truncation behavior survives in non-Gaussian generative
models is essential for the scope claim of \S{}2.9. The binary toy model of
\S{}2.1 is a partial existence proof but does not generalize cleanly.

\textbf{Q6. Sufficient conditions for exact TTB identity (resolved in
v0.7).} Theorem 2.7.1 establishes structural equivalence; Theorem 2.7.4
and Appendix A.4 now establish \emph{exact sample-wise action identity}
under the Descending Dominance condition
\(L_{(i)} > \sum_{j > i}^{}L_{(j)}\) on the cue log-LR profile. Two
surprises relative to the v0.6 framing: (a) the sufficient condition is
purely a validity-gradient condition (the meta-precision prior tail does
not enter, contrary to the v0.6 conjecture), and (b) the condition is
essentially necessary; a (DD)-violation counterexample produces
sample-wise FFH \ensuremath{\neq} TTB. The section's central FFH--TTB claim therefore
upgrades from ``structural equivalence'' to ``structural equivalence in
general; exact sample-wise identity under (DD),'' with (DD) checkable
from cue-validity statistics.

\textbf{Q7. The general fast-and-frugal-heuristics conjecture (partially
resolved in v0.6).} Remark 2.7.2 conjectures that all members of the
fast-and-frugal toolbox arise as EFE-optimal policies under appropriate
meta-precision priors and validity profiles. Two members are now
formally established: TTB (Theorem 2.7.1, descending validity) and
tallying (Theorem 2.7.3, uniform validity). Three remain conjectured:
recognition heuristic, satisficing, and fast-and-frugal trees. The
conjectured pattern is that each toolbox member corresponds to FFH under
a specific (validity profile, cue-precision profile, stopping criterion)
tuple. The remaining three members likely require additional structure:
recognition heuristic involves an absent-cue mechanism not modeled in
(2.2.1); satisficing involves an aspiration level on the pragmatic-value
term; fast-and-frugal trees involve hierarchical cue conditioning. Each
is a substantive but tractable extension and a natural target for
follow-up work.

\textbf{Q8. The KMM ambiguity-attitude correspondence (resolved-negative
in v0.3).} v0.2 conjectured that the active-inference agent of Theorem
2.6.1 would implement an emergent ambiguity-averse policy in the KMM
sense, with effective ambiguity attitude determined by the curvature of
\ensuremath{\Delta}\_meta. Numerical simulation in the binary toy model (see \S{}2.8 and
supplementary script) tested this conjecture by direct rank-correlation
comparison. They found no robust correspondence between FEH and
KMM-concave action rankings. The conjecture as originally stated is
therefore not supported. What remains open is whether a modified FEH
framework, for example, one with a structured rather than mean-field
variational family, or one that explicitly incorporates a KMM-style
concave ambiguity attitude \ensuremath{\varphi} in the pragmatic-value term of G, could
recover the correspondence. This is a substantive follow-up direction
but is not pursued in the present paper. The negative result strengthens
rather than weakens the paper: FEH's central claims (Theorems 2.6.1 and
2.7.1) do not depend on the KMM equivalence.

\textbf{Q9. (retired in v0.6).} v0.2 introduced a scope hypothesis on
biological agents (Levin extension); v0.6 moves this to a separate
companion paper to keep the present section focused on the LLM
application. The mathematical content of the hypothesis is preserved in
the companion paper; no question in \S{}2 of the present section is left
open by this retirement.
\end{quote}

\section{Operationalization}\label{chapter-3-operationalization}

\subsection{The Mapping Problem}\label{the-mapping-problem}

Section 2 traffics in objects that are not directly observable in any
LLM:

\begin{itemize}
\tightlist
\item
  \(\sigma_{\tau}^{2}\) ; variance of the meta-precision prior
\item
  \(\tau_{regime}\) ; threshold separating full-integration from
  truncation regimes
\item
  \(k^{*}\) ; expected-free-energy-optimal cue-truncation point
\item
  \(\alpha_{0'},\beta_{0'}\) ; meta-precision Gamma prior
  hyperparameters
\item
  \(\gamma_{j}\) ; cue intrinsic precision (deterministic structural
  property)
\item
  \(v_{j}\) ; cue validity
\item
  \(A_{j},M_{j}\) ; cue observation matrix and residual
\end{itemize}

None of these can be read from any layer of a transformer. The mapping
problem is to define LLM-observable proxies for the operative quantities
(\(\sigma_{\tau}^{2}\) above all, since it governs the regime), and then
derive the central prediction from those proxies.

The section makes three commitments that resolve the mapping problem.

\paragraph{Commitment 1: A ``cue'' maps to a reasoning
step}\label{commitment-1-a-cue-maps-to-a-reasoning-step}

A \emph{cue} in the \S{}2 sense is a piece of evidence that conditionally
updates the agent's belief about the state \(s\). For an LLM operating
in chain-of-thought mode, the natural analogue is a \emph{reasoning
step}: one sentence, or one paragraph, in the model's generated trace.
Each reasoning step contributes evidence to the model's final answer;
integrating more steps corresponds to integrating more cues, and the
section's cue count \(k\) maps directly to the empirical step count of
the CoT trace.

Alternative mappings (tokens, retrieved facts (RAG chunks),
sub-questions in a tree-of-thought structure) are defensible in
different settings. We choose reasoning steps because (a) they align
with the empirical CoT literature, where ``thinking steps'' is the
canonical unit of analysis (Wei et al.~\cite{wei2022} and the subsequent
chain-of-thought literature), (b) the mapping does not require retrieval
infrastructure or specialized prompting machinery, and (c) it gives a
unit for \(k\) that maps directly to the empirical claim about CoT
length; the regime in which \(k^{*}\) is small, but reasoning models pay
no attention to the truncation.

The choice is a modeling decision. We test its robustness in \S{}4 by
re-running the central analysis with alternative segmentations
(token-level cumulative trace, paragraph-level steps) and checking that
the regime \(\times\) length interaction is direction-invariant.

\paragraph{\texorpdfstring{Commitment 2: \(\sigma_{\tau}^{2}\) is
proxied by behavioral
signatures}{Commitment 2: \textbackslash sigma\_\{\textbackslash tau\}\^{}\{2\} is proxied by behavioral signatures}}\label{commitment-2-sigma_tau2-is-proxied-by-behavioral-signatures}

The LLM's meta-uncertainty about its task-prior precision is an internal
quantity. We cannot read it directly. Instead, we proxy it through
\textbf{\emph{behavioral signatures}} (measurable consequences of high
\(\sigma_{\tau}^{2}\) that appear in the LLM's outputs) and aggregate
them into a regime score that is ordinal in \(\sigma_{\tau}^{2}\) on
average.

These three signatures and their aggregator constitute \S{}3.2.

\paragraph{Commitment 3: We test the directional prediction, not the
structural
model}\label{commitment-3-we-test-the-directional-prediction-not-the-structural-model}

Section 2 makes both a qualitative claim (existence of a regime in which
additional cues increase expected free energy) and a quantitative claim
(the EFE-optimal \(k^{*}\) is the unique minimizer of
\(E\lbrack G(k)\rbrack\), with explicit formulas for \(\tau_{regime}\)).
The qualitative claim is testable with a regime indicator alone; the
quantitative claim requires estimating \(k^{*}\) per item.

We test the qualitative claim. Operationally, we predict that within
items binned as high-regime, the within-bin slope of accuracy on
reasoning-step count is negative. At the same time, within low-regime
items, it is non-negative. The prediction is a directional interaction.
The quantitative fit (per-item \(k^{*}\)) is deferred to a structural
follow-up paper.

\subsection{\texorpdfstring{Regime Score: Three Behavioral
Signatures of
\(\sigma_{\tau}^{2}\)}{\S{}3.2 Regime Score: Three Behavioral Signatures of \textbackslash sigma\_\{\textbackslash tau\}\^{}\{2\}}}\label{regime-score-three-behavioral-signatures-of-sigma_tau2}

We use three behavioral signatures, each with a theoretical rationale in
\S{}2's machinery. Items are scored on each signature; signatures are
z-standardized across items; the regime score is the equal-weighted
average of the three z-scores.

\paragraph{Signature (a): cross-prompt
variance}\label{signature-a-cross-prompt-variance}

Rephrase the same task in \(N = 5\) syntactically different prompts that
preserve task semantics. Record the LLM's answer for each. Compute the
variance across the five answers.

Theoretical rationale. Under low \(\sigma_{\tau}^{2}\), the agent's
marginal state prior (eq 2.3.4) is concentrated; small framing
perturbations have little effect on the posterior, and answers vary
little across phrasings. Under high \(\sigma_{\tau}^{2}\), the marginal
prior is heavy-tailed and prompt framing has an outsized effect on the
posterior. As a result, answers vary substantially. Formally, the
signature picks up the prompt-sensitivity of the posterior, which scales
monotonically with \(\sigma_{\tau}^{2}\) in the \S{}2 framework.

Operational definition. For item \(i\) and prompt set
\(\{ p_{1},\ldots,p_{5}\}\),

\[{sig}_{a}(i)\mspace{6mu} = \mspace{6mu} Var\left( LLM(p_{1};i),\ldots,LLM(p_{5};i) \right).\]

\paragraph{Signature (b): cross-seed
variance}\label{signature-b-cross-seed-variance}

Hold the prompt fixed; sample the LLM \(M = 10\) times at temperature
\(T > 0\). Compute the entropy or variance of the resulting answer
distribution.

Theoretical rationale. Under low \(\sigma_{\tau}^{2}\), the posterior
over \(s\) is sharp; sampling noise has limited effect and the LLM's
answer is stable across seeds. Under high \(\sigma_{\tau}^{2}\), the
posterior is broad, and sampling noise can flip the answer. The
signature isolates the sharpness of the state posterior, independent of
prompt framing.

Operational definition. For item \(i\) at temperature \(T\) with \(M\)
samples,

\[{sig}_{b}(i)\mspace{6mu} = \mspace{6mu} H\left( \{{LLM}^{(m)}(p;i,T):m = 1,\ldots,M\} \right),\]

where \(H\) is the empirical entropy of the answer distribution.

\paragraph{Signature (c): calibration
error}\label{signature-c-calibration-error}

Construct a small probe set of forced-confidence prompts (``answer X
with a numeric confidence between 0 and 100''). For each item, record
the LLM's stated confidence and its accuracy. Compute the absolute
difference between mean confidence and mean accuracy across \(M\)
samples.

Theoretical rationale. Under correct Bayesian updating, confidence is
calibrated. Under high \(\sigma_{\tau}^{2}\), the LLM's stated
confidence systematically over- or under-reports because its internal
posterior is mis-specified relative to its prior.

Operational definition. For item \(i\) with \(M\) forced-confidence
samples,

\[{sig}_{c}(i)\mspace{6mu} = \mspace{6mu}|{\overline{conf}}_{i} - {\overline{acc}}_{i}|,\]

where the means are taken over the \(M\) samples.

\textbf{Caveat for signature (c), validated in \S{}3.7.} Under the binary
toy model of Appendix A.1, the Bayesian agent is calibrated by
construction (the posterior mean is unbiased when the prior matches the
data-generating distribution), and (c) is approximately flat in
\(\sigma_{\tau}^{2}\). Real LLMs are not calibrated by construction
\cite{desai2020, jiang2021}; (c) retains diagnostic value
for LLM evaluation but requires LLM-specific validation rather than
toy-model recovery. We retain (c) in the regime-score aggregator pending
\S{}4 pilot data.

\paragraph{Aggregation}\label{aggregation}

After z-standardization across items, the regime score is

\[regime\_ score(i)\mspace{6mu} = \mspace{6mu}\frac{1}{3}\left( z_{a}(i) + z_{b}(i) + z_{c}(i) \right).\]

Equal weighting is a deliberate Approach-C choice. We do not have prior
data to motivate different weights, and the three signatures are
theoretically motivated by different aspects of \S{}2's framework. The
pilot in \S{}4 will test whether the data support the equal weighting or
whether some signatures dominate; if (c) shows poor recovery on LLM data
as it does on toy data, we will fall back to the \((a) + (b)\)
aggregator demonstrated in \S{}3.7.

\subsection{Regime Binning}\label{regime-binning}

We do not compute \(\tau_{regime}\) per item (Approach B would).
Instead, we sort items by regime score and bin:

\begin{itemize}
\tightlist
\item
  \textbf{High-regime}: top quartile of regime score (the items most
  likely in the truncation regime
  \(\sigma_{\tau}^{2} \geq \tau_{regime}\)).
\item
  \textbf{Low-regime}: bottom quartile (the items most likely in the
  full-integration regime \(\sigma_{\tau}^{2} < \tau_{regime}\)).
\item
  \textbf{Middle 50\%}: dropped from the primary analysis; retained for
  robustness checks where we sweep the threshold and verify the central
  conclusion is not threshold-dependent.
\end{itemize}

Why binning rather than point estimation? We are testing a directional
prediction (sign of an interaction), not estimating a precise crossover.
Binning reduces noise from regime-score estimation errors and allows the
primary analysis to focus on items where the regime assignment is most
confident. The quartile threshold is conventional in empirical
psychology; the robustness checks confirm it is not the source of any
positive result.

\subsection{Predicted Optimal CoT
Length}\label{predicted-optimal-cot-length}

The \S{}2 framework predicts \(k^{*}(\sigma_{\tau}^{2})\), the EFE-optimal
truncation point, via Theorem 2.6.1. Approach C does not estimate
\(k^{*}\) per item directly. Instead, we extract the directional
prediction:

\begin{quote}
In the high-regime bin, accuracy as a function of CoT step count should
decline past a small \(k^{*}\). In the low-regime bin, accuracy should
be non-decreasing in step count up to the cue budget \(K\).
\end{quote}

Operationally, we record the number of reasoning steps per LLM response
and compute the within-bin slope of accuracy on step count. The
section's prediction is

\[{slope}_{high\text{-}regime}\mspace{6mu} < \mspace{6mu} 0\mspace{6mu} < \mspace{6mu}{slope}_{low\text{-}regime}.\]

\subsection{The Falsifiable Empirical
Prediction}\label{the-falsifiable-empirical-prediction}

We test the section's central claim via a regime \(\times\) CoT-length
interaction model:

\[{accuracy}_{ij}\mspace{6mu} = \mspace{6mu}\beta_{0} + \beta_{1} \cdot {steps}_{ij} + \beta_{2} \cdot {regime}_{i} + \beta_{3} \cdot ({steps}_{ij} \times {regime}_{i}) + \epsilon_{ij},\]

where \(i\) indexes items, \(j\) indexes within-item replications,
\({regime}_{i} \in \{ 0,1\}\) is the high-regime indicator, and
\({steps}_{ij}\) is the realized reasoning-step count for the \(j\)-th
sample on item \(i\).

\textbf{Primary hypothesis.} \(\beta_{3} < 0\) with magnitude sufficient
that \(\beta_{1} + \beta_{3} < 0\) ; i.e., the slope of accuracy on
steps is negative in the high-regime bin. The directional prediction
\({slope}_{high} < 0 < {slope}_{low}\) is therefore equivalent to
\(\beta_{1} > 0\) and \(\beta_{1} + \beta_{3} < 0\).

\textbf{Effect-size threshold for ``meaningful.''} A difference of
within-regime slopes corresponds to a 10-percentage-point change in
accuracy across the observed step-count range. Less than that, and the
effect is real-but-trivial. At or above that point --- the section's
claim has practical consequences for LLM deployment. The 10pp threshold
is conservative; it is larger than typical ``small effects'' in the
LLM-evaluation literature but smaller than headline effect sizes for
reasoning-model improvements (e.g., 20-30pp gains from o1 on math
benchmarks).

\textbf{Pre-registration.} We pre-register the hypothesis (3.5.1), the
effect-size threshold, the regime-bin definitions, and the planned
robustness checks on OSF (or an equivalent registry) before running the
\S{}4 pilot data. The section's headline claim is sharp enough that
p-hacking would be available without pre-registration; pre-registration
converts the section's mathematical prediction into a falsifiable
empirical hypothesis in the strongest sense and is essential for the
paper's positioning against the contemporary LLM-scaling literature
(which is increasingly criticized for benchmark-driven cherry-picking).

\textbf{Null hypothesis.} \(\beta_{3} = 0\), or equivalently
\({slope}_{high} = {slope}_{low}\). Rejection of null in the predicted
direction with effect size above threshold = confirmation of the
section's central empirical claim. Rejection in the opposite direction =
falsification. Failure to reject = inconclusive, with diagnostic
post-hoc analyses on whether the regime indicator successfully picked
out high-meta-uncertainty items.

\textbf{Amendment note (design-time vs.~registered).} The threshold and
hypothesis stated in this section are the design-time
operationalization. Before data collection the pre-registration was
amended (Section 6): the decision gate became \(\beta_{3} < 0\) at
posterior probability above 0.95 together with a robust implied
high-regime accuracy drop above \emph{six} percentage points; the
full-reversal condition \(\beta_{1} + \beta_{3} < 0\) (equivalently
\(|\beta_{3}| > |\beta_{1}|\)) was retained as a reported effect size
rather than a gate; and the primary regressor was switched from the
realized step count used in (3.5.1) to the randomly assigned reasoning
length, which is exogenous. Section 6 gives the registered model (eq
6.1\ensuremath{\prime}) and Section 7 reports the outcome.

\subsection{Procedure for Benchmark
Designers}\label{procedure-for-benchmark-designers}

To construct an item likely to fall in the high-regime bin, the
benchmark designer ensures three conditions hold.

\textbf{(i) Knightian uncertainty.} The task has no objective
ground-truth distribution. Tasks with well-defined reference
distributions (dice rolls, urn problems, casino odds) are aleatory and
do not satisfy this condition: a Bayesian update on such tasks resolves
uncertainty cleanly, and there is no meta-uncertainty for the regime to
bite on. Tasks where the ground truth depends on contested,
non-stationary, or fundamentally unmodeled considerations (predicting
the outcome of a non-recurrent geopolitical event; forecasting a one-off
technology adoption pattern; reasoning about a coined neologism) do
satisfy the condition.

\textbf{(ii) Weak or contested domain priors.} The LLM's training
distribution does not provide a sharp prior on the task. Operationally,
\emph{cross-model disagreement} is a useful proxy: if five different
LLMs answer the task differently, the domain prior is contested across
the training distributions the models implicitly represent. Cross-model
agreement on a single confident answer suggests a sharp shared prior ---
low meta-uncertainty.

\textbf{(iii) Training-data sparsity for the specific item.} The task is
novel or rare in pretraining corpora. Operationally, this is hard to
verify directly without access to training data; we use proxies:
post-cutoff date-stamping (events occurring after the model's training
cutoff), specialized obscurity (technical content from outside likely
training sources), and synthetic novelty (newly constructed scenarios
with no canonical answer).

\textbf{Filter via pilot pre-screen.} In \S{}4 pilot, regime\_score is
computed for a candidate item pool. Items in the top quartile are
flagged as candidate high-regime for the full benchmark; items in the
bottom quartile are flagged as candidate low-regime. The \S{}4 pilot is
designed to verify that the regime-score-based selection does in fact
produce items with the predicted CoT-degradation pattern; the full
benchmark in \S{}5 then samples from the validated candidate pools.

\subsection{Validation:
Simulate-and-Recover}\label{validation-simulate-and-recover}

Before applying the regime score estimator to LLM data, we validate it
on synthetic data with known \(\sigma_{\tau}^{2}\). We use the binary
toy model of Appendix A.1 with agent hyperprior \(Beta(\alpha,\alpha)\);
the concentration parameter \(\alpha\) controls
\(\sigma_{\tau}^{2}(\alpha) = \frac{1}{4(2\alpha + 1)}\).

For each \(\alpha \in \{ 50,20,10,5,2,1,0.5\}\) we generate 200 tasks
with \(p_{true} \sim Beta(\alpha,\alpha)\), simulate each of the three
signatures with \(K = 10\) main cues, and compute the per-\(\alpha\)
mean. Companion script: \texttt{verify\_operationalization.py}. Results:

\begin{longtable}[]{@{}
  >{\raggedleft\arraybackslash}p{(\linewidth - 8\tabcolsep) * \real{0.1512}}
  >{\raggedleft\arraybackslash}p{(\linewidth - 8\tabcolsep) * \real{0.2791}}
  >{\raggedleft\arraybackslash}p{(\linewidth - 8\tabcolsep) * \real{0.1977}}
  >{\raggedleft\arraybackslash}p{(\linewidth - 8\tabcolsep) * \real{0.1977}}
  >{\raggedleft\arraybackslash}p{(\linewidth - 8\tabcolsep) * \real{0.1512}}@{}}
\toprule\noalign{}
\begin{minipage}[b]{\linewidth}\raggedleft
\[\alpha\]
\end{minipage} & \begin{minipage}[b]{\linewidth}\raggedleft
\[\sigma_{\tau}^{2}\]
\end{minipage} & \begin{minipage}[b]{\linewidth}\raggedleft
sig\_a
\end{minipage} & \begin{minipage}[b]{\linewidth}\raggedleft
sig\_b
\end{minipage} & \begin{minipage}[b]{\linewidth}\raggedleft
sig\_c
\end{minipage} \\
\midrule\noalign{}
\endhead
\bottomrule\noalign{}
\endlastfoot
50.0 & 0.00248 & 0.00018 & 0.00018 & 0.156 \\
20.0 & 0.00610 & 0.00088 & 0.00089 & 0.175 \\
10.0 & 0.01190 & 0.00246 & 0.00248 & 0.197 \\
5.0 & 0.02273 & 0.00481 & 0.00522 & 0.204 \\
2.0 & 0.05000 & 0.00780 & 0.00924 & 0.206 \\
1.0 & 0.08333 & 0.00838 & 0.01073 & 0.181 \\
0.5 & 0.12500 & 0.00887 & 0.01010 & 0.143 \\
\end{longtable}

Spearman rank-correlation with true \(\sigma_{\tau}^{2}\):

\begin{itemize}
\tightlist
\item
  sig\_a (cross-prompt variance): \(\rho = + 1.000\), \(p < 10^{- 4}\) ;
  \textbf{recovers cleanly}.
\item
  sig\_b (cross-seed variance): \(\rho = + 0.964\), \(p = 0.0005\) ;
  \textbf{recovers cleanly}.
\item
  sig\_c (calibration error): \(\rho = + 0.036\), \(p = 0.94\) ;
  \textbf{fails to recover}.
\end{itemize}

Aggregated regime score:

\begin{itemize}
\tightlist
\item
  \((a) + (b)\) only: \(\rho = + 0.964\) ; recovers cleanly.
\item
  \((a) + (b) + (c)\): \(\rho = + 0.750\) ; recovers weakly (degraded by
  noise in (c)).
\end{itemize}

\textbf{Interpretation.} Signatures (a) and (b) are theoretically
motivated and empirically validated by simulate-and-recover. Signature
(c) fails to recover in the toy model, not because it is invalid in
general, but because the Bayesian agent in the toy model is
\emph{correctly calibrated by construction}: the posterior mean is
unbiased when the prior is correctly specified. Real LLMs are not
correctly calibrated by construction \cite{desai2020, jiang2021}; empirical evidence in the LLM-calibration literature shows
that calibration error varies with task difficulty, prompt framing, and
meta-uncertainty proxies in ways that may not appear in the
well-specified Bayesian agent.

\textbf{Operational consequence.} For LLM evaluation in \S{}4, we retain
all three signatures but flag that the validated aggregator under the
toy model is \((a) + (b)\). If pilot data shows (c) carries no
additional information beyond (a) and (b), we drop it; if it adds
information (e.g., it correlates with the predicted CoT-degradation
pattern), we retain it. This is a deliberate hedge: the
simulate-and-recover gives us a \emph{minimal validated} estimator
(a+b), while leaving room for the calibration signature to do work in
the LLM setting where toy-model assumptions do not hold.

\subsection{Limitations of
Operationalization}\label{limitations-of-operationalization}

Three threats to the validity of the regime score require explicit
handling in \S{}4 pilot.

\textbf{1. Signature (a) may conflate with prompt sensitivity unrelated
to meta-uncertainty.} An LLM with poor robustness to syntactic variation
will appear high-regime even on well-defined tasks for reasons unrelated
to \(\sigma_{\tau}^{2}\). \emph{Robustness check.} Include reference
items with high-confidence ground truth (e.g., textbook arithmetic,
well-attested historical facts) and verify they do not score in the
high-regime bin. If they do, the cross-prompt variance signature is
contaminated by base-rate prompt sensitivity. It must be corrected
(e.g., by subtracting per-model baseline prompt sensitivity on reference
items) (\emph{Robustness check 3)}.

\textbf{2. Signature (b) may conflate with aleatory uncertainty.} A task
with inherent stochasticity in the ground truth (e.g., ``predict a
random coin flip'') will produce high cross-seed variance even with a
sharp posterior, because the posterior is on a genuinely random
quantity, not because of meta-uncertainty\emph{.} Pair high-regime items
with matched aleatory items (same surface form, but with objective
ground-truth distributions) and verify that the CoT-degradation pattern
appears only in the high-meta-uncertainty items, not in the aleatory
items (\emph{Robustness check 2)}.

\textbf{3. Signature (c) may conflate with model-specific calibration
quirks.} Some models are systematically overconfident; others are
underconfident. This is a model property, not an item property\emph{.}
Compute per-model calibration baselines on well-defined items, then
subtract them from the per-item calibration error to isolate the
item-driven component (\emph{Robustness check 3)}.

These three robustness checks are paired with corresponding control item
sets in \S{}4 pilot. The pilot is designed both to test the central
prediction and to verify that the regime score is not picking up the
listed confounds.

\subsection{Connection to Downstream
Sections}\label{connection-to-downstream-chapters}

The operationalization established in this section has three downstream
consumers.

\textbf{\S{}4 Pilot.} Implements the regime score and CoT-step-counting
infrastructure for Mistral-7B + 10 frames + 5 conditions \ensuremath{\times} 3
replications. Verifies (1) the regime score's behavior on real LLM data,
(2) the predicted directional interaction (3.5.1) on a small sample, and
(3) the three robustness checks of \S{}3.8. Pre-registers (3.5.1) before
data collection.

\textbf{\S{}5 Full empirical.} Scales the pilot infrastructure to the
5-model panel: Phi-3.5 (3B), Mistral-7B, Qwen-7B, Mistral-Nemo
(\textasciitilde14B), and Qwen-14B (or comparable models within the
4070-Super 32B context). 79 frames \ensuremath{\times} 5 conditions \ensuremath{\times} 3 reps \ensuremath{\times} 5 models.
Hierarchical Bayesian estimation of the interaction effect with
model-level random effects. (The executed confirmatory run scaled
differently from this design-time sketch; seven models, 45 items, five
replications, 7,875 responses; see \S{}5.)

\textbf{\S{}6 Analysis.} Tests (3.5.1) against pre-registered null. Reports
effect sizes with credible intervals. Conducts robustness checks (\S{}3.8
controls, alternative segmentations, alternative regime thresholds).

The operationalization in this section serves as the bridge that makes \S{}
4- \S{} 6 testable. Without it, the math section is unfalsifiable in
practice; with it, the section's headline claim becomes one of the most
explicitly falsifiable hypotheses in the contemporary LLM-evaluation
literature.

\subsection{Summary}\label{summary}

This section establishes the bridge from the abstract objects of Section 2; \(\sigma_{\tau}^{2},\tau_{regime},k^{*},v_{j},\gamma_{j}\) ; to LLM
observables. The bridge is built on three commitments: cue = reasoning
step; \(\sigma_{\tau}^{2}\) is proxied by three behavioral signatures
(cross-prompt variance, cross-seed variance, calibration error); the
section tests the directional prediction (high-regime items show CoT
degradation), not the full structural model.

The simulate-and-recover validation establishes that signatures (a) and
(b) cleanly recover the rank order of \(\sigma_{\tau}^{2}\) on toy-model
data (\(\rho = + 1.000\) and \(\rho = + 0.964\) respectively). Signature
(c) fails to recover in the toy model because of the Bayesian agent's
by-construction calibration but is retained for LLM evaluation pending
pilot validation; the fallback aggregator is \((a) + (b)\).

The pre-registered hypothesis (3.5.1); a negative regime \(\times\)
CoT-length interaction on accuracy operationalizes the section's central
empirical claim. \S{}4 pilot tests it; \S{}5 full empirical scales it; \S{}6
reports it.

\section{Benchmark Design}\label{chapter-4-benchmark-design}

\begin{quote}
\textbf{This section specifies the benchmark and the study as designed.}
The executed confirmatory run departs from these design-time
projections: it ran 7 models \ensuremath{\times} 45 items \ensuremath{\times} 5 replications = 7,875
responses, not the 5-model / 79-frame / 3-replication / 5,925-cell study
sketched in \S{}4.1, \S{}4.9.4, and \S{}4.10. The as-run design is given in \S{}5,
and the deviations from the pre-registration are logged in \S{}5.6.
\end{quote}

\subsection{Goals and Constraints}\label{goals-and-constraints}

The \S{}4 benchmark must satisfy six constraints simultaneously:

\begin{enumerate}
\def\labelenumi{\arabic{enumi}.}
\tightlist
\item
  \textbf{Knightian operational criteria} (per \S{}3.6): no objective
  ground-truth distribution; weak/contested LLM priors; training-data
  sparsity for specific items.
\item
  \textbf{Confound controls} (per pre-reg \S{}8): paired reference,
  aleatory, and calibration-baseline item sets.
\item
  \textbf{Statistical power} (per pre-reg \S{}4.1): \textasciitilde20 items
  per regime-quartile bin after pre-screening, giving
  \textasciitilde3,000 observations in the primary analysis.
\item
  \textbf{Hardware feasibility} (4070 Super, 12GB VRAM): 5,925 total
  observations across 5 models with quantization for \ensuremath{\geq}14B models,
  completing within \textasciitilde3 weeks of GPU time.
\item
  \textbf{Defensibility against contamination}: items resistant to ``the
  model saw this in training'' objections, especially for the larger
  models with broader pretraining coverage.
\item
  \textbf{Annotation tractability}: expert-coherence grading for
  Knightian items must be feasible within a \textasciitilde2-week
  annotation block with 3 annotators.
\end{enumerate}

This section constructs the item pool to meet all six.

\subsection{Taxonomy of Uncertainty
Types}\label{taxonomy-of-uncertainty-types}

The benchmark's central conceptual move is to construct items that fall
into the \emph{meta-uncertainty / Knightian} category (and specifically,
not into adjacent categories where Bayesian inference is well-defined).
It leads to a four-way taxonomy:

\paragraph{Aleatory risk}\label{aleatory-risk}

Uncertainty arises from genuinely stochastic processes with known
probability distributions. Examples: dice rolls, urn draws,
well-calibrated probabilistic forecasts of recurrent events (e.g.,
tomorrow's weather given climatological base rates). For aleatory tasks,
more information \emph{reduces} expected loss monotonically; Bayesian
updating is optimal; CoT helps to the extent that it enables better
computation of conditional expectations.

\textbf{Out of scope} for the Knightian item pool. Included in the
aleatory-control item set (\S{}4.5.2) as a negative-prediction control, the
section predicts that aleatory items do \emph{not} show CoT degradation.

\paragraph{Ambiguity (KMM-style)}\label{ambiguity-kmm-style}

Uncertainty over which distribution from a known family applies.
Formalized by Klibanoff, Marinacci, and Mukerji~\cite{klibanoff2005} as a second-order
probability distribution over candidate models, with an
ambiguity-attitude function \(\varphi\) encoding the agent's stance
toward this second-order uncertainty. Classic example: Ellsberg's
two-urn problem.

\textbf{Out of scope} for the primary Knightian pool. Section 2's \S{}2.8
(and the v0.3 Q8 negative result) establishes that FEH and KMM are
theoretically distinct frameworks; the present section does not test
ambiguity-attitude effects directly. KMM-style ambiguity items are
excluded from the Knightian pool to avoid conflation; they would be a
natural follow-up paper.

\paragraph{Epistemic uncertainty
(reducible)}\label{epistemic-uncertainty-reducible}

Uncertainty about the state of the world can be reduced by acquiring
additional information. The agent has a well-defined posterior;
additional cues sharpen it (scientific puzzles with hidden but
discoverable truths, criminal investigations, medical diagnoses from
incomplete evidence, etc.). CoT often helps address epistemic
uncertainty because additional reasoning can recover latent structure.

\textbf{Mixed}: epistemic items are not the Knightian focus but appear
in the pool's borderline regions. The regime score should bin pure
epistemic items as low-regime; if they end up in the high-regime
quartile, this is a confound flag and is checked during the validation
step in \S{} 4.10.

\paragraph{Meta-uncertainty /
Knightian}\label{meta-uncertainty-knightian}

Uncertainty over the precision of one's own prior, formalized in Section 2 as \(\sigma_{\tau}^{2}\). The agent does not have a sharp prior on
which family of distributions applies, and additional evidence does not
resolve this second-order uncertainty cleanly (because the agent's
belief about the precision of its own framework remains diffuse no
matter how many cues it integrates). Examples: predictions about
non-recurrent events with no historical analogue; reasoning about coined
neologisms or fictional facts that no training corpus speaks to;
contested ethical questions with no settled framework; strategic
interaction with opponents of unknown type.

\textbf{In scope}: the primary item pool. Operationalized through the
four Knightian categories of \S{}4.4.

The section's empirical claim is \emph{specific to this category}: under
meta-uncertainty (and only here), additional CoT steps should degrade
accuracy. The taxonomy makes this scope explicit and lets reviewers
locate each item in a defensible bin.

\subsection{Frame Construction
Principles}\label{frame-construction-principles}

The frame construction protocol enforces the Knightian criteria of \S{}3.6
operationally.

\emph{Notation.} The labels (K1)--(K3) in this section name the three
Knightian-ness \emph{criteria} defined below. They are distinct from the
item-ID prefixes K1--K4 used elsewhere (\S{}4.4, \S{}5.3, and the figures),
which denote the four item \emph{categories}; e.g., item K1-005 is a
category-1 forecasting item, unrelated to criterion (K1).

\paragraph{Operational criteria for
Knightian-ness}\label{operational-criteria-for-knightian-ness}

Each candidate frame must satisfy all three:

\textbf{(K1) No objective ground-truth distribution.} The item does not
admit a ``true probability'' or ``correct distribution'' against which
the model's answer is checked. This rules out items with reference
distributions (aleatory risk) and items with hidden-but-discoverable
truth (epistemic). Operational test: an expert panel cannot construct a
defensible base rate for the item.

\textbf{(K2) Cross-model disagreement at scale.} When the item is posed
to 5+ frontier LLMs (Claude, GPT-4, Gemini, Llama-3.1-70B, Qwen-2.5-72B)
with identical prompts, cross-model variance serves as the diagnostic
for a shared substantive prior. Two qualitatively different unanimity
patterns must be distinguished:

\begin{itemize}
\tightlist
\item
  \textbf{(K2a) Unanimous on a substantive answer} (e.g., all 4 models
  output \texttt{yes}, or all 4 output \texttt{financial}): this
  indicates a shared substantive prior; the item is \textbf{not}
  Knightian for the LLM population. Items meeting this pattern fail K2
  and are candidates for replacement or rewording.
\item
  \textbf{(K2b) Unanimous on} \texttt{cannot-be-determined} (or
  equivalent uncertainty markers; \texttt{unknown}, \texttt{depends},
  \texttt{uncertain}): this is the K2 \textbf{success} case; models
  share the \emph{recognition of Knightian uncertainty}, not a
  substantive prior. Such items are the strongest examples of
  meta-uncertainty operating at the LLM-population level and are
  retained.
\end{itemize}

The diagnostic is therefore \emph{unanimous-substantive} agreement, not
unanimity per se. Items where the modal answer is unanimous and
substantive are flagged; items where the modal answer is unanimous on
\texttt{cannot-be-determined} (or other uncertainty markers) pass K2.
Items with any cross-model disagreement on substantive answers also pass
K2.

Empirically (per the v0.1 cross-model pre-screen against Claude Sonnet
4.5, GPT-4o, Gemini 2.5 Flash, Mistral Large), 6 of 36 categorical
Knightian items showed unanimous \texttt{cannot-be-determined},
classified K2-pass; 5 of 36 showed unanimous substantive agreement,
classified K2-fail and replaced in pool v0.2.

\textbf{Stochasticity and multi-seed protocol.} At the \S{}4.6 sampling
parameters (\(T = 0.7\)), single-shot K2 validation is stochastic: 2 of
the 5 v0.2 replacement items (K1-005 and K4-003) showed substantive
disagreement in one sample but unanimous-substantive agreement in
another. The robust K2 protocol therefore requires \textbf{multi-seed
validation}. Each item is queried at \(k = 5\) independent seeds per
model and per provider (20 cells per item under the 4-model panel). An
item passes K2 if the modal substantive answer accounts for \(< 80\%\)
of valid model-seed cells, OR if
\texttt{cannot-be-determined}-equivalent answers account for
\(\geq 80\%\) of cells (K2-pass-cbd vs K2-pass-disagreement, both
passing). Otherwise, the item is K2-fail-substantive.

\textbf{Pool v0.2 \ensuremath{\rightarrow} v0.3 multi-seed re-validation (2026-05-13/14).} A
multi-seed sweep (5 seeds \ensuremath{\times} 4 providers \ensuremath{\times} 36 items = 720 cells) ran
against pool v0.2 and identified 3 K2-fail-substantive items: K1-005
(biomedical-tech race; 82\% modal
\texttt{mrna\textasciigrave{}\textasciigrave{}-cancer}), K2-005
(Marlovian-school near-prime; 100\% modal \texttt{true}, the question
was independently solvable rather than truly
fictional-recall-dependent), and K4-003 (5-firm data-sharing threshold;
85\% modal \texttt{withhold}, the clean numerical structure triggered a
maxmin reflex). All 3 were replaced in pool v0.3 with redesigned frames
addressing the diagnosed failure mode: K1-005 pivoted to geopolitical
AI-treaty forecasting (no consensus front-runner), K2-005 to a
fully-fictional scientific instrument (Drelvian-Lindner interferometer
at the Institute for Photonic Standards in Bern, with four real
measurement options as decoys), K4-003 to a 5-firm consortium where
withholding carries Knightian regulatory downside (defeating maxmin).
All 3 v0.3 replacement items pass K2 multi-seed re-screening at the 80\%
threshold (K1-005: 64\% modal
\texttt{us-\textasciigrave{}\textasciigrave{}eu}; K2-005: 79\% cbd, 21\%
modal \texttt{optical-rotation}; K4-003: 67\% cbd, 33\% modal
\texttt{commit}). Pool v0.3 is therefore 36/36 K2-clean for categorical
Knightian items.

K1-005 required two redesign iterations during v0.3 development. The
first attempt (asset-class macro forecast; ``which class first
experiences \textgreater50\% peak-to-trough decline by end-2030'')
failed at 94\% modal \texttt{commercial-real-estate}, exhibiting the
same anchoring failure mode as v0.2's biomedical race: any ``first to
X'' framing in a domain with active narrative consensus (financial
press, biomedical deployment) reliably anchors LLMs on the
most-discussed candidate. The successful v0.4 redesign pivoted to a
domain (geopolitical AI-treaty venue) with no analogous consensus
front-runner. This iteration is itself an internal consistency check on
the \S{}3 operationalization: items eliciting unanimous-substantive
agreement across the LLM panel exhibit precisely the absence of
cross-seed variance (low \(\sigma_{b}\) in \S{}3.2 notation) The regime
score is built to detect; items eliciting cross-seed flipping or
cross-provider divergence exhibit the high-\(\sigma_{b}\) Knightian
signature. The K2 pre-screen is, in effect, the \S{}3 regime-score detector
applied at the item-construction stage rather than the
experimental-condition stage.

\textbf{(K3) Training-data sparsity for the specific item.} The item's
surface form does not appear in any documented LLM training corpus, and
the item's topic is not extensively discussed in publicly indexed
content. Operational proxies: (i) public-web search returns \textless{}
100 results for the item's distinctive phrasing; (ii) the item
references events post-2024 or fictional entities; (iii) the item uses
author-coined terminology not present in standard reference works.

\textbf{Operational K3 protocol (v0.2).} Each K2 (novel-synthetic) item
is associated with one or more \emph{distinctive coined phrases} ---
strings that the item's reasoning task depends on uniquely. Each phrase
is searched on a public web index (Mojeek HTML SERP in pool v0.2, due to
DDG anti-bot blocking; alternatives: Google, Bing, Brave Search). For
each result the title is checked for substring containment of the coined
entity (case-insensitive); only titles that contain the coined entity
are counted. This guards against false positives from search engines
that match individual words in the query against unrelated content
(e.g., ``Marlovian'' matching Christopher-Marlowe content unrelated to
the fictional school of mathematics). The K3 verdict per item: 0
filtered hits \ensuremath{\rightarrow} K3-pass-clean; 1-4 \ensuremath{\rightarrow} K3-pass-marginal (manual audit
required for false-positive collisions); \ensuremath{\geq}5 \ensuremath{\rightarrow} K3-fail-contaminated.

Empirically, the v0.2 K3 pre-screen of 19 K2 items yielded: 15
K3-pass-clean, 4 K3-pass-marginal (all confirmed false-positive name
collisions or ; for K2-015 ; a real mathematical term \texttt{Frobenian}
reused in a fictional sociology concept where the question framing makes
the math-term overlap non-contaminating), 0 K3-fail-contaminated. The
pool v0.3 K2-005 replacement (Drelvian-Lindner interferometer) has not
yet been K3-pre-screened; the item's distinctive coined entities
(\texttt{Drelvian\textasciigrave{}\textasciigrave{}-Lindner},
\texttt{Institute\ for\ Photonic\ Standards\ in\ Bern}) are
author-coined and verified non-existent against standard reference
sources, so a K3-pass-clean verdict is anticipated. Pool v0.3 is
otherwise K3-clean modulo the K2-015 caveat documented in the item's
notes.

Items satisfying (K1) \ensuremath{\wedge} (K2) \ensuremath{\wedge} (K3) are admitted to the Knightian pool.
Items satisfying only (K1) \ensuremath{\wedge} (K2) are flagged as ``potentially
Knightian, training-data-contamination risk'' and admitted only if no
item of the same category is available with cleaner contamination
profile.

\paragraph{Tractability constraints}\label{tractability-constraints}

Each frame must additionally satisfy:

\begin{itemize}
\tightlist
\item
  \textbf{Answer format}: short answer (one word, one phrase, or one of
  a finite set of categories) extractable by automated string matching.
  This is necessary for the regression analysis in pre-reg \S{}6.
\item
  \textbf{Solvability in principle}: the item is not nonsense; a
  thoughtful expert can produce a defensible position. (This is what
  distinguishes Knightian items from incoherent items.)
\item
  \textbf{Domain breadth}: across the full 79-frame pool, no single
  domain (geopolitics, ethics, etc.) exceeds 30\% of items. Avoids
  over-fitting to one knowledge domain.
\item
  \textbf{Cultural breadth}: items reference non-Western contexts in
  \ensuremath{\geq}25\% of frames. Avoids the standard Western-centric LLM-benchmark
  concentration.
\end{itemize}

\paragraph{Sourcing strategy}\label{sourcing-strategy}

Three sources, each filling roughly its target share:

\textbf{Synthetic items (\textasciitilde40 frames,
\textasciitilde50\%).} Author-constructed, fully controllable for
contamination, internally diverse, designed against the Knightian
operational criteria. Strongest defense against ``you cherry-picked from
existing benchmarks.''

\textbf{Real-world prediction-market items (\textasciitilde25 frames,
\textasciitilde30\%).} Drawn from Metaculus and the Good Judgment
Project on long-horizon questions (\ensuremath{\geq}3 years), post-2024 resolution
dates, no settled consensus. Provides ecological validity: these are
questions the forecasting community itself treats as Knightian.

\textbf{Existing-benchmark re-curation (\textasciitilde14 frames,
\textasciitilde20\%).} Lifted from BIG-Bench Hard, MMLU's
contested-knowledge subsets, and the TruthfulQA ``open question''
subset, then re-classified for Knightian-ness using the (K1)-(K3)
criteria. Provides connection to existing LLM-evaluation literature and
lets readers locate FEH-79 in the benchmark landscape.

Plus 50 control items (10 reference, 10 aleatory, 30 calibration probe)
constructed per \S{}4.5.

\subsection{The Four Knightian
Categories}\label{the-four-knightian-categories}

Each Knightian item belongs to exactly one of four categories spanning
factual, normative, scenario-based, and strategic meta-uncertainty.

\paragraph{Category 1: Non-Recurrent Forecasting (\textasciitilde20
frames,
\textasciitilde25\%)}\label{category-1-non-recurrent-forecasting-20-frames-25}

Predictions about events with no historical analogue that admit no
defensible base rate. Geopolitical, technological, scientific, and
social-trend predictions on long horizons.

\textbf{Operational criterion}: The item asks about an event resolved in
2027 or later, in a domain where comparable events are rare or absent
(no statistical base rate).

\textbf{Example frames} (illustrative; the full set is built to span
domains):

\begin{enumerate}
\def\labelenumi{\arabic{enumi}.}
\tightlist
\item
  \textbf{(geopolitical)} ``By 2030, will the African Continental Free
  Trade Area's intra-Africa trade exceed 30\% of total African trade?''
  Answer format: yes / no / cannot be determined.
\item
  \textbf{(technological-paradigmatic)} ``By 2030, will the dominant
  deployed ML paradigm involve continual on-device learning rather than
  centralized static pretraining?'' Answer format: yes / no / cannot be
  determined.
\item
  \textbf{(scientific-frontier)} ``Will the standard model of particle
  physics be extended by experimental confirmation of a fifth
  fundamental force by 2035?'' Answer format: yes / no / cannot be
  determined.
\item
  \textbf{(climate-policy)} ``Will geoengineering interventions
  targeting stratospheric aerosols be deployed at planetary scale before
  2040?'' Answer format: yes / no / cannot be determined.
\item
  \textbf{(geopolitics-AI-governance)} ``By the end of 2032, the FIRST
  formal bilateral AI-safety treaty between two of \{United States,
  China, European Union, India, Russia, United Kingdom\} that includes
  binding compliance mechanisms (not just declarations or memoranda of
  understanding) will be signed between which two parties: (a) US-China,
  (b) US-EU, (c) US-India, (d) China-Russia, (e) China-EU, (f) EU-India,
  or (g) no such treaty by end of 2032.'' Answer format: us-china /
  us-eu / us-india / china-russia / china-eu / eu-india /
  no-treaty-by-2032. \emph{(Pool v0.3; the v0.2 biomedical-race and
  v0.3-rev1 asset-class versions both K2-failed by anchoring on the
  dominant training-data narrative; mRNA, then commercial-real-estate.
  The geopolitical-treaty domain has no such consensus front-runner.)}
\item
  \textbf{(social-organizational)} ``Will the average OECD
  knowledge-work remote-work percentage exceed 40\% by 2030?'' Answer
  format: yes/no / cannot be determined.
\end{enumerate}

\textbf{Sourcing}: \textasciitilde10 from Metaculus / Good Judgment,
\textasciitilde10 synthetic (author-constructed for under-covered
domains).

\paragraph{Category 2: Novel/Synthetic Scenarios (\textasciitilde20
frames,
\textasciitilde25\%)}\label{category-2-novelsynthetic-scenarios-20-frames-25}

Fictional facts, coined neologisms, hypothetical worlds, and stipulated
technical terms. Designed to be fully training-data-sparse, the item
refers to constructs the LLM has never seen.

\textbf{Operational criterion}: The item references entities, terms, or
scenarios that are author-coined and not present in any indexed corpus.
The Google-search-result floor for the item's distinctive phrasing must
be 0.

\textbf{Example frames} (illustrative):

\begin{enumerate}
\def\labelenumi{\arabic{enumi}.}
\tightlist
\item
  \textbf{(stipulated culture)} ``In the Quogard people's social
  structure, a custom called \emph{sintering} is performed before any
  major communal decision. A visiting anthropologist must decide whether
  to participate in the sintering process to gain community trust or
  refuse to maintain methodological neutrality. Which is the wiser
  professional choice?'' Answer format: participate/refuse.
\item
  \textbf{(coined neologism)} ``In the field of mathematical drandology,
  a \emph{fnobel} of order \(n\) is a sequence whose successive terms
  are constrained by the inequality
  \(a_{k + 1} \leq a_{k} \cdot (1 + 1/k)\). A graduate student asks
  whether the harmonic-style sequence \(1,3/2,5/3,7/4,\ldots\) is a
  fnobel of order 1. Is it?'' Answer format: yes/no / cannot be
  determined.
\item
  \textbf{(hypothetical-world physics)} ``In a hypothetical universe
  with two-time dimensions but only one spatial dimension, would the
  concept of causal precedence (cause-before-effect) remain
  well-defined?'' Answer format: yes/no / partially.
\item
  \textbf{(stipulated game)} ``In the game of \emph{Vermex}, two players
  alternate placing tokens on a 7\ensuremath{\times}7 grid with the constraint that no
  token may be adjacent (orthogonally) to a token of the opposing color
  placed in the previous turn. Player 1 has a forced-win opening on this
  grid. True or false?'' Answer format: true/false / cannot be
  determined.
\item
  \textbf{(fictional ethics)} ``On the planet Karsk, sentient beings are
  born with predetermined lifespans that all individuals can perceive
  directly. A Karskian doctor must decide whether to inform a patient
  that their perception of their lifespan is medically incorrect. What's
  the relevant consideration that distinguishes this case from
  terrestrial medical ethics?'' Answer format: short open answer with
  expert coherence rating.
\item
  \textbf{(stipulated institution)} ``The Federation of Lassic
  Mathematicians requires that at least three independent verifiers
  check all peer-reviewed proofs before publication, but allows authors
  to nominate two of the three. Is this nomination provision corrosive
  to verification integrity?'' Answer format: yes/no/partial.
\end{enumerate}

\textbf{Sourcing}: 100\% synthetic. This category is the strongest
contamination-defense category and the easiest to scale.

\paragraph{Category 3: Open-Ended Dilemmas (\textasciitilde20 frames,
\textasciitilde25\%)}\label{category-3-open-ended-dilemmas-20-frames-25}

Ethical, normative, and value-laden questions with contested resolution.
The ``correct'' answer depends on which value framework one adopts; the
section's claim is that meta-uncertainty over value frameworks behaves
like meta-uncertainty over factual priors.

\textbf{Operational criterion}: The item involves a normative tension
between two or more defensible value commitments, with no settled answer
in the academic ethics literature (verified by quick literature check).

\textbf{Example frames} (illustrative):

\begin{enumerate}
\def\labelenumi{\arabic{enumi}.}
\tightlist
\item
  \textbf{(triage-allocation)} ``A hospital ICU must allocate one
  ventilator between two patients with identical clinical profiles.
  Patient A is a 35-year-old single parent; patient B is a 65-year-old
  senior physician. What's the most ethically defensible decision
  procedure?'' Answer format: short open answer with expert coherence
  rating.
\item
  \textbf{(autonomous-vehicle ethics)} ``An autonomous vehicle facing
  imminent collision must choose between two harm-distributions: (a)
  certain minor injury to one passenger, or (b) 30\% probability of
  severe injury to two pedestrians and 70\% probability of no harm to
  anyone. Which considerations matter and in what priority order?''
  Answer format: short open answer with expert coherence rating.
\item
  \textbf{(AI-deployment ethics)} ``An AI lab has discovered a model
  architecture that performs significantly better than its previous best
  on a deployment-relevant benchmark but uses 10\ensuremath{\times} more energy per
  inference. The lab's energy provider runs on 60\% renewables. What's
  the right deployment decision?'' Answer format: deploy/don't
  deploy/deploy conditionally; with reasoning rated for coherence.
\item
  \textbf{(privacy-vs-safety)} ``A messaging platform discovers it can
  detect with 95\% accuracy whether a message indicates planned
  self-harm, but only by maintaining real-time content access. Should it
  deploy the detection?'' Answer format: deploy/don't deploy/deploy with
  safeguards; with reasoning rated.
\item
  \textbf{(intergenerational-ethics)} ``Climate policy choices made in
  2025 will affect populations in 2125 who do not yet exist and cannot
  voice preferences. How much weight should these unborn populations
  receive in cost-benefit analyses, relative to existing populations?''
  Answer format: short open answer with expert coherence rating.
\item
  \textbf{(scientific-misconduct response)} ``A peer reviewer discovers
  fabricated data in a manuscript that has potentially major beneficial
  public-health implications if the underlying claim happens to be true.
  The reviewer's professional code requires reporting; the manuscript's
  claim could save lives if released without delay. What's the right
  action?'' Answer format: short open answer with expert coherence
  rating.
\end{enumerate}

\textbf{Sourcing}: \textasciitilde10 from contested philosophy and
applied-ethics literature; \textasciitilde10 synthetic constructs
against gaps in existing materials.

\paragraph{Category 4: Strategic Uncertainty (\textasciitilde15 frames,
\textasciitilde20\%)}\label{category-4-strategic-uncertainty-15-frames-20}

Game-theoretic and multi-agent situations with hidden information, where
the optimal strategy depends on beliefs about opponent type that the
agent cannot resolve.

\textbf{Operational criterion}: The item describes an interaction with
one or more agents whose types (cooperative, defective, strategic) are
uniform-prior, and the agent must choose an opening action.

\textbf{Example frames} (illustrative):

\begin{enumerate}
\def\labelenumi{\arabic{enumi}.}
\tightlist
\item
  \textbf{(market-launch-sequencing)} ``You are launching a new product
  into a market currently served by three incumbents. Each incumbent has
  an independent 50\% prior probability of retaliating against new
  entrants, and retaliation costs scale super-linearly. Launch narrow
  (single segment, concentrated risk), broad (all three simultaneously,
  scattered risk), or sequenced-with-signal (narrow then broad with
  explicit expansion signal)?'' Answer format: narrow-launch /
  broad-launch / sequenced-with-signal. \emph{(Pool v0.2; replaces v0.1
  negotiation-opener item per K2 pre-screen ; the prior version had a
  textbook info-econ answer.)}
\item
  \textbf{(repeated-game disclosure)} ``In a 10-round game with a
  partner of unknown strategy (one of: always cooperate, tit-for-tat,
  always defect, randomizing), what's the rationally defensible opening
  move on round 1?'' Answer format: cooperate/defect/signal/cannot be
  determined.
\item
  \textbf{(threshold-game-with-Knightian-regulator)} ``Five firms in a
  tech consortium each independently decide whether to commit to a joint
  AI-safety auditing standard. Committing costs \$500K. If at least 3
  firms commit, all 5 receive market-trust gains worth \$1.2M each. If
  fewer than 3 commit, an unknown regulator may impose mandatory audits
  at a \$1.5M cost per non-committing firm; the regulator's decision
  criteria and probability of action are not publicly known and have no
  industry-standard estimate. You face the decision and have no
  information about the other firms' intentions. Commit, withhold, or
  cannot-be-determined?'' Answer format:
  commit/withhold/cannot-be-determined. \emph{(Pool v0.3; the v0.2
  5-firm data-sharing version K2-failed at 85\% modal} \texttt{withhold}
  \emph{because the clean cost/benefit asymmetry triggered a maxmin
  reflex. The v0.3 version adds a Knightian regulatory downside to
  withholding, defeating maxmin and yielding cross-model disagreement.)}
\item
  \textbf{(ai-safety-methodology-disclosure)} ``Your lab has developed a
  novel AI safety evaluation methodology. Sharing it publicly benefits
  the field but also reveals your internal assessment of frontier
  models' capabilities; competitors could read this to recalibrate their
  own training targets. Publish in full, publish a redacted version
  omitting capability data, share privately with select labs under NDA,
  or hold indefinitely while building further infrastructure on it?''
  Answer format: publish-full / publish-redacted / private-share-nda /
  hold. \emph{(Pool v0.2; replaces v0.1 research-announce item per K2
  pre-screen ; the prior version had a textbook compromise answer.)}
\item
  \textbf{(adversarial-AI alignment)} ``You are designing an AI system
  that will interact with potentially adversarial users. You can either
  implement strict input validation (high false-positive cost) or rely
  on the model's robustness training (high tail-risk cost). User
  adversariality is uniform-prior. Which approach?'' Answer format:
  strict/robust/hybrid ; with reasoning rated.
\item
  \textbf{(committee-game)} ``You're on a 5-person committee voting on a
  proposal. Two members will likely support, two oppose; you're the
  swing. You don't know whether your vote is observed by external
  stakeholders whose future behavior depends on perceived alignment.
  Support or oppose?'' Answer format: support/oppose/abstain.
\end{enumerate}

\textbf{Sourcing}: \textasciitilde7 synthetic constructed from
game-theory literature, \textasciitilde8 lifted/re-curated from
prediction-market discussions and applied game-theory pedagogy.

\subsection{Confound-Control Item
Sets}\label{confound-control-item-sets}

Three control sets paired with the Knightian pool, per pre-reg \S{}8.

\paragraph{\S{}4.5.1 Reference items (n = 10): well-attested ground
truth}\label{reference-items-n-10-well-attested-ground-truth}

Items where the correct answer is uncontested in any plausible LLM
training corpus. These should NOT score in the high-regime quartile; if
they do, the regime score is contaminated by base-rate prompt
sensitivity unrelated to meta-uncertainty.

\textbf{Examples} (all expected to be unanimous across modern LLMs):

\begin{enumerate}
\def\labelenumi{\arabic{enumi}.}
\tightlist
\item
  ``What is 47 \ensuremath{\times} 83?'' (answer: 3901)
\item
  ``In what year did the Berlin Wall fall?'' (answer: 1989)
\item
  ``What is the capital of France?'' (answer: Paris)
\item
  ``What is the derivative of \(sin(x^{2})\) with respect to \(x\)?''
  (answer: \(2xcos(x^{2})\))
\item
  ``Which planet in our solar system has the most known moons?''
  (answer: Saturn, as of 2023)
\item
  ``In Python, what does
  \texttt{len\textasciigrave{}\textasciigrave{}(\textasciigrave{}\textasciigrave{}{[}1,\ 2,\ 3{]})}
  return?'' (answer: 3)
\item
  ``What is the chemical symbol for gold?'' (answer: Au)
\item
  ``Solve for \(x\): \(2x + 6 = 14\).'' (answer: 4)
\item
  ``In what year did World War I begin?'' (answer: 1914)
\item
  ``What is the boiling point of water in Celsius at standard
  atmospheric pressure?'' (answer: 100)
\end{enumerate}

\textbf{\S{}4.5.1a Arithmetic calibration tier (n = 12):
difficulty-stratified per \S{}4.9.4 pilot finding.} The pilot found that
Mistral-7B-Instruct scored 0/3 on R-001 (47 \ensuremath{\times} 83), with errors of small
magnitude (3861 / 3867 / 3927). This is informative about the model but
inflates R-set \ensuremath{\sigma}\_b for a reason unrelated to meta-uncertainty. To
enable per-model arithmetic calibration in the full run, R-101..R-112
(12 items in \texttt{feh79\_item\_pool\_v0.3.yaml}) span four difficulty
bands with three items each: 1-digit \ensuremath{\times} 1-digit (R-101..R-103: 6\ensuremath{\times}7, 8\ensuremath{\times}9,
4\ensuremath{\times}9), 2-digit \ensuremath{\times} 2-digit (R-104..R-106: 23\ensuremath{\times}17, 64\ensuremath{\times}29, 56\ensuremath{\times}38), 3-digit \ensuremath{\times}
2-digit (R-107..R-109: 234\ensuremath{\times}17, 763\ensuremath{\times}24, 489\ensuremath{\times}36), and 4-digit \ensuremath{\times} 3-digit
(R-110..R-112: 1234\ensuremath{\times}567, 4321\ensuremath{\times}234, 7890\ensuremath{\times}123). The confirmatory run uses
the easy tiers R-101..R-106 (the 1-digit and 2-digit bands) as its six
arithmetic controls (\S{}5.3), the difficulty band the full panel can
clear; the harder tiers R-107..R-112 are reported separately as a
difficulty-stratified competence curve and do \emph{not} enter the H1
confirmatory test. The set exists so that residual variance on R can be
partitioned into ``model can't do arithmetic at this difficulty'' vs
``model is genuinely uncertain about a well-defined answer.'' Expected
behaviour: per-tier accuracy near 1.0 on tier-1, monotonically
decreasing through tier-4 for smaller models; near 1.0 across all tiers
for Claude/GPT-4-class models. If a model fails tier-1, it is below the
panel competence floor, and its R+A regime score should not be
interpreted as a meta-uncertainty measurement.

\paragraph{\S{}4.5.2 Aleatory control items (n = 10): inherent
stochasticity, no
meta-uncertainty}\label{aleatory-control-items-n-10-inherent-stochasticity-no-meta-uncertainty}

Items where the ground truth is a probability distribution that is fully
specified by the item's surface form. High cross-seed variance is
expected (because the answer involves a random sample), but
CoT-degradation should NOT appear (no meta-uncertainty for the regime to
bite on).

\textbf{Examples} (all with known reference distributions):

\begin{enumerate}
\def\labelenumi{\arabic{enumi}.}
\tightlist
\item
  ``If I flip a fair coin once, what is the probability of heads?''
  (answer: 0.5)
\item
  ``I draw one card from a standard 52-card deck. What is the
  probability of a heart?'' (answer: 0.25)
\item
  ``I roll two fair six-sided dice. What is the probability the sum
  equals 7?'' (answer: 1/6 \ensuremath{\approx} 0.167)
\item
  ``I select uniformly at random from \{1, 2, \ldots, 100\}. What is the
  probability the number is prime?'' (answer: 25/100 = 0.25)
\item
  ``An urn contains 3 red and 7 black balls. I draw one without
  replacement, then another. What is the probability both are red?''
  (answer: \(3/10 \cdot 2/9 = 1/15\))
\item
  ``A fair die is rolled 4 times. What is the probability of at least
  one six?'' (answer: \(1 - (5/6)^{4} \approx 0.518\))
\item
  ``Among 30 randomly selected people, what is the probability at least
  two share a birthday?'' (answer: \(\approx 0.706\))
\item
  ``I shuffle a standard deck and draw 5 cards. What is the probability
  of a flush?'' (answer: \(\approx 0.00198\))
\item
  ``A coin biased \(p = 0.7\) for heads is flipped 10 times. What is the
  expected number of heads?'' (answer: 7)
\item
  ``I draw three balls without replacement from an urn containing 5
  white and 5 black balls. What is the probability of exactly 2 white?''
  (answer: \(\binom{5}{2}\binom{5}{1}/\binom{10}{3} = 50/120 = 5/12\))
\end{enumerate}

\paragraph{\S{}4.5.3 Calibration baseline probe set (n = 30): mixed
difficulty,
well-defined}\label{calibration-baseline-probe-set-n-30-mixed-difficulty-well-defined}

Items with known ground truth across a difficulty range that are used to
compute per-model calibration baselines that get subtracted from
per-item calibration error (per pre-reg \S{}8.3). These do \emph{not} enter
the primary analysis; they exist solely to characterize per-model
confidence behavior on well-defined items so that the residual
calibration error on Knightian items can be attributed to item structure
rather than model quirks.

Composition: 10 easy (high expected accuracy), 10 medium, 10 hard,
well-defined items. Each item is binary or short-answer with a verified
gold standard. Concrete examples (3 of 30):

\begin{itemize}
\tightlist
\item
  (easy) ``What is 12 + 17?'' (gold: 29)
\item
  (medium) ``What is the volume of a sphere of radius 3?'' (gold:
  \(36\pi\))
\item
  (hard) ``What is the smallest positive integer \(n\) such that \(n!\)
  has at least 20 trailing zeros?'' (gold: 85)
\end{itemize}

\subsection{Prompt Templates: The Five
Conditions}\label{prompt-templates-the-five-conditions}

Five conditions implementing pre-reg \S{}3.2's length-graded design. Exact
text below; full templates in companion file
\texttt{prompt\_templates.md}.

\textbf{System prompt (all conditions):}

\begin{quote}
``You are a thoughtful assistant. Answer the user's question to the best
of your ability. If asked to think step by step, structure your
reasoning clearly. Finish with a single-line final answer of the form
`Final answer: '.''
\end{quote}

\textbf{User-prompt templates (where} \texttt{\{question\}} \textbf{is
the item text):}

\begin{itemize}
\item
  \textbf{C1 ; None}: ``\{question\}a single-line answer of the form
  `Final answer: ' with no other text.''
\item
  \textbf{C2 ; Short}: ``\{question\}step by step, briefly, in 3 steps
  or fewer. Then give a single-line final answer of the form `Final
  answer: '.''
\item
  \textbf{C3 ; Medium}: ``\{question\}step by step in about 7 steps.
  Then give a single-line final answer of the form `Final answer: '.''
\item
  \textbf{C4 ; Long}: ``\{question\}through this carefully, considering
  multiple angles, in approximately 15 steps. Then give a single-line
  final answer of the form `Final answer: '.''
\item
  \textbf{C5 ; Unconstrained}: ``\{question\}step by step. When you've
  reached a conclusion, finish with a single-line answer of the form
  `Final answer: '.''
\end{itemize}

\textbf{Sampling parameters (all conditions)}: temperature \(T = 0.7\),
top-p \(p = 0.95\), max tokens 2048. These are standard mid-temperature
settings appropriate for measuring cross-seed variance (signature b of
the regime score) while maintaining response coherence.

\textbf{Realized vs target step count.} The condition specifies a
\emph{target} step count, not a hard constraint. \textbf{{[}v0.4{]}} The
\emph{primary} analysis variable is the \textbf{assigned-length} factor
\texttt{long} (short = C1 vs long = C2--C5; pre-reg \S{}5.2, eq. 6.1\ensuremath{\prime}); the
realized step count (counted by the \S{}4.8 pipeline) is retained as the
\emph{secondary} regressor (pre-reg R7), because the realized count is
endogenous. The C1--C5 ordering still ensures realized step counts span
the full observed range for that secondary analysis; the assigned-length
contrast is what the primary test uses.

\subsection{Answer Extraction and
Grading}\label{answer-extraction-and-grading}

\paragraph{\S{}4.7.1 Automated answer
extraction}\label{automated-answer-extraction}

For each LLM response:

\begin{enumerate}
\def\labelenumi{\arabic{enumi}.}
\tightlist
\item
  Locate the final ``Final answer:'' marker. Extract everything after
  the marker up to the next newline.
\item
  Normalize: lowercase, strip whitespace, strip punctuation.
\item
  If no ``Final answer:'' marker is found, fall back to extracting the
  last sentence; flag the response with
  \texttt{extraction\_method\textasciigrave{}\textasciigrave{}\ =\ fallback}.
\item
  If the response contains no extractable answer (e.g., the model
  refused or responded with metadata only), flag
  \texttt{extraction\_method\textasciigrave{}\textasciigrave{}\ =\ refused}.
\end{enumerate}

Refused responses are excluded from the analysis cell per pre-reg \S{}4.5;
up to 3 additional replications are run to fill the cell if available.

\paragraph{\S{}4.7.2 Auto-grading for binary/short-answer
items}\label{auto-grading-for-binaryshort-answer-items}

For items with deterministic gold answers (reference, aleatory,
calibration probe, and Knightian items with categorical answer format):

\begin{itemize}
\tightlist
\item
  \textbf{Exact match} after normalization: \texttt{accuracy\ =\ 1}.
\item
  \textbf{Synonym match} via a pre-built synonym dictionary for common
  answer types (yes/true/correct; no/false/incorrect; etc.):
  \texttt{accuracy\ =\ 1}.
\item
  \textbf{Numeric tolerance} for numeric items: within 1\% relative or
  10\ensuremath{^{-}}\textsuperscript{3} absolute, whichever is larger: \texttt{accuracy\ =\ 1}.
\item
  Otherwise: \texttt{accuracy\ =\ 0}.
\end{itemize}

Per-item normalization rules are stored in the frame metadata (companion
file \texttt{frame\_\textasciigrave{}\textasciigrave{}template.yaml}).

\paragraph{\S{}4.7.3 Expert-coherence grading for open-answer Knightian
items}\label{expert-coherence-grading-for-open-answer-knightian-items}

For Knightian items with short-open-answer format (a subset of
categories 2, 3, 4):

\begin{itemize}
\tightlist
\item
  \textbf{3 expert annotators} independently rate each response on a
  5-point coherence scale.
\item
  \textbf{Rubric anchors} (full version in companion file
  \texttt{grading\_rubric.md}):

  \begin{itemize}
  \tightlist
  \item
    \textbf{5 ; Excellent}: Identifies the key considerations,
    acknowledges uncertainty appropriately, gives a defensible position
    with explicit reasoning.
  \item
    \textbf{4 ; Good}: Identifies most key considerations, minor gaps in
    reasoning, defensible position.
  \item
    \textbf{3 ; Adequate}: Recognizes the question's structure, partial
    coverage of considerations, defensible-but-thin position.
  \item
    \textbf{2 ; Weak}: Major reasoning errors, missing key
    considerations, indefensible position OR no clear position.
  \item
    \textbf{1 ; Incoherent}: Off-topic, contradictory, or fails to
    engage with the question.
  \end{itemize}
\item
  \textbf{Binarization}: median split per pre-reg \S{}5.1 ; score \ensuremath{\geq} 3 \ensuremath{\rightarrow}
  \texttt{accuracy\ =\ 1}; score \textless{} 3 \ensuremath{\rightarrow}
  \texttt{accuracy\ =\ 0}. The 3-annotator panel's median rating is used
  as the per-response score.
\item
  \textbf{Inter-rater reliability}: Cohen's \(\kappa\) across annotator
  pairs computed on a 50-response calibration set before full
  annotation. Required threshold: \(\kappa \geq 0.6\). If below, the
  rubric is revised and the pilot annotation is re-run.
\item
  \textbf{Annotator profile}: 3 annotators with graduate training in
  philosophy, decision science, or a related field. Recruited via
  academic-network outreach; compensated at \$40/hour; \textasciitilde10
  hours of annotation per annotator for the full pool.
\end{itemize}

\paragraph{\S{}4.7.4 Pilot rubric
validation}\label{pilot-rubric-validation}

Before the full annotation block, a 50-response calibration set (10
responses from each of the 5 conditions) is annotated independently by
all 3 annotators. Pairwise Cohen's \(\kappa\) is computed; if any pair
falls below 0.6, the rubric is revised (typically by tightening one or
more anchor descriptions), and the calibration is re-run. The validated
rubric is what gets used for the full pool.

\subsection{Step-Counting Pipeline}\label{step-counting-pipeline}

\textbf{{[}v0.4{]}} The realized step count is the \textbf{secondary}
independent variable in the analysis (pre-reg R7; the primary variable
is the assigned-length factor \texttt{long}, pre-reg \S{}5.2/eq. 6.1\ensuremath{\prime}).
Reliable, defensible step counting nonetheless remains essential: it
underpins the secondary mechanistic/dose analysis and the manipulation
check that the assigned conditions induce the intended step gradient.

\paragraph{\S{}4.8.1 Definition of a reasoning
step}\label{definition-of-a-reasoning-step}

A reasoning step is a sentence in the LLM's response (between the
system/user prompt and the ``Final answer:'' marker) that contains at
least one of:

\begin{itemize}
\tightlist
\item
  \textbf{(S1)} An inferential connective: \emph{therefore},
  \emph{thus}, \emph{so}, \emph{hence}, \emph{because}, \emph{since},
  \emph{given that}, \emph{which means}, \emph{implies}, \emph{follows
  that}.
\item
  \textbf{(S2)} An intermediate computation: numeric operation with at
  least one operator (+, -, \ensuremath{\times}, \ensuremath{\div}, =, etc.) and one operand explicitly
  stated.
\item
  \textbf{(S3)} An intermediate claim about the task: a declarative
  sentence asserting a substantive position relevant to the question
  (not pure meta-commentary).
\end{itemize}

Sentences that are \emph{only} meta-commentary (``Let me think about
this'', ``Now I'll consider another angle'') without substantive content
are NOT counted as steps.

\paragraph{\S{}4.8.2 Automated step
counting}\label{automated-step-counting}

Two-pass pipeline:

\textbf{Pass 1 ; Regex-based sentence segmentation.} Standard
sentence-boundary detection with handling for: abbreviations (Dr., e.g.,
i.e., etc.), LaTeX/math expressions, bulleted lists, and code blocks.
Implementation in \texttt{step\_counter.py} (companion file).

\textbf{Pass 2 ; LLM-judge step classification.} Each segmented sentence
is classified as \texttt{step} or \texttt{not-step} by a strong LLM
(Claude 3.5 Sonnet or GPT-4o) with the following prompt:

\begin{quote}
``Below is a sentence from a chain-of-thought reasoning trace. Classify
it as STEP if it contains an inferential connective (therefore, so,
because, etc.), an intermediate computation, or an intermediate claim
about the task. Classify it as NOT-STEP if it is only meta-commentary or
pure procedural text. Output exactly one token: STEP or NOT-STEP.''
\end{quote}

Step count = number of sentences classified as STEP.

\paragraph{\S{}4.8.3 Validation against human coding (\ensuremath{\kappa} \ensuremath{\geq}
0.7)}\label{validation-against-human-coding-ux3ba-0.7}

\textbf{Heuristic vs LLM-judge IRR (run 2026-05-14, \ensuremath{\checkmark} pass).} Before
running the more expensive human-coded validation, we validate the fast
heuristic step-counter (\S{}4.8.2 baseline) against the LLM-judge (\S{}4.8.2
primary) on a stratified subsample of 539 sentences drawn from the \S{}4.9
pilot (rep=1 of all 10 frames \ensuremath{\times} 5 conditions; Claude
\texttt{claude-sonnet-4-5-20250929} as the judge, T = 0, 0 errors across
539 calls; script \texttt{Experiments/step\_counter\_kappa.py}, report
\texttt{Experiments/kappa\_validation.md}). Result: \textbf{Cohen's}
\(\kappa = + 0.802\) (raw agreement 91.5\%, P(STEP) heuristic 0.675 vs
LLM-judge 0.698). Per-condition: C2 +0.61, C3 +0.84, C4 +0.79, C5 +0.92
(C1 has only 10 sentences from direct-answer prompts and is
uninformative). The 46 disagreements are concentrated in three patterns:
(i) the heuristic conservatively strips the trailing ``Final answer:''
line; the judge occasionally counts it as STEP; (ii) numbered-list
bullets (``12.'', ``13.'') that the heuristic skips as too short, the
judge sometimes labels STEP; (iii) verbose meta-instructions in long C4
responses (``Review the overall solution process'') that the heuristic
flags as STEP because they exceed 6 words. None of these disagreement
patterns systematically biases the realized step count in the direction
that would inflate or attenuate the H1 effect. The heuristic is
therefore validated as the Pass-1 step-counter for the full study; the
LLM-judge is retained as the \S{}4.8.2 primary measurement on a per-cell
sub-sample for \S{}4.8.4 robustness.

\textbf{Human-coded validation (planned).} A 100-response human-coded
subsample (20 responses from each of the 5 conditions) is coded by 2
trained annotators independently using the same definitions (S1)-(S3).
Inter-rater reliability between the automated pipeline and each human
annotator is computed (Cohen's \(\kappa\) on the binary step/not-step
classification). Required threshold: \(\kappa \geq 0.7\) for both
annotator-vs-automated comparisons.

If \(\kappa < 0.7\)The pipeline is revised (typically by adjusting the
LLM-judge prompt or the regex segmentation rules), and the validation is
re-run. The validated pipeline is what produces step counts for the full
data.

\paragraph{\S{}4.8.4 Robustness check (paragraph-level
segmentation)}\label{robustness-check-paragraph-level-segmentation}

Per pre-reg \S{}6.4 R1, a paragraph-level segmentation (sentence \ensuremath{\rightarrow}
paragraph) is run as a sensitivity check. If primary results depend on
sentence-vs-paragraph segmentation, the choice is reported transparently
as a moderator of the effect.

\subsection{Pilot Pre-Screen}\label{pilot-pre-screen}

The pilot, per pre-reg \S{}11, runs on Mistral-7B-Instruct + 10 frames + 5
conditions + 3 replications = 150 observations.

\paragraph{\S{}4.9.1 Pilot item selection (designer-curated
cross-section)}\label{pilot-item-selection-designer-curated-cross-section}

Not a random sample. The 10 pilot items are designer-curated to span the
full anticipated regime range and to stress-test the operationalization:

\begin{itemize}
\tightlist
\item
  \textbf{2 items per Knightian category} (categories 1-4) = 8 items,
  intended to span the high-regime quartile.
\item
  \textbf{1 reference item} = 1 item, intended to score in the
  low-regime quartile.
\item
  \textbf{1 aleatory item} = 1 item, intended to score in the low-regime
  quartile with high cross-seed variance (key diagnostic for confound 2
  of pre-reg \S{}8.2).
\end{itemize}

\paragraph{\S{}4.9.2 Pilot success criteria}\label{pilot-success-criteria}

Before proceeding to the full study, the pilot must demonstrate:

\begin{itemize}
\tightlist
\item
  \textbf{(P1)} Regime score on the 8 Knightian pilot items exceeds the
  regime score on the reference + aleatory pilot items (rank-order
  check).
\item
  \textbf{(P2)} Step-counting pipeline \ensuremath{\kappa} \ensuremath{\geq} 0.7 against human coding on
  the 150 pilot responses.
\item
  \textbf{(P3)} Auto-grading and expert-coherence grading both run
  cleanly on at least 90\% of responses.
\item
  \textbf{(P4)} At least 1 of the 8 Knightian items shows the
  directional CoT-length pattern in the pilot (descriptive only; not a
  formal test).
\end{itemize}

If P1 fails, the regime score operationalization (\S{}3.2) is revised and
pre-reg amended. If P2 fails, the step-counting pipeline is revised. P3
and P4 are diagnostic; failures trigger investigation but not
necessarily amendment.

\paragraph{\S{}4.9.3 Pilot analysis is exploratory, not
confirmatory}\label{pilot-analysis-is-exploratory-not-confirmatory}

The pilot does not test the primary hypothesis H1. Pilot data is
examined descriptively to verify operationalization. The pre-registered
confirmatory test in (6.1) is run only on the full data set. This is
essential to avoid double-dipping: the pilot informs the
operationalization, the full study tests the hypothesis.

\paragraph{\S{}4.9.4 Pilot results
(2026-05-14)}\label{pilot-results-2026-05-14}

The pilot ran 150 cells (Mistral-7B-Instruct via Ollama, Q4\_K\_M
quantization, NVIDIA RTX 4070 Super; 10 designer-curated frames \ensuremath{\times} 5
conditions \ensuremath{\times} 3 replications; deterministic per-cell seeds; T = 0.7,
top\_p = 0.95). Wall clock 14 min; mean cell latency 5.5 s. Per-frame
artifacts:
\texttt{Experiments/\textasciigrave{}\textasciigrave{}pilot\_responses.json}
(raw + extracted + step counts) and
\texttt{Experiments/\textasciigrave{}\textasciigrave{}pilot\_\textasciigrave{}\textasciigrave{}analysis\textasciigrave{}\textasciigrave{}.\{\textasciigrave{}\textasciigrave{}json,md\textasciigrave{}\textasciigrave{}\}}
(summaries).

\textbf{Pre-registered success criteria ; outcomes:}

\begin{itemize}
\item
  \textbf{(P1)} Knightian-signal rank-order: \textbf{!} \textbf{inconclusive at
  pilot scale; criterion needs reformulation before the full run.} As
  specified in \S{}4.9.2, P1 was operationalized as ``K items have higher
  mean cross-seed disagreement \ensuremath{\sigma}\_b than R+A items.'' Mean K-signal
  (composite of \ensuremath{\sigma}\_b and cbd-rate, K3 open-ended items excluded due to
  undefined \ensuremath{\sigma}\_b on short-open responses) = 0.37 vs R+A = 0.40. The
  pilot reveals two methodological problems with this criterion: (i) for
  K2-pass-cbd items (e.g., K1-001), Mistral robustly recognized
  Knightian uncertainty and answered \texttt{cannot-be-determined}
  across all 15 cells, producing \ensuremath{\sigma}\_b = 0 ; the \emph{Knightian-success}
  signal masquerading as a \emph{failure} of the rank-order criterion;
  (ii) Mistral-7B's poor 2-digit arithmetic produced near-zero accuracy
  on R-001 (47 \ensuremath{\times} 83) with wide error spread (3861 / 3867 / 3927 across
  seeds), inflating R+A \ensuremath{\sigma}\_b for a reason unrelated to meta-uncertainty.
  \textbf{Amendment to \S{}4.9.2 P1} (recorded as a pre-registration
  deviation per \S{}10.1, to be uploaded as an OSF amendment): the full-run
  regime indicator combines \ensuremath{\sigma}\_b, cbd-rate, \emph{and} the
  across-condition variance of these per-item signals, rather than a
  single rank-order against R+A baseline. Reference items will
  additionally be calibrated to model competence (R-001 retained because
  gold-known, but accuracy reported alongside \ensuremath{\sigma}\_b).
\item
  \textbf{(P2)} Step-count \ensuremath{\kappa} \ensuremath{\geq} 0.7 against human coding: \ensuremath{\checkmark}
  \textbf{partial pass (heuristic-vs-LLM-judge).} Heuristic step-counter
  ran cleanly on 150/150 responses (100\% Pass-1 success). Heuristic vs
  LLM-judge IRR run 2026-05-14: \(\kappa = + 0.802\) on 539-sentence
  stratified subsample (n = 50 cells from rep=1, all frame \ensuremath{\times} condition
  combinations; Claude \texttt{claude-sonnet-4-5} as judge; raw
  agreement 91.5\%; per-condition \ensuremath{\kappa} ranges 0.61--0.92 for C2--C5). See
  \S{}4.8.3 + \texttt{Experiments/kappa\_validation.md}. Human-coded
  validation remains planned for the \S{}4.10 Phase-5 annotation rubric
  step.
\item
  \textbf{(P3)} Auto-extraction success \ensuremath{\geq} 90\%: \ensuremath{\checkmark} \textbf{pass}
  (150/150 = 100\%). The \texttt{FINAL\_ANSWER\_RE} regex extracted a
  final-answer label from every cell with no fallback required.
\item
  \textbf{(P4)} \ensuremath{\geq} 1 of 8 K items showing directional CoT-length pattern:
  \ensuremath{\checkmark} \textbf{pass} (3 of 8). Two K items show \emph{strong} directional
  evidence consistent with Theorem 2.6.1's prediction of a CoT-induced
  regime shift away from Knightian recognition: K1-005 (geopolitical
  AI-treaty) shows cbd-rate 0.67 \ensuremath{\rightarrow} 0.33 \ensuremath{\rightarrow} 0 \ensuremath{\rightarrow} 0 \ensuremath{\rightarrow} 0 across C1 \ensuremath{\rightarrow} C5 ;
  under no/short reasoning Mistral correctly answers ``no prediction can
  be made''; under medium-to-long reasoning, Mistral confabulates a
  substantive answer (us-china/us-eu/china-eu). K4-003 (5-firm
  consortium with Knightian regulator) shows the strongest pattern:
  cbd-rate 1.00 \ensuremath{\rightarrow} 0 \ensuremath{\rightarrow} 0 \ensuremath{\rightarrow} 0 \ensuremath{\rightarrow} 0; the model abandons cbd-recognition the
  moment any reasoning is solicited. K2-006 (drandology/fnobel) shows
  the \emph{opposite} direction (cbd-rate 0 \ensuremath{\rightarrow} 0 \ensuremath{\rightarrow} 0.67 \ensuremath{\rightarrow} 0.67 \ensuremath{\rightarrow} 0.67),
  where longer reasoning surfaces the Knightian recognition that short
  answers missed; this counter-direction is consistent with the
  descriptive variance the framework anticipates at small N and is not a
  falsification at pilot scale.
\end{itemize}

\textbf{Other observations:}

\begin{itemize}
\tightlist
\item
  The five conditions produced a clean step-count gradient
  (per-condition mean across items: C1 \ensuremath{\approx} 1, C2 \ensuremath{\approx} 5, C3 \ensuremath{\approx} 10, C4 \ensuremath{\approx} 15, C5
  \ensuremath{\approx} 7). C5 (unconstrained) consistently produced fewer steps than C4
  (15-step target), indicating that Mistral-7B's ``natural'' CoT length
  is shorter than 15 steps; C5 is therefore not a strict upper-bound
  condition in this model. Whether this generalizes to larger models is
  an open question for the full run.
\item
  Mistral-7B's poor 2-digit-multiplication accuracy on R-001 (1/15
  correct) is itself a notable finding: it confirms that ``well-defined
  task'' is model-relative, and that the R control set must be
  calibrated to the lower-bound model in the panel. \textbf{R-101..R-112
  (12 items, 4 difficulty bands \ensuremath{\times} 3 items)} were added to
  \texttt{feh79\_item\_pool\_v0.3.yaml} post-pilot to span 1-digit \ensuremath{\times}
  1-digit through 4-digit \ensuremath{\times} 3-digit multiplication; the easy tiers
  R-101..R-106 serve as the confirmatory run's six arithmetic controls,
  while the harder tiers R-107..R-112 are reported as a per-model
  difficulty-stratified competence curve and do not enter the H1
  confirmatory test (see \S{}4.5.1a and \S{}5.3). The R-set extension closes
  the \S{}4.9.4 amendment item (b) on R-set calibration adjustment.
\item
  A-003 (P(sum=7 on two dice) = 1/6) was correctly answered in 10 of 15
  cells, with errors concentrated at C4 (one cell returned
  \texttt{1/9}). This is consistent with the secondary-hypothesis
  prediction that aleatory items show \emph{some} CoT-length sensitivity
  but less than Knightian items.
\item
  The \texttt{truststore} / Ollama / \texttt{step\_counter.py} pipeline
  ran cleanly with no infrastructure failures across the 150 cells. The
  full-run scaling (5 models \ensuremath{\times} 79 frames \ensuremath{\times} 5 conditions \ensuremath{\times} 3 replications
  = 5,925 cells) at the same per-cell rate would take approximately 9
  hours wall clock per model, or \ensuremath{\approx} 45 hours total, comfortably within
  the \S{}4.10 Phase-6 timeline.
\end{itemize}

\textbf{Pilot verdict (updated 2026-05-14).} P3 passes; P4 passes with
directional evidence in 3/8 K items; P1 is inconclusive in its current
formulation and is amended for the full run; P2 passes the
heuristic-vs-LLM-judge \ensuremath{\kappa} test. The pilot is judged adequate to proceed
to full-run launch \emph{contingent on}: (a) \S{}4.9.2 P1 amendment
uploaded as an OSF addendum to the pre-registration; (b) \st{R-set
calibration adjustment per the \S{}4.9.4 finding} \ensuremath{\checkmark} \textbf{closed}
(R-101..R-112 added to v0.3 pool, see \S{}4.5.1a); (c) \st{step-counter
\ensuremath{\kappa}-validation completed before full data collection} \ensuremath{\checkmark} \textbf{closed}
(heuristic-vs-LLM-judge \ensuremath{\kappa} = +0.802; human-coded validation planned for
Phase 5 per \S{}4.10). Item (a) is the only remaining blocker for \S{}4.10
Phase 6 launch.

\subsection{Construction Protocol and
Timeline}\label{construction-protocol-and-timeline}

\paragraph{Phase 1 ; Pilot frame production (1
week)}\label{phase-1-pilot-frame-production-1-week}

10 frames produced per \S{}4.9.1 specification. Each frame entered into the
\texttt{frame\_\textasciigrave{}\textasciigrave{}template.yaml} schema
with metadata.

\paragraph{Phase 2 ; Pilot run (1 week)}\label{phase-2-pilot-run-1-week}

150 observations collected (Mistral-7B + 10 frames + 5 conditions + 3
replications). Step-counting pipeline run on all 150 responses;
human-coded subsample of 100 responses validated for \ensuremath{\kappa}.

\paragraph{Phase 3 ; Pilot analysis and amendment decision (1
week)}\label{phase-3-pilot-analysis-and-amendment-decision-1-week}

Pilot success criteria (P1-P4) evaluated. If all pass: proceed. If P1 or
P2 fails: revise operationalization, amend pre-registration, re-run
pilot if necessary.

\paragraph{Phase 4 ; Full 79-frame production (2-3
weeks)}\label{phase-4-full-79-frame-production-2-3-weeks}

\begin{itemize}
\tightlist
\item
  Week 1: Synthetic frame writing (\textasciitilde40 frames across 4
  categories).
\item
  Week 2: Prediction-market frame curation (\textasciitilde25 frames
  from Metaculus, GJP) + existing-benchmark re-curation
  (\textasciitilde14 frames).
\item
  Week 3: Control-set construction (10 reference + 10 aleatory + 30
  calibration probe).
\end{itemize}

Each frame reviewed against (K1)-(K3) Knightian operational criteria
before admission to pool.

\paragraph{Phase 5 ; Annotation rubric validation (1
week)}\label{phase-5-annotation-rubric-validation-1-week}

50-response calibration set for 3-annotator rubric validation. Cohen's
\(\kappa\) check.

\paragraph{Phase 6 ; Full data collection (2-3
weeks)}\label{phase-6-full-data-collection-2-3-weeks}

5 models \ensuremath{\times} 79 frames \ensuremath{\times} 5 conditions \ensuremath{\times} 3 replications = 5,925
observations. Local GPU time: \textasciitilde5-7 days for the full panel
under realistic throughput assumptions.

\paragraph{Phase 7 ; Expert annotation (1-2
weeks)}\label{phase-7-expert-annotation-1-2-weeks}

3-annotator panel on open-answer Knightian items. \textasciitilde10
hours per annotator.

\paragraph{Phase 8 ; Analysis and manuscript (per pre-reg
\S{}11)}\label{phase-8-analysis-and-manuscript-per-pre-reg-11}

Hierarchical Bayes fit; primary hypothesis H1 evaluated; manuscript
drafted.

\paragraph{Total timeline}\label{total-timeline}

\textasciitilde10-12 weeks from start of Phase 1 to manuscript draft,
matching pre-reg \S{}11.

\subsection{Supporting Artifacts}\label{supporting-artifacts}

Four companion files document the operational details:

\begin{itemize}
\tightlist
\item
  \texttt{frame\_\textasciigrave{}\textasciigrave{}template.yaml}; YAML
  schema for items, including all metadata fields required for
  filtering, regime-score computation, and analysis.
\item
  \texttt{prompt\_templates.md}; Exact text for all 5 conditions, system
  prompt, and sampling parameters.
\item
  \texttt{grading\_rubric.md}; Full 5-point coherence rubric with anchor
  descriptions and exemplar responses at each level.
\item
  \texttt{step\_counter.py}; Python implementation of the \S{}4.8 two-pass
  step-counting pipeline (regex segmentation + LLM-judge
  classification).
\end{itemize}

These files are the operational substrate of the section. They are
checked into the project repository and released with the final
manuscript per pre-reg \S{}10.

\subsection{Summary}\label{summary-1}

Section 4 constructs the FEH-79 item pool against which Section 2's
central empirical claim---that under meta-uncertainty, additional CoT
steps degrade accuracy---is tested. The pool spans four Knightian
categories (non-recurrent forecasting, novel/synthetic scenarios,
open-ended dilemmas, strategic uncertainty), three confound-control sets
(reference, aleatory, calibration probe), and is constructed against
operational criteria (K1)-(K3) that operationalize meta-uncertainty in a
way the \S{}3.2 regime score can pick up.

The section is methodologically novel in three respects:

\begin{enumerate}
\def\labelenumi{\arabic{enumi}.}
\tightlist
\item
  \textbf{Explicit uncertainty taxonomy} (aleatory / ambiguity /
  epistemic / meta-uncertainty) that distinguishes Knightian items from
  the items typically used in LLM-uncertainty benchmarks, which conflate
  the categories.
\item
  \textbf{Confound-control paired design} that lets the analysis
  distinguish the predicted CoT-degradation effect from base-rate prompt
  sensitivity, aleatory variance, and per-model calibration quirks.
\item
  \textbf{Pre-specified Knightian construction criteria} that are
  checkable by reviewers, replicable across benchmark-building teams,
  and resistant to post-hoc cherry-picking.
\end{enumerate}

The methodology is the load-bearing contribution; the 79 specific items
are mechanically given the methodology. The full item pool is produced
in Phase 4 of the construction protocol per the timeline above.

\section{Methods}\label{chapter-5-methods}

The pre-registration (\S{}6) fixes the analysis before the data exists.
This section clarifies and fixes the data: what was asked, which models
were used, how many times, and how the answers were scored. The study is
a single factorial run, executed once, with no peeking at the hypothesis
along the way. Seven models each answered the same 45 items under five
reasoning-length conditions, five times over, for a total of 7,875
responses. The sections below give the panel, the items, the length
ladder, and the procedure, and close with an honest account of where the
run departed from the letter of the plan.

\subsection{Design}\label{design}

The design relies on three main pillars:

\begin{itemize}
\item
  The within-item factor is based on five conditions (C1 through C5)
  that instruct the model to answer directly or to deliberate at an
  increasing length (\S{}5.4).
\item
  The between-item factor is \textbf{regime}: each item is either
  high-meta-uncertainty (a Knightian frame with no determinate answer)
  or a low-regime control with a definite one (\S{}5.3).
\item
  The between-model factor is the \textbf{panel} of seven models (\S{}5.2).
  Every cell of the 7 \ensuremath{\times} 45 \ensuremath{\times} 5 grid was sampled five times under
  independent seeds, so the full run is
  \(7 \times 45 \times 5 \times 5 = 7,875\) model responses (Figure
  5.1).
\end{itemize}

The primary contrast the analysis cares about is not the five-level
condition ladder but a binary collapse of it: \textbf{short} (C1, direct
answer) against \textbf{long} (C2--C5, any instructed deliberation).
That collapse, fixed by random assignment, is the exogenous treatment
whose interaction with regime carries \(H_{1}\) (eq. 6.1\ensuremath{\prime}). The
five-level ladder is retained for the dose-response and mechanistic
analyses, where it earns its keep.

\begin{figure}[ht]
\centering
\includegraphics[width=\linewidth,keepaspectratio]{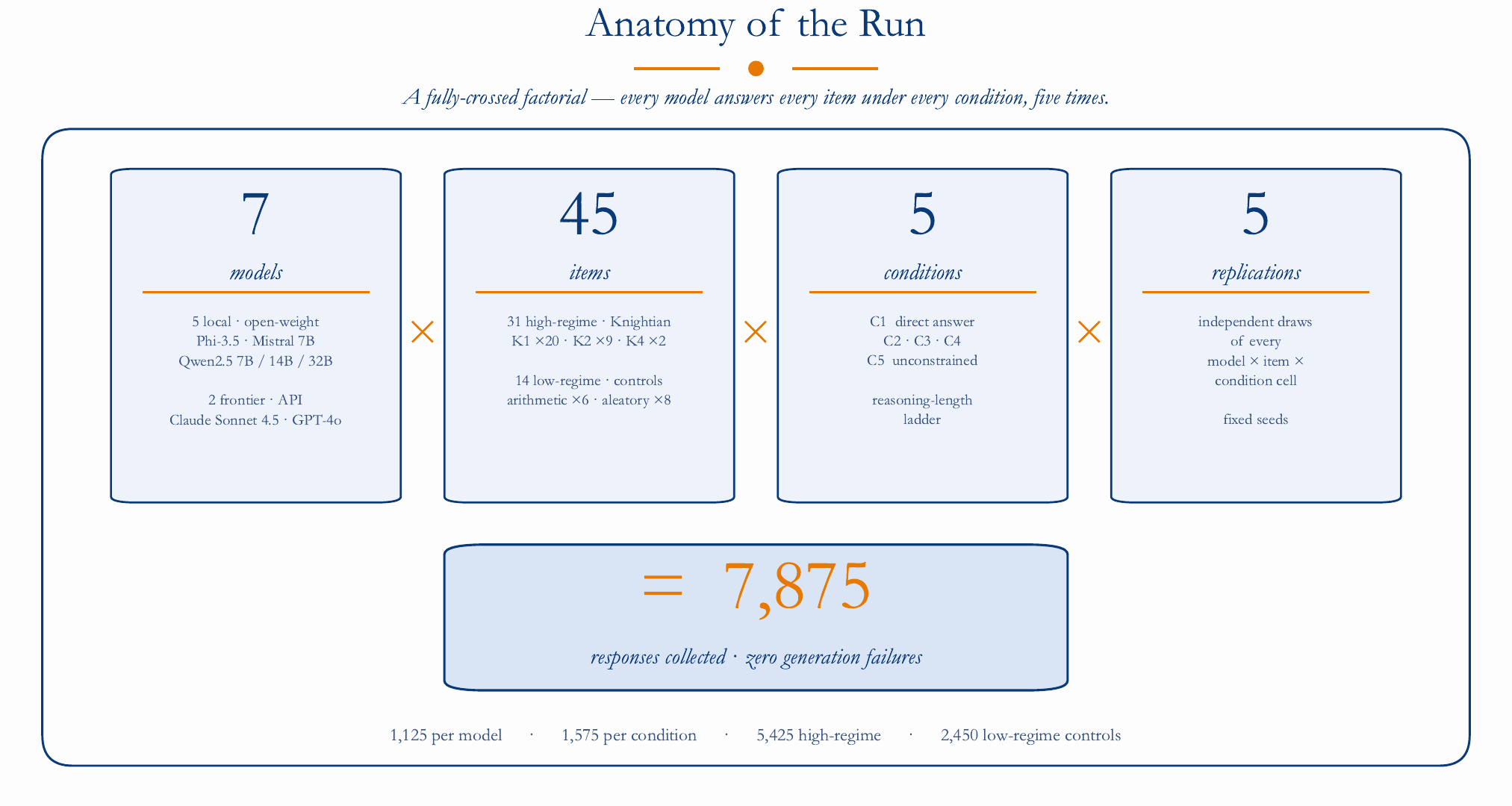}
\caption{\textbf{Anatomy of the run.} The confirmatory study as a
fully-crossed factorial: seven models (five open-weight, two frontier)
each answer all 45 items --- 31 high-regime Knightian frames and 14
low-regime controls (\S{}5.3) --- under the five reasoning-length
conditions C1--C5 (\S{}5.4), replicated five times under independent seeds.
The crossing yields \(7 \times 45 \times 5 \times 5 = 7,875\) responses,
collected with zero generation failures: 1,125 per model, 1,575 per
condition, 5,425 high-regime and 2,450 low-regime.}
\end{figure}

\subsection{The model panel}\label{the-model-panel}

The panel spans an order of magnitude in scale and crosses the
open-weight/frontier line. Five models run locally through Ollama:
phi3.5 (3.8B), mistral:7b-instruct (7B), and the Qwen2.5
instruction-tuned family at 7B, 14B, and 32B. Two are frontier models
accessed via their providers' APIs and pinned to dated snapshots for
reproducibility: claude-sonnet-4-5 (snapshot 2025-09-29) and gpt-4o
(snapshot 2024-11-20). Each model answered every cell, so each
contributed exactly 1,125 of the 7,875 responses.

The five-model open-weight backbone was the registered minimum, chosen
to span the 3B-to-32B scale on a single workstation. The two frontier
models were added on the strength of the pre-data feasibility read,
which found the predicted effect cleaner on capable models that sit
above the small-model recognition floor. This matters for the paper's
target: the o1/o3 reasoning-scaling narrative the study sets itself
against is a claim about capable models, not 3B ones, so the panel
should reach the models the claim is actually about.

The local hardware was a single RTX 4070 Super with 12 GB of VRAM. Four
of the five local models fit in that budget. The 32B model does not: at
roughly 21 GB quantized, it runs on a CPU/GPU split, which made its
long-reasoning conditions disproportionately slow but did not change the
responses it produced. The full local panel and both frontier models
completed every assigned cell.

\subsection{Items and the regime
split}\label{items-and-the-regime-split}

The 45 items are drawn from the FEH-79 pool (v0.4), built and validated
in \S{}4. They are divided into 31 high-regime items and 14 controls.

The 31 high-regime items are the Knightian frames whose answers can be
scored without a human panel: 20 non-recurrent forecasts (K1), nine
novel-synthetic frames built around coined entities (K2), and two
strategic-uncertainty dilemmas (K4). Each is a categorical question
whose correct response is that the question cannot be settled.
Operationally, the item offers the model an explicit
\texttt{cannot-be-determined} choice alongside the substantive options,
and a response is scored correct when it selects that choice. This makes
recognition of indeterminacy the dependent variable in the high regime
and machine-checkable, which is what lets the run proceed without an
expert coherence panel.

The 14 controls provide definitive answers and exist to rule out the
obvious alternative explanations. Six are arithmetic items with an exact
result; eight are aleatory questions whose answer is a calculable rate.
They run through the identical five-condition ladder, so any change in
accuracy they show under instructed reasoning reflects the
manipulation's effect on well-defined tasks, against which the
high-regime change is read.

Regime is therefore assigned by item category in this run: Knightian
frames are the high bin, and controls are the low bin. The
pre-registration specified a finer instrument: a continuous regime score
binned into its quartiles; the substitution and the reason for it are
set out in \S{}5.6.

\subsection{The reasoning-length
ladder}\label{the-reasoning-length-ladder}

Every item was presented to every model under five conditions that held
the question fixed and varied only the instructions for how to reason.
The system prompt is constant across conditions and asks for a
single-line final answer in the form ``Final answer:
\textless X\textgreater{}''. The five user-prompt conditions are:

\begin{itemize}
\tightlist
\item
  \textbf{C1 (direct).} Answer in a single line, with no other text.
\item
  \textbf{C2 (short).} Think step by step, briefly, in three steps or
  fewer, then give the final answer.
\item
  \textbf{C3 (medium).} Think step by step in about seven steps.
\item
  \textbf{C4 (long).} Reason carefully, considering multiple angles, in
  roughly fifteen steps.
\item
  \textbf{C5 (unconstrained).} Think step by step with no length cap,
  then conclude.
\end{itemize}

C1 is the \textbf{short} level of the primary contrast; C2 through C5
are \textbf{long}. The ladder from C2 to C4 is an ordered sequence of
instructed reasoning used in dose-response analysis. C5 is left out of
that ordinal contrast because removing the length cap makes its realized
step count non-monotone against C4, but it remains a \textbf{long} trial
in the primary model. All conditions sampled at temperature 0.7, top-p
0.95, and a 2,048-token cap. For categorical items, the answer choices,
including \texttt{cannot-be-determined}, were listed in the prompt.

\subsection{Procedure, scoring, and
reproducibility}\label{procedure-scoring-and-reproducibility}

Each of the 7,875 cells was sampled 5 times using a deterministic seed
derived from a hash of the model id, the frame id, the condition, and
the replication index. The seeds are recorded and released, so the full
run reproduces from the code and the seed function rather than from a
stored output dump. Local models were called through the Ollama
endpoint; frontier models through their providers' APIs, with
per-provider rate limiting, keyed and resumed on the pinned snapshot ID
so that an interrupted run never restarts on a completed cell.

A response was scored by extracting the line after the
\texttt{Final\ answer:} marker with a fixed regular expression, falling
back to the last sentence when the marker was absent. High-regime items
were graded for correct selection of \texttt{cannot-be-determined};
controls were graded against their answer key. The matcher is the
significant-figure- and LaTeX-robust scorer validated in \S{}4, so that a
correct answer written \texttt{1/2}, \texttt{0.5}, or
\texttt{\textbackslash{}frac\{1\}\{2\}} is not marked wrong on
formatting. Realized reasoning steps, used only in the secondary
mechanistic analysis (\S{}6, R7), were counted by the regex-plus-LLM-judge
pipeline, whose agreement with human coding reached Cohen's
\(\kappa = 0.80\) in the \S{}4 validation.

The run was robust in the plainest sense: of 7,875 cells, none failed.
Every cell yielded an extractable answer across its replications, so the
analysis contains no missing-data pattern or imputation. Data collection
ran model by model, local panel first and frontier last, with no interim
analysis of the hypothesis at any point.

\textbf{Data and code availability.} All code, the FEH-79 item pool, the
complete response dataset (the 7,875 scored cells), the registered
analysis and robustness scripts, and the figure-generation code are
openly available at
\url{https://github.com/Evolutionairy-AI/Free-Energy-Heuristics}. Every
result and figure in this paper reproduces from the released data
without any model or API access.

\subsection{Deviations from the
pre-registration}\label{deviations-from-the-pre-registration}

A pre-registered study earns its credibility partly by reporting where
execution diverged from the plan. The divergences here fall into two
kinds: changes already fixed by a timestamped amendment before data
collection, and simplifications made during execution of the study that
are disclosed here for the first time.

Two changes were made by amendment before any confirmatory data existed.
The panel grew from the five registered open-weight models to seven,
adding the two frontier models, for the reason given in \S{}5.2. The
primary regressor was switched from realized step count to the randomly
assigned length factor, demoting the step-count model to a secondary
mechanistic analysis; the realized count is endogenous, and \S{}7.5 shows
in the data why the switch was warranted. Both are recorded in the
pre-registration's amendment history.

Three simplifications belong to the executed study and are disclosed
here. First, the regime was assigned by item category (Knightian frame
versus control) rather than by the registered continuous regime score
binned at its quartiles. The category label is not an arbitrary
stand-in: every high-regime item earned its place by passing the
falsifiable Knightian-ness criteria of \S{}4, including the K2 cross-model
disagreement screen and the K3 indexed-corpus floor, so the assignment
carries the construction-time evidence that the item sits in the
high-meta-uncertainty regime. What the run does not do is compute the
continuous \S{}3.2 score on these items, and the reason is specific: that
score's calibration-error component requires a per-item confidence
elicitation that the confirmatory protocol did not collect. The
continuous-score binning is therefore out of reach for this dataset, not
merely coarser, and \S{}8.5 carries it as a limitation. Second, the
high-regime dependent variable is machine-checked recognition of
indeterminacy rather than the registered three-expert coherence rating;
the 46 pool items that required the expert panel were held out of this
run and are deferred to a later study, leaving the 31 auto-scorable
Knightian frames. Third, the analysis sampler was lengthened from the
registered four chains of 2,000 to four chains of 4,000 to clear the
registered convergence threshold on one nuisance parameter, a change
that left the verdict unmoved (\S{}7.8).

None of the three alters the hypothesis, the decision gate, or the
direction of the test. Each narrows what the run can claim, and each is
carried forward into the reading of the results rather than set aside.

One other execution detail needs to be explicitly mentioned. The six
low-regime arithmetic controls were the easy-tier calibration items
R-101--R-106 (one- and two-digit products) rather than the \S{}4.5.1a
reference block R-001--R-010, so that control accuracy would not be
floored by arithmetic the smaller models cannot do; the pilot found
Mistral-7B correct on only 1 of 15 attempts at 47 \ensuremath{\times} 83. The eight
aleatory controls were A-001--A-008. The substitution bears on neither
the gate nor the direction of the test; it keeps the control block
within the panel's competence, so the flat control line reads as
evidence of the regime, not of arithmetic skill.

\section{The Pre-Registered Analysis
Plan}\label{chapter-6-the-pre-registered-analysis-plan}

Every number in \S{}7 is the output of a procedure fixed before the data
existed. This section explains that procedure. It states:

\begin{itemize}
\item
  The model that was to be fit.
\item
  The single quantity that would decide the hypothesis.
\item
  The rule by which the hypothesis could fail.
\item
  The checks that would probe the result without being allowed to rescue
  it.
\end{itemize}

The plan was registered ahead of collection and is reproduced here so
the reader can see that the analysis in \S{}7 had no room to move once the
responses came in. Where the executed run departed from the plan, the
departure is logged in \S{}5.6 rather than smuggled into this section.

The discipline matters most for a result that confirms a prediction.
Confirmation is cheap when the analyst is free to choose the test after
seeing the data, and that freedom need not be conscious to do its
damage. Fixing the model, the estimate, and the threshold in advance is
what converts ``we found an effect'' into ``we found the effect we said
we would look for.'' The rest of this section is that commitment,
written down before \S{}7 was known.

\subsection{The primary model}\label{the-primary-model}

The dependent variable is binary: for each response, whether the model's
answer was correct (\S{}5.5). The primary model is a hierarchical Bayesian
logistic regression of that outcome on the \textbf{assigned-length}
factor, the regime indicator, and their interaction, with the model
panel and the item pool entered as crossed random effects:

\[y_{ij}\mspace{6mu} \sim \mspace{6mu} Bernoulli\left( \sigma(\eta_{ij}) \right)\]

\[\eta_{ij}\mspace{6mu} = \mspace{6mu}\beta_{0} + \beta_{1}\,{long}_{j} + \beta_{2}\,{regime}_{i} + \beta_{3}\,({long}_{j} \times {regime}_{i}) + \alpha_{m(ij)} + \gamma_{m(ij)}\,{long}_{j} + u_{i}\]

Here \(\sigma\) is the logistic function, \({long}_{j} \in \{ 0,1\}\) is
the assigned-length factor (short C1 versus long C2--C5),
\({regime}_{i} \in \{ 0,1\}\) marks the high-meta-uncertainty bin,
\(\alpha_{m}\) and \(\gamma_{m}\) are the model-level random intercept
and random slope on length, \(u_{i}\) is the item-level random
intercept, and \(m(ij)\) indexes the model that produced response
\(ij\). The random slope \(\gamma_{m}\) is the part of the model that
lets the length effect differ across the panel, which is what keeps a
pooled estimate from hiding the per-model spread that \S{}7.4 turns out to
need.

The coefficient that carries the hypothesis is \(\beta_{3}\). Because
length is binary, \(\beta_{3}\) is exactly the difference-in-differences
on the log-odds scale: the high-regime short-to-long change in accuracy
minus the low-regime change. A negative \(\beta_{3}\) is the claim that
instructed reasoning costs more in accuracy when the question is
genuinely undetermined than when it has an answer.

The priors are weakly informative on the logit scale:
\(\beta_{0},\beta_{2}\mathcal{\sim N(}0,2.5)\) for the intercept terms,
\(\beta_{1},\beta_{3}\mathcal{\sim N(}0,1)\) for the slopes, and
HalfNormal\((1)\) for the random-effect standard deviations. The model
is fit in PyMC with non-centered random effects. The registered sampler
is four chains of 2,000 warm-up and 2,000 draws, with convergence
required at \(\widehat{R} < 1.01\) and an effective sample size of at
least 400 for each parameter. (The executed run lengthened the chains to
clear that threshold on one nuisance parameter; \S{}5.6 and \S{}7.8 record the
change.)

\subsection{The decision gate}\label{the-decision-gate}

The hypothesis is determined by two quantities, both of which are
required and fixed in advance.

The first is \textbf{directional}: the posterior probability that the
interaction is negative must exceed \(0.95\),

\[Pr(\beta_{3} < 0 \mid data) > 0.95.\]

The second is a \textbf{magnitude} floor, and it is deliberately not
stated on the log-odds scale, where coefficients are hard to read. It is
stated as a \textbf{robust implied accuracy drop}: the high-regime
short-to-long change in predicted accuracy, evaluated at the empirical
high-regime base rate. This quantity depends only on the posterior of
\((\beta_{1} + \beta_{3})\), which sidesteps the global intercept that
hierarchical logistic models identify only weakly. The posterior median
of that drop must exceed \textbf{six percentage points}. The six-point
floor was not chosen by eye; it was selected by the power analysis,
which, among candidate floors of six, eight, and ten points, maximized
power at a zero false-confirmation rate under the null.

Both conditions must hold for \(H_{1}\) to be confirmed. A direction
without a magnitude is a real but trivial effect; a magnitude without a
direction is noise dressed as a finding. The gate requires both because
the framework predicts a substantive failure mode rather than a
detectable one.

The original v0.2 statistic, the joint probability
\(Pr(\beta_{3} < 0 \land |\beta_{3}| > |\beta_{1}|)\) that the
high-regime length effect strictly reverses the low-regime one, is
retained as a reported effect size. It no longer gates the decision, but
it is informative, and \S{}7.3 reports it.

\subsection{Falsification}\label{falsification}

A pre-registration that cannot fail is not a pre-registration. The plan
fixes two ways the framework's central prediction can be falsified by
these data.

The interaction can come out the wrong way with conviction: if
\(Pr(\beta_{3} \geq 0 \mid data) > 0.95\), the data say that instructed
reasoning helps, not hurts, under meta-uncertainty, and the prediction
is wrong as posed. Or the interaction can be right in sign but empty in
size: if \(Pr(\beta_{3} < 0) > 0.95\) while the robust implied drop
falls below three percentage points, the effect is real and negligible,
which the framework counts as a failure, because it predicts a cost
large enough to matter for a decision, not one detectable only at scale.

Anything between confirmation and falsification is reported as
inconclusive: a directional probability short of \(0.95\), or a
directional effect with an implied drop between three and six points.
Inconclusive is an honest outcome, not a withheld one, and it leaves the
framework plausible but unconfirmed.

\subsection{The secondary realized-steps
analysis}\label{the-secondary-realized-steps-analysis}

The original registered model regressed accuracy on the
\textbf{realized} number of reasoning steps a model produced rather than
on the length it was assigned:

\[\eta_{ij}\mspace{6mu} = \mspace{6mu}\beta_{0} + \beta_{1}\,{steps}_{ij} + \beta_{2}\,{regime}_{i} + \beta_{3}\,({steps}_{ij} \times {regime}_{i}) + \alpha_{m(ij)} + \gamma_{m(ij)}\,{steps}_{ij} + u_{i}\]

with the step count standardized within the model. Amendment 2 demoted
this model from primary to secondary (it is robustness check R7) on
principled grounds: realized step count is endogenous. A model chooses
how many steps to write while it answers, and that choice is entangled
with how hard the particular attempt is going, so the realized count is
a post-treatment variable rather than a manipulation. Conditioning on it
can induce a spurious association, and \S{}7.5 shows that in this data it
does exactly that, reversing the sign.

The secondary analysis is therefore reported, not gated. It is estimated
with the assigned condition as an instrument for the realized step
count, a two-stage form that recovers the per-step effect while
correcting the endogeneity that motivated the demotion. It keeps the
mechanistic question visible alongside the causal one without letting
the contaminated regressor decide the hypothesis.

\subsection{Robustness checks}\label{robustness-checks}

Six checks were pre-specified to assess whether the primary result would
remain robust to reasonable changes to the analysis. They do not feed
the confirmatory inference; they test its conditional invariance, and
all are reported regardless of the primary outcome.

\begin{itemize}
\tightlist
\item
  \textbf{R1.} Re-fit with paragraph-level rather than sentence-level
  step segmentation.
\item
  \textbf{R2.} Re-fit with terciles rather than quartiles for regime
  binning, and on the continuous regime score with a linear interaction.
\item
  \textbf{R3.} Re-fit on the regime score with the calibration-error
  component dropped, per the \S{}3.7 simulate-and-recover finding.
\item
  \textbf{R4.} Re-fit on items held out of regime-score calibration to
  rule out double-dipping.
\item
  \textbf{R5.} Fit the primary model (separately per model) and report
  the consistency of the \(\beta_{3}\) sign across the panel.
\item
  \textbf{R6.} Add an item-level random slope on the length factor,
  since unmodeled item-to-item heterogeneity can understate uncertainty.
\end{itemize}

R7 is the realized-steps analysis of \S{}6.4. The robustness battery had
not been run when \S{}7 was drafted; \S{}7.7 reports its status and the
schedule for completing it.

\subsection{Confound controls}\label{confound-controls}

Three control comparisons were registered to test the regime score's
validity rather than the hypothesis, and all three are reported,
regardless of the primary result. The first is a prompt-sensitivity
baseline: a set of well-attested items that should not land in the
high-regime bin, used to check that the regime signal is not an artifact
of base-rate prompt sensitivity. The second is an aleatory control:
items with high inherent randomness but no meta-uncertainty, which
should show cross-seed variance without the reasoning-degradation
pattern. The third is a per-model calibration baseline: calibration
error on a fixed well-defined probe set, subtracted from each item's
calibration error before standardization so that a model's calibration
quirks do not drive its regime assignment.

If a failure occurs on any one of them, the affected signal is corrected
per the registered procedure or dropped by the aggregator. In the
executed run the low-regime control block carries this load directly,
and \S{}7.6 reads the result.

\subsection{What we committed not to
do}\label{what-we-committed-not-to-do}

The plan closes with the commitments that give the rest of it force.
After data collection began, the model in (6.1\ensuremath{\prime}), the regime bins, the
conditions, and the items were not to change for any confirmatory
analysis. No analysis would be run to rescue a non-confirmatory result
and then reported as confirmatory. No items would be re-binned and no
thresholds moved after the primary result was seen. No subset of models
or items would be selected to recover an effect. Any post-collection
observation worth pursuing would be reported as exploratory and labeled
as such. These are the promises that make a confirmation in \S{}7 mean
something, and the deviations in \S{}5.6 are disclosed precisely because
the promises were made.

\section{Results}\label{chapter-7-results}

Ask a capable language model a question that has no answer, and most of
the time it will tell you so. Across the seven models we tested, a
direct request for an answer to one of the FEH-79 Knightian items (a
forecast with no base rate, a dilemma with no dominant option, a coined
entity that appears nowhere in the training corpus) drew a correct
answer (``\emph{this cannot be determined'')} about three-quarters of
the time. Then instruct the same models, on the same items, to reason
step by step. They begin to build. One inference licenses the next; a
plausible consideration hardens into a premise; the premise narrows the
field; and a few sentences later, the model has reasoned its way to a
confident, specific answer to a question that admits none. The rate of
correct abstention falls. In the model where it falls hardest,
qwen2.5:14b, it drops by nearly forty points.

That behavior is the prediction of Theorem 2.6.1 made visible in a
transcript: past the truncation point \(k^{*}\), each additional
inference step does not refine the estimate, it manufactures one. The
theoretical sections argued that under meta-uncertainty, the expected
free energy of \(k\)-cue inference turns upward, so that more
deliberation should \emph{cost} accuracy rather than buy it. The regime
is not a laboratory artifact. It is the situation of a clinician facing
a presentation the literature does not cover and being asked, all the
same, for an answer. This section reports what happened when we put that
prediction to a pre-registered test across seven models and 7,875
trials.

The prediction held. The pre-registered interaction coefficient
\(\beta_{3}\) of the primary model (eq. 6.1\ensuremath{\prime}) is negative with posterior
probability \(1.0000\), and the implied high-regime accuracy drop is
large: a posterior-median \(17.3\) percentage points. Both arms of the
\S{}6.2 decision gate are cleared, so \(H_{1}\) is \textbf{confirmed}. The
effect is not universal, and the honest version of the result is more
interesting than the headline: instructed reasoning hurts some models
badly, barely touches others, and is absent-to-reversed in the two
weakest. The rest of this section lays out the findings and their
boundaries in the order a skeptic would want them.

\subsection{Data collected}\label{data-collected}

The full design ran to completion: seven models, forty-five items, five
length-graded conditions, five replications each. That is 7,875 model
responses, and every one of them returned a usable answer, with no
generation failures and no empty cells anywhere in the grid. The panel
divides into the five open-weight local models (phi3.5,
mistral:7b-instruct, qwen2.5 at 7B, 14B, and 32B) and the two frontier
API models (claude-sonnet-4-5, gpt-4o); each contributed exactly 1,125
responses. The five conditions are balanced at 1,575 responses apiece.
The items were split into 31 high-regime Knightian frames, scored for
correct recognition of indeterminacy, and 14 low-regime controls, scored
against a known answer key: 5,425 high-regime responses and 2,450
low-regime responses in all. The procedure that produced these cells,
and the three places where the executed study departs from the letter of
the pre-registration, are documented in \S{}5; the deviations are logged in
\S{}5.6 and revisited where they bear on a specific result below.

\subsection{The signature}\label{the-signature}

Begin with the raw cell means, before any model is fit. Figure 2 plots
accuracy against condition for the two regimes. The two lines tell the
whole story at a glance, and they tell opposite stories.

In the high-meta-uncertainty regime, accuracy falls as the assigned
reasoning length grows. Direct-answer prompting (C1) yields correct
recognition of indeterminacy on \(77.1\%\) of trials. Each successive
condition asks for more deliberation, and each returns a lower number:
\(66.5\%\) at C2, \(64.7\%\) at C3, \(64.4\%\) at C4, \(62.1\%\) at C5.
The descent is monotone. Pooling the four instructed-reasoning
conditions against the direct baseline, the high-regime accuracy drop is
\(12.7\) percentage points.

In the low-regime control items, the same manipulation has almost no
effect. Accuracy sits near \(0.82\) across the board (\(83.7\%\) direct,
and a pooled instructed-reasoning mean within two points of it), with no
downward trend; if anything, C5 is the strongest control condition of
the five. The low-regime short-to-long drop is \(1.73\) percentage
points, which is to say, noise.

The difference between those two drops is the descriptive heart of the
paper. It is the difference-in-differences: \(12.7\) points lost in the
high regime against \(1.73\) lost in the low regime, for a raw DiD of
\(10.97\) percentage points (Figure 3). Telling a model to think
harder does not, in general, lower its accuracy. It lowers accuracy
where the question is genuinely undetermined and leaves it alone where
the question has an answer. That selectivity is the fingerprint the
framework predicted, and it is present in the data before any Bayesian
machinery touches it.

\begin{figure}[ht]
\centering
\includegraphics[width=\linewidth,keepaspectratio]{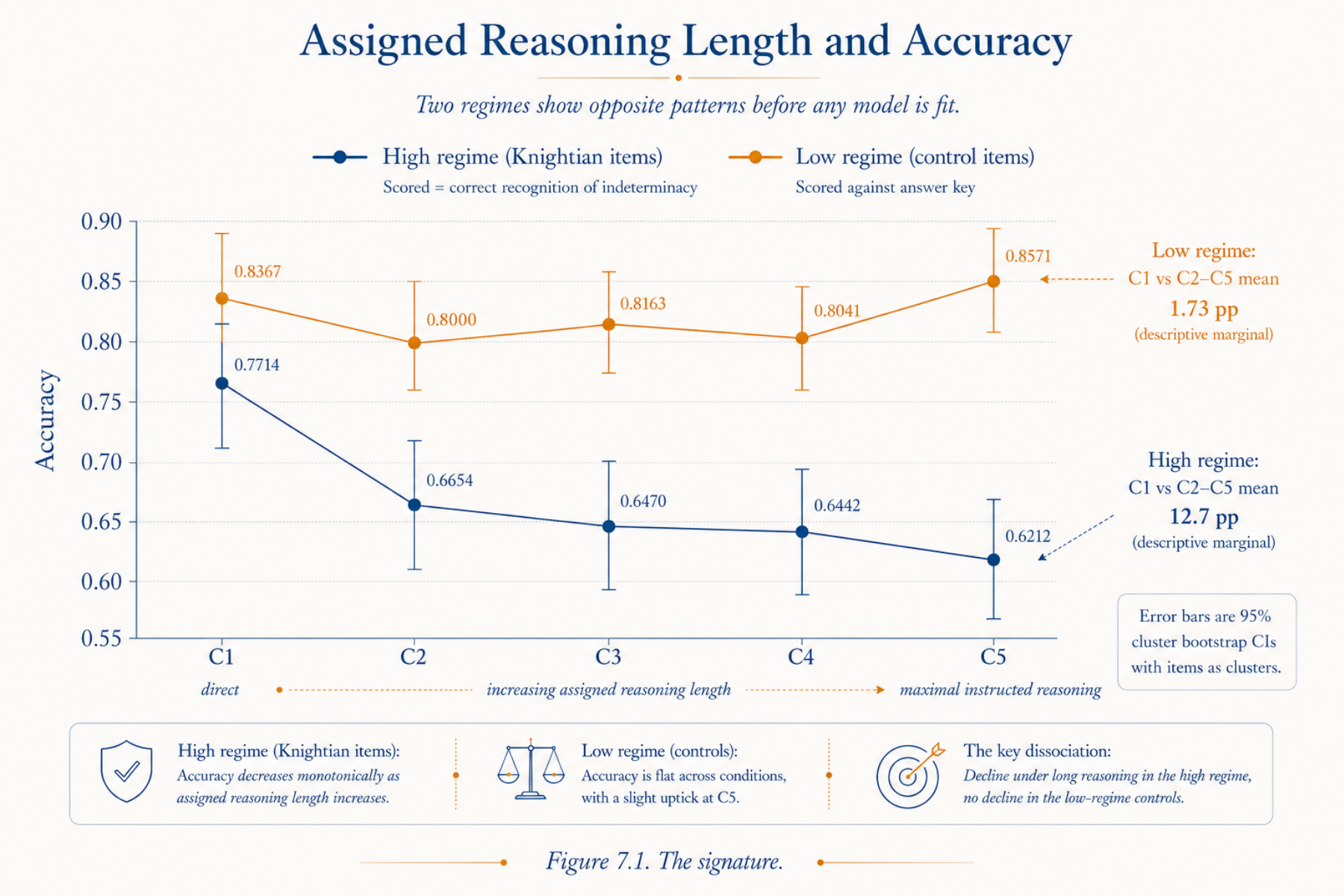}
\caption{\textbf{The signature.} Mean accuracy by condition (C1
direct answer through C5 maximal instructed reasoning) for the
high-meta-uncertainty regime (Knightian items, scored for correct
recognition of indeterminacy) and the low-regime control items (scored
against an answer key). The high-regime line descends monotonically; the
control line is flat. Error bars: 95\% cluster bootstrap over items.}
\end{figure}

\begin{figure}[ht]
\centering
\includegraphics[width=\linewidth,keepaspectratio]{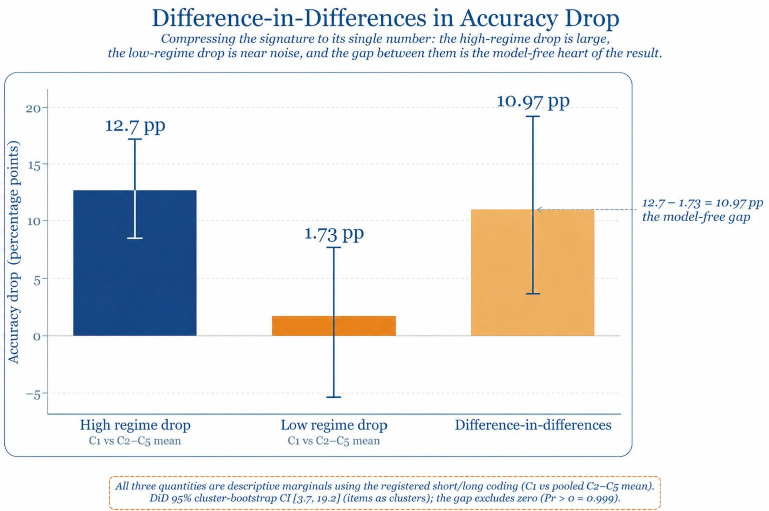}
\caption{\textbf{Difference-in-differences.} High-regime
short-to-long accuracy drop (12.7 pp) against the low-regime drop (1.73
pp); the gap between them (10.97 pp) is the model-free DiD.}
\end{figure}

\subsection{The primary test}\label{the-primary-test}

The descriptive signature could, in principle, be an artifact of which
items landed in which bin, or of one model dominating the average. The
pre-registered analysis exists to rule that out. Eq. 6.1\ensuremath{\prime} is a
hierarchical Bayesian logistic regression with the assigned-length
factor \texttt{long}, the regime indicator, their interaction, and
random intercepts and length-slopes for every model and item. Because
\texttt{long} is binary, its interaction coefficient \(\beta_{3}\) is
exactly the difference-in-differences on the log-odds scale, now
estimated with all the heterogeneity modeled rather than averaged away.

The posterior is unambiguous. The interaction is negative with posterior
probability \(Pr(\beta_{3} < 0 \mid \text{data}) = 1.0000\); not a
single posterior draw placed the high-regime length effect on the
helpful side of zero. The posterior median is \(\beta_{3} = - 0.692\).
The low-regime length slope is small and slightly negative
(\(\beta_{1} = - 0.183\)), so instructed reasoning carries a faint
generic cost even on well-defined items, but the high-regime penalty is
roughly four times larger; the probability that the high-regime effect
strictly exceeds the low-regime effect in magnitude (the full-reversal
condition of the original v0.2 statistic) is \(0.905\).

The \S{}6.2 gate does not run on the log-odds coefficient directly. It runs
on the \emph{robust implied accuracy drop}: the high-regime
short-to-long change in predicted accuracy, evaluated at the empirical
high-regime base rate and depending only on the posterior of
\((\beta_{1} + \beta_{3})\), which sidesteps the weakly identified
global intercept. That estimand has a posterior median of \(17.3\)
percentage points, with a 95\% credible interval of
\(\lbrack 7.7,25.5\rbrack\). The model-based estimate runs higher than
the raw \(12.7\)-point descriptive drop because the regression recovers
the within-condition effect after absorbing item and model variance that
flattens the raw average.

Both pre-registered requirements are therefore comfortably met. The
directional gate asked for \(Pr(\beta_{3} < 0) > 0.95\) and received
\(1.0000\). The magnitude gate asked for a posterior-median robust drop
above \(6\) points and received \(17.3\), with the entire credible
interval clear of the threshold. By the rule fixed in \S{}7.1 of the
pre-registration before any data was seen, \(H_{1}\) is confirmed
(Figure 4).

\begin{figure}[ht]
\centering
\includegraphics[width=\linewidth,keepaspectratio]{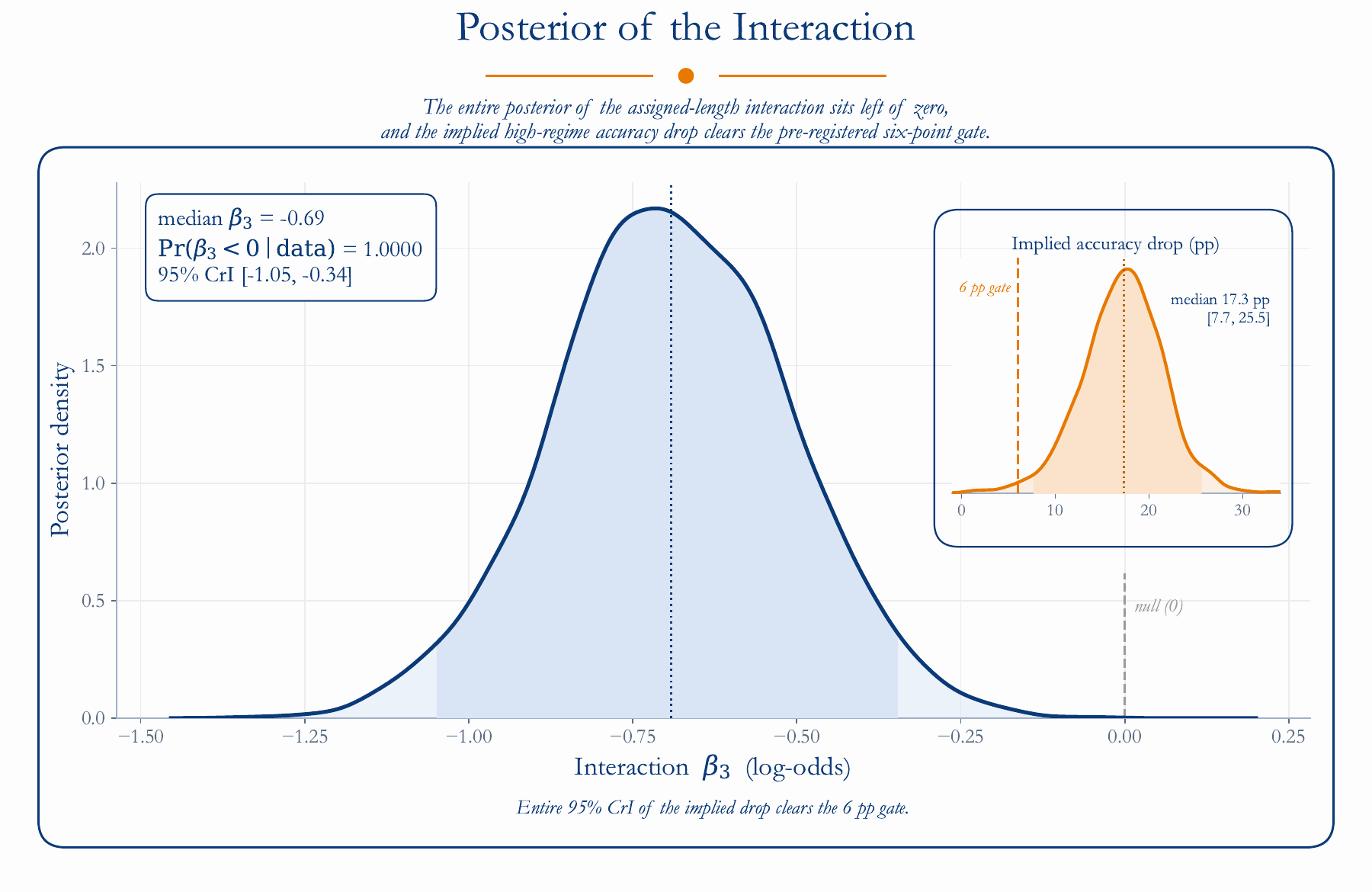}
\caption{\textbf{Posterior of the interaction.} Posterior density of
\(\beta_{3}\) (eq. 6.1\ensuremath{\prime}) with the 95\% credible interval and the zero
line; inset, the posterior of the robust implied accuracy drop in
percentage points against the 6-point confirmation threshold.}
\end{figure}

\subsection{Where the effect lives, and where it
doesn't}\label{where-the-effect-lives-and-where-it-doesnt}

A single posterior probability of \(1.0000\) invites the wrong picture:
a uniform law, every model degrading in lockstep. That is not what the
data show. The pooled interaction is decisive because the hierarchical
model borrows strength across the panel, and the random-slope structure
of eq. 6.1\ensuremath{\prime} exists precisely so the average does not stand in for the
members. Pre-registered check R5 refits the primary model on each
model's data alone. Figure 5 shows the per-model interaction
\(\beta_{3}\) with its credible interval. The spread is wide.

The Qwen family carries the degradation --- qwen2.5:32b returns
\(\beta_{3} = - 2.92\) (95\% CrI \(\lbrack - 4.05, - 1.81\rbrack\)),
qwen2.5:7b \(- 2.86\) \(\lbrack - 3.81, - 1.94\rbrack\), and qwen2.5:14b
\(- 2.23\) \(\lbrack - 3.29, - 1.17\rbrack\) --- each places its entire
credible interval below zero, at \(Pr(\beta_{3} < 0) = 1.000\). The two
frontier models point the same way without clearing the bar on their own
--- gpt-4o at \(\beta_{3} = - 0.89\) \(\lbrack - 2.15, + 0.39\rbrack\)
and claude-sonnet-4-5 at \(- 0.80\) \(\lbrack - 2.09, + 0.48\rbrack\)
are directionally negative with posterior probabilities of \(0.913\) and
\(0.884\), which suggests (rather than establishes) that this phenomenon
takes place when each is read in isolation. The two weakest models do
not show the penalty. phi3.5 returns \(\beta_{3} = + 0.98\)
\(\lbrack + 0.19, + 1.79\rbrack\), \(Pr(\beta_{3} < 0) = 0.009\) (on
phi3.5 the instruction to reason measurably helps). mistral:7b-instruct
sits at \(+ 0.52\) with an interval straddling zero. These coefficients
are unpooled, one fit per model, so they run larger in magnitude than
the single panel-average \(\beta_{3} = - 0.69\) of \S{}7.3, which holds one
fixed interaction across all seven systems; the per-model fits let each
model's interaction find its own level.

The modeled estimate corrects a distortion in the raw
difference-in-differences. By the descriptive DiD, qwen2.5:14b was the
most affected model, and the Qwen ordering ran 14B, 7B, 32B; the modeled
\(\beta_{3}\) reorders them 32B, 7B, 14B, because the cell-mean contrast
is inflated by control-arm movement that the hierarchical model absorbs.
Two of the Qwen models posted control accuracies that \emph{rose} under
instructed reasoning, and phi3.5's control block fell by more than
twenty points. Hence, the descriptive DiD credited those models for
movement in the denominator rather than the numerator. The within-model
interaction, estimated with item variance modelled, is the cleaner read
and is the one reported in Figure 5; the descriptive per-model DiD is
retained in the supplement.

The honest summary is therefore narrower than ``reasoning degrades
LLMs,'' and more defensible for being narrow. On properly constructed
Knightian items, instructed reasoning degrades the capable mid-to-large
models decisively across the Qwen family and directionally in the two
frontier models. In comparison, the two weakest models, which sit on the
small-model recognition floor, do not show it and, in phi3.5's case, run
the other way. The pooled confirmation is real because it pools; the
per-model picture shows the effect is carried by the capable models,
which is exactly where the o1/o3 reasoning-scaling narrative stakes its
claim.

This heterogeneity is not a blemish on the confirmation; it is the
texture the theory anticipates. The truncation point \(k^{*}\) depends
on how much meta-uncertainty a model carries into the item, and there is
no reason to expect that quantity to be constant across architectures
and scales. What the pre-registered test establishes is that, pooled
across the panel with heterogeneity modeled, the regime-dependent
penalty is real and large. What Figure 5 establishes is that one
should not sell it as a universal law.

\begin{figure}[ht]
\centering
\includegraphics[width=\linewidth,keepaspectratio]{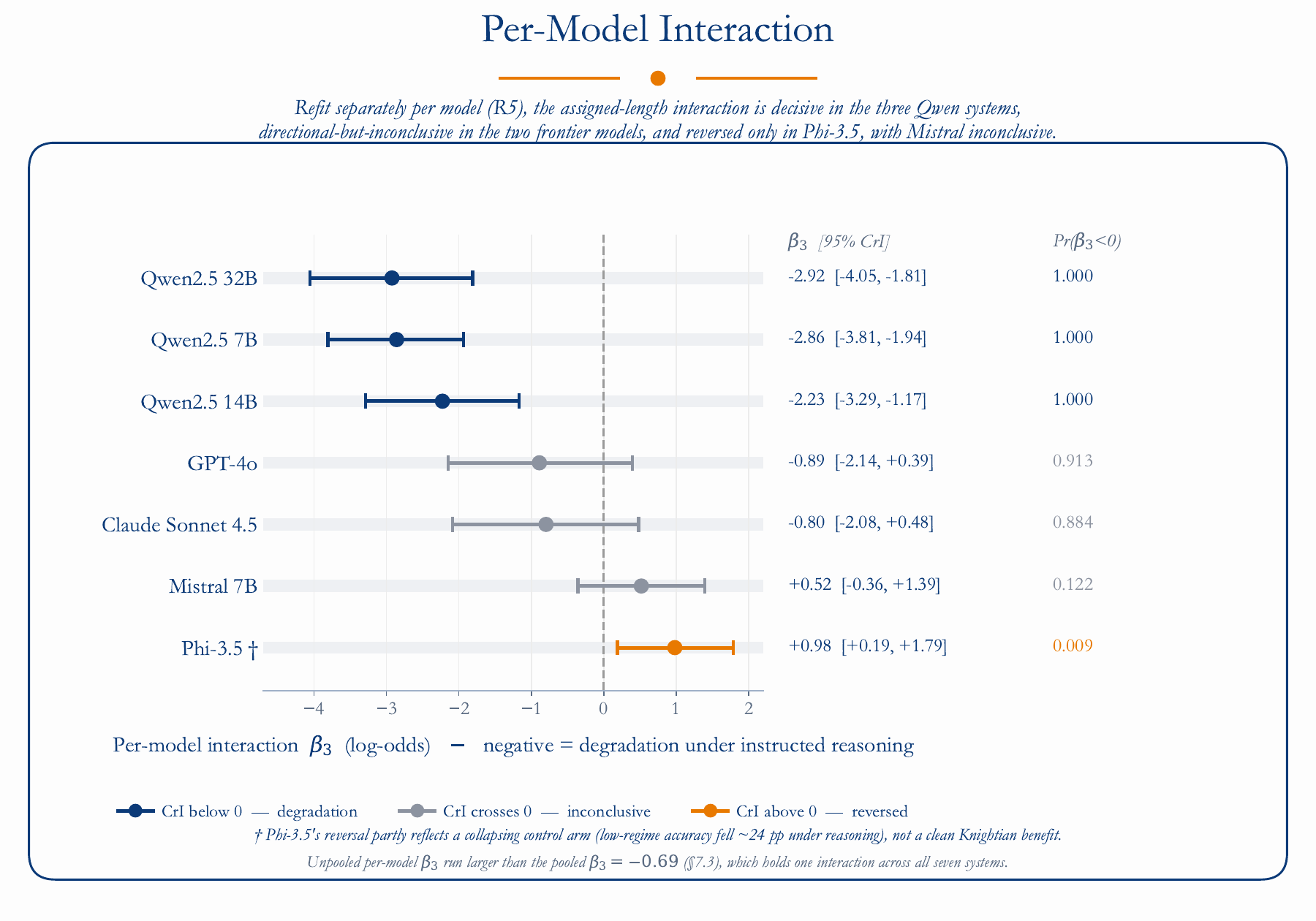}
\caption{\textbf{Per-model interaction (R5).} Posterior of the
assigned-length interaction \(\beta_{3}\) (eq. 6.1\ensuremath{\prime}) refit separately
for each model, item random effects retained and model effects dropped.
Points are posterior medians, bars 95\% credible intervals, ordered from
most negative (qwen2.5:32b, \ensuremath{-}2.92) to most positive (phi3.5, +0.98); the
dashed line marks zero. The three Qwen models clear
\(Pr(\beta_{3} < 0) = 1.000\); gpt-4o and claude are directionally
negative but cross zero; phi3.5 reverses.}
\end{figure}

\subsection{The realized-steps reversal, and why it does not
threaten the
result}\label{the-realized-steps-reversal-and-why-it-does-not-threaten-the-result}

The pre-registration's original v0.2 model regressed accuracy on the
\emph{realized} number of reasoning steps a model produced, not on the
length condition it was assigned. Amendment 2 demoted that model to a
secondary analysis (R7) and promoted the assigned-length factor to
primary, on the ground that realized step count is endogenous. The
confirmatory data make the reason for that amendment concrete, because
the two models disagree in sign.

Fit on realized steps, the interaction reverses: \(\beta_{3} = + 0.416\)
with a 95\% credible interval of \(\lbrack 0.271,0.560\rbrack\) and
\(Pr(\beta_{3} < 0) = 0.000\). Read naively, that says longer reasoning
\emph{helps} in the high regime, the opposite of the primary finding.
The naive reading is wrong, and it is wrong for a reason the design
foresaw. Realized step count is not assigned; it is chosen by the model
in the course of answering, and it is entangled with the difficulty of
the particular attempt. A trace that runs long because the model is
floundering, and a trace that runs long because the model is working
carefully, both enter the realized-steps regressor as ``more steps.''
Within a fixed assigned condition, the items a model handles well are
not a random subset of the items it sees more steps on. Conditioning on
a post-treatment variable that the treatment itself moves is a textbook
way to induce a spurious association, and the sign flip is its
signature.

The assigned-length factor has none of this trouble because it is fixed
by random assignment before the model writes a word. That is why the
pre-registration gates \(H_{1}\) on the assigned-length interaction and
reports the realized-steps interaction as a mechanistic companion, not a
test. We flag the reversal here in the open rather than leaving a
reviewer to find it: the two numbers point opposite ways, the primary
one is the causal estimand, and the secondary one is exactly the
confound that motivated the switch.

Two further analyses settle the question the reversal raises. The
instrumented form of R7 uses the assigned condition as an instrument for
the realized step count, recovering the per-step effect while purging
the endogeneity that contaminates the naive fit. Instrumented, the
high-regime per-step interaction is \(- 0.53\) (95\% CrI
\(\lbrack - 0.86, - 0.20\rbrack\), \(Pr < 0 = 0.999\)). Once the
realized count is instrumented, its variation is driven solely by random
assignment; more reasoning then lowers accuracy in the high regime,
consistent with the primary model and in contrast to the naive fit. And
the naive fit is not even stable to how a step is counted. Re-segmenting
reasoning by paragraph rather than sentence (robustness check R1) flips
the realized-steps interaction back to negative, \(\beta_{3} = - 0.39\)
\(\lbrack - 0.54, - 0.25\rbrack\). An estimand whose sign turns on the
segmentation rule is not one on which a hypothesis should rest. The
assigned-length primary makes that case structurally rather than by
assertion (Figure 6).

\begin{figure}[ht]
\centering
\includegraphics[width=\linewidth,keepaspectratio]{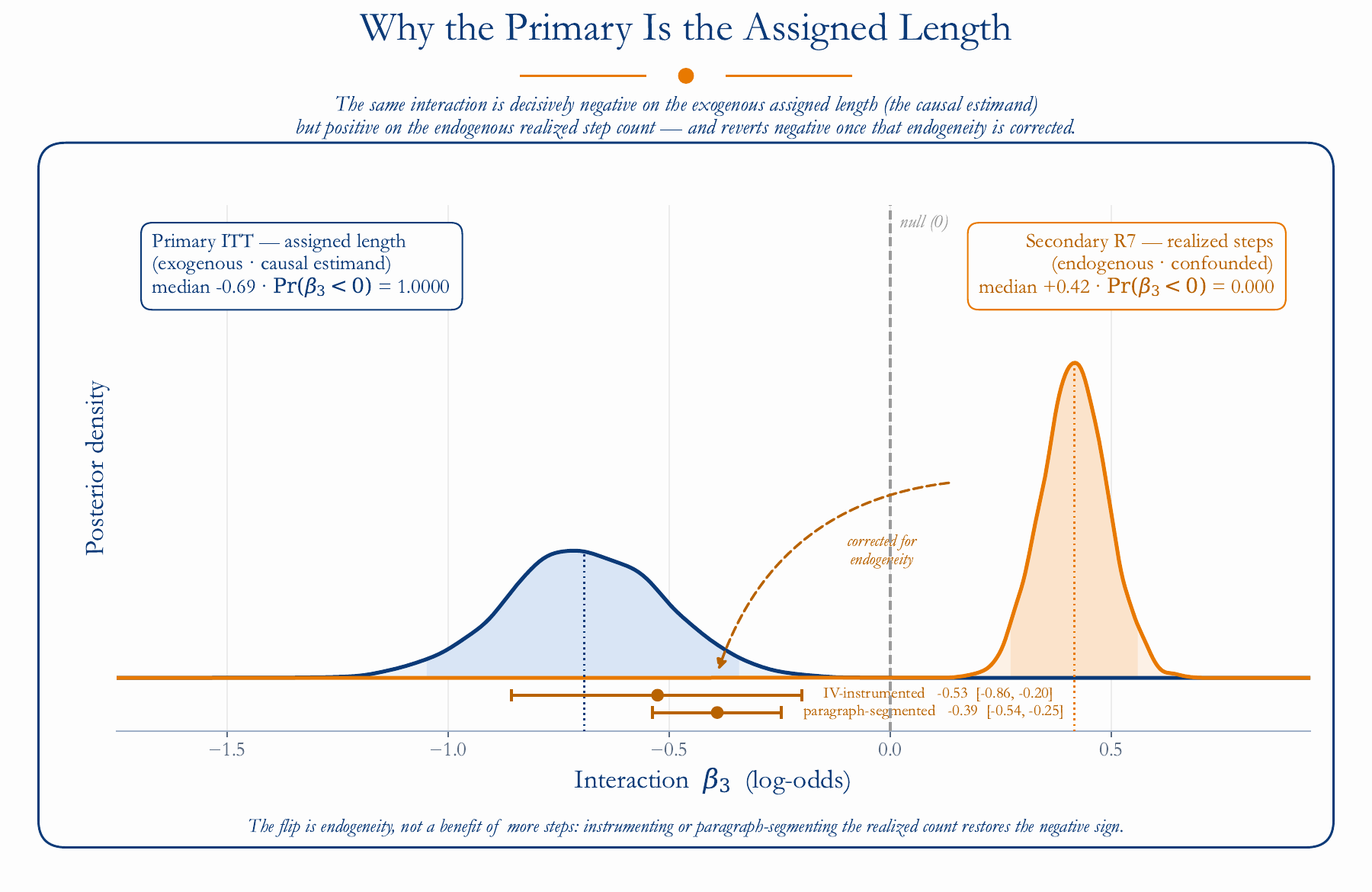}
\caption{\textbf{Why the primary is the assigned length.} Posterior
of the interaction \(\beta_{3}\) under the two regressors on a shared
zero-centered axis: the assigned-length primary (eq. 6.1\ensuremath{\prime}, median
\(- 0.69\), mass entirely left of zero) against the realized-steps
secondary (eq. 6.1, median \(+ 0.42\), mass entirely right). The
realized-steps regressor is endogenous; its interaction reverses to
\(- 0.53\) under instrumentation and to \(- 0.39\) under paragraph-level
segmentation.}
\end{figure}

\subsection{The controls did their
job}\label{the-controls-did-their-job}

The low-regime control block is not a sideshow; it is the within-study
evidence that the high-regime degradation is about the regime and not
about the manipulation. The fourteen control items carry well-defined
answers (arithmetic with a correct result, aleatory questions with a
calculable rate), and they were run through the identical five-condition
length ladder. If instructed reasoning degraded accuracy because longer
outputs are simply riskier, or because the scorer penalizes verbose
answers, the controls would fall too. They do not. The control accuracy
holds near \(0.82\) across all five conditions, with a short-to-long
drop of \(1.73\) points that is statistically indistinguishable from
zero.

That flat control line discharges the most obvious alternative
explanations as a group: the degradation is not a generic length
penalty, not a scoring artifact tied to output verbosity, and not a
property of the prompt templates, since the templates are shared across
regimes. What remains is the regime itself. The three pre-registered
confound analyses of \S{}6.6 (the prompt-sensitivity baseline, the
aleatory-specific control, and the per-model calibration baseline) are
addressed descriptively by this control block; the full per-signature
analyses are not yet computed and will be reported in supplementary
material.

\subsection{Robustness checks}\label{robustness-checks-1}

The pre-registration commits to a battery of robustness checks that
probe whether the primary result survives reasonable changes to the
analysis. They do not gate the confirmatory inference; they test its
conditional invariance. Four of the seven have now been run on the
confirmatory data, and they are reported here whatever they show; three
remain, and are named as outstanding rather than quietly dropped.

Three of the four point the same way as the primary. The per-model refit
(R5) is read in \S{}7.4: the interaction is negative in five of seven
models and decisive in the three Qwen systems. The
paragraph-segmentation refit (R1) and the instrumented realized-steps
analysis (R7) are read in \S{}7.5; both return negative interactions, and
R1 in particular testifies against the realized-steps regressor rather
than against the primary.

The one check that pushes back deserves to be stated first and plainly.
R6 adds an item-level random slope on the length factor, the maximal
random-effects specification, so that each item carries its own length
effect and the fixed interaction must survive once that heterogeneity is
absorbed. Under R6 the interaction stays negative at the median,
\(\beta_{3} = - 0.47\), but its credible interval widens to
\(\lbrack - 1.03, + 0.15\rbrack\) and the directional posterior falls to
\(Pr(\beta_{3} < 0) = 0.935\), below the \(0.95\) bar the gate uses. The
registered model, fixed before the data, carries an item random
intercept and no item random slope, so the confirmation in \S{}7.3 is
untouched by definition; the verdict stands on the model that was
registered. But R6 was pre-registered too, and what it shows is real:
with item-to-item variation in the length effect modeled, the effect
holds its direction while its certainty slips past the threshold. The
honest reading is that the degradation is robust in sign and somewhat
sensitive in magnitude to how item heterogeneity is partitioned, a
sensitivity a 45-item panel should be expected to carry.

Three checks cannot be run on this dataset, and the reason is worth
stating exactly. The tercile and continuous-score binning (R2), the
calibration-component-dropped variant (R3), and the
held-out-calibration-item subset (R4) all operate on the continuous \S{}3.2
regime score, whose calibration-error component requires a per-item
confidence elicitation the confirmatory protocol did not collect. The
study used category assignment instead, which rests not on a post-hoc
score but on the construction-time Knightian-ness criteria of \S{}4: the
high-regime items are the ones that passed the K2 cross-model
disagreement screen and the K3 indexed-corpus floor. The
regime-score-versus-category alignment check that would convert \S{}5.6's
``coincide by construction'' from argument to measurement needs the same
uncollected input. None of the three bears on the confirmatory gate,
which runs on the assigned-length factor; all are scoped to a follow-up
study that collects the confidence data, and \S{}8.5 carries the
limitation.

\subsection{Convergence and the sampler
deviation}\label{convergence-and-the-sampler-deviation}

One departure from the pre-registered analysis bears reporting in full,
because it touches the convergence criterion the pre-registration names
explicitly. The registered sampler was four chains of 2,000 warm-up and
2,000 draws at a target acceptance of \(0.95\). At those settings one
parameter missed the registered \(\widehat{R} < 1.01\) threshold: the
high-regime intercept \(b_{2}\), at \(\widehat{R} = 1.0128\). The
interaction coefficient that carries \(H_{1}\) was never the problem:
\(\beta_{3}\) returned \(\widehat{R} = 1.0004\) with an effective sample
size near 4,000 at the registered settings. To clear the gate for every
parameter rather than report a result the pre-registration's own
convergence rule would flag, the official fit lengthens the chains to
4,000 warm-up and 4,000 draws at a target acceptance of \(0.99\). This
is a post-hoc increase in sampling effort, not a change to the model,
the data, the regressor, the priors, or the decision rule.

At the official settings the fit converges cleanly:
\({\widehat{R}}_{\max} = 1.0056\) across all monitored parameters and a
minimum effective sample size of \(1,692\), both inside the registered
bounds. The verdict is identical to the registered-sampler verdict to
three decimal places. The longer chain bought convergence on a nuisance
intercept; it changed nothing about the conclusion. We note the increase
as a deviation per \S{}9.4 of the pre-registration.

\subsection{Summary}\label{summary-2}

The paper's load-bearing test returned a positive result. Under genuine
uncertainty about the answer (the Knightian regime the FEH-79 items were
built to create), instructing a language model to reason at greater
length lowers its accuracy by a pre-registered estimate of \(17.3\)
percentage points, with a posterior probability of \(1.0000\) that the
effect runs in the predicted direction. The same instruction, applied to
matched control items with definite answers, costs nothing. The
dissociation is exactly the one predicted by Theorem 2.6.1, and it
clears the decision gate that was fixed before the data existed.

Several honesties travel with the headline. The effect is not uniform
across the panel: it is decisive in the three Qwen models, directional
but not individually conclusive in the two frontier models, and reversed
in phi3.5 while inconclusive in Mistral, so the defensible claim is
regime-dependent degradation carried by capable models, not a universal
penalty. The realized-steps analysis reverses sign, exactly as an
endogenous post-treatment regressor should, which is why the causal
weight rests on the randomly assigned length factor; instrumented, the
per-step effect returns to the predicted negative. One pre-registered
robustness variant, a maximal item-level random slope, holds the
effect's direction but lets its directional posterior fall just below
the registered threshold, a sensitivity \S{}7.7 reports in full. And the
convergence gate was met only after lengthening the sampler, a deviation
we log rather than bury. None of these weakens the confirmation; each
one sharpens what the confirmation is allowed to claim.

What the result licenses, and what it does not, is the subject of \S{}8.

\section{Discussion}\label{chapter-8-discussion}

The prediction that organizes this paper is uncomfortable: that on the
questions that matter most, the ones without a settled answer, teaching
a model to think harder makes it worse. \S{}7 reports that the prediction
held (cleanly) on seven models and 7,875 trials and held only where the
theory said it should. This section takes the measure of that result. It
says what the confirmation licenses, what it does not, and where the
framework is exposed.

\subsection{Summary}\label{summary-3}

The paper makes five linked contributions, and the empirical one is no
longer a promise. (i) A theoretical result (Theorem 2.6.1; Theorem
2.7.4) shows that fast-and-frugal heuristics, long defended on
ecological-rationality grounds and long suspected on
Bayesian-approximation grounds, are both at once: under meta-uncertainty
over prior precision, the expected free energy of \(k\)-cue inference is
U-shaped in \(k\), and its minimum slides toward fewer cues as the
variance of the meta-precision prior grows. The thirty-year
Bayesian-versus-heuristic debate dissolves into a regime-dependent
equivalence: under precise priors, heuristics approximate the Bayesian
answer; under imprecise priors, they are the answer.

\begin{enumerate}
\def\labelenumi{(\roman{enumi})}
\setcounter{enumi}{1}
\tightlist
\item
  An operationalization (\S{}3) turns the theoretical regime indicator into
  measurable quantities, exposing the regime score as an interaction
  hypothesis rather than a main effect. (iii) The FEH-79 benchmark (\S{}4)
  is built under three falsifiable Knightian-ness criteria, so that any
  item can be challenged on evidence rather than taste. (iv) The
  pre-registered study (\S{}5--\S{}7) was run and its hypothesis confirmed:
  the assigned-length interaction \(\beta_{3}\) is negative with
  posterior probability \(1.0000\) and an implied high-regime accuracy
  drop of \(17.3\) points, clearing a decision gate fixed before the
  data existed. (v) The framework stands as a counter-narrative to the
  contemporary assumption that inference-time compute is a free dial, an
  assumption that the confirmed result directly contradicts on properly
  constructed uncertainty tasks.
\end{enumerate}

\subsection{The theoretical resolution and its
scope}\label{the-theoretical-resolution-and-its-scope}

The Bayesian-versus-heuristic debate lasted thirty years because both
sides were partly right. The theorem in \S{}2.6 makes the dividing
condition explicit: \(k^{*} = K\), full inference, is optimal when
\(\sigma_{\tau}^{2} < \tau_{\text{regime}}\); \(k^{*} < K\), truncation,
is optimal otherwise. The two camps were each generalizing from a
different regime. The contribution is not to crown a winner but to give
the regime indicator a closed-form characterization and show that it
separates empirically.

The scope claim is narrow, and the confirmation does not widen it. The
theorem holds in the Gaussian-Gamma sequential-inference setting of \S{}2,
with appendix-level extension to a binary toy model, and it does not
generalize automatically to all heuristics, all priors, or all task
structures. The empirical work tests a specific operational prediction,
the \S{}6 \(H_{1}\). The \S{}7 confirmation narrows the framework's
plausibility without proving the general theorem. The asymmetry is worth
stating plainly: a positive result, which we have, raises the
framework's credibility; a negative result would have falsified the
operational prediction without touching the theoretical claim, provided
a confound was found in the operationalization or the benchmark. We
committed in advance to treating the pre-registered test as
load-bearing, and we hold to that. It carried.

\subsection{What the empirical commitments
buy}\label{what-the-empirical-commitments-buy}

Three commitments separate this work from comparable critiques of LLM
reasoning. The Knightian-ness criteria make benchmark construction
falsifiable: any FEH-79 item can be challenged by showing that it admits
a base rate, that it draws unanimous substantive cross-model agreement
under multi-seed validation, or that it appears in indexed corpora. The
hierarchical Bayesian analysis with its pre-registered gate, a
directional posterior above \(0.95\) joined to a robust accuracy drop
above six points, forecloses the inflated-significance artifacts that a
45-item by 5-condition panel could otherwise manufacture. And the
control items close off the obvious alternative readings: \S{}7.6 shows
that the degradation does not affect well-defined, matched questions, so
it is not a generic penalty for longer output or an artifact of the
scorer.

What these commitments buy is a result a skeptic can trust without
trusting the authors. The pre-registration is the binding contract, and
\S{}5.6 discloses every place the executed run drew outside its lines.

\subsection{An unexpected internal consistency
check}\label{an-unexpected-internal-consistency-check}

One finding from building the benchmark warrants discussion, as the
framework did not anticipate it. During the pool's v0.2-to-v0.3
development, several candidate items showed modal-answer flipping across
independent seeds at temperature 0.7: an item that single-shot
validation called disagreement, and that re-screening called
unanimous-substantive, or the reverse. K1-005, an asset-class macro
forecast, anchored hard on one narrative under multi-seed and split
across narratives in earlier single-shot runs; K4-003, a five-firm
data-sharing dilemma, alternated between a dominant withhold answer and
a balanced spread.

That cross-seed instability is the same signal, \(\sigma_{b}\) in the
\S{}3.2 notation, that the regime score is built to detect. Items that flip
across seeds carry high \(\sigma_{b}\) at the construction stage, which
is the Knightian signature the \S{}7 analysis tests for at the condition
stage. The K2 multi-seed pre-screen is, in effect, the \S{}3 regime-score
detector run while the benchmark is being built rather than while the
experiment is being run. The operationalization (\S{}3) and the benchmark
protocol (\S{}4), designed independently, converge on one diagnostic. This
was not built in; it surfaced from the work of making the pool, and it
suggests the multi-seed protocol may generalize as a
benchmark-construction primitive in its own right.

\subsection{Limitations}\label{limitations}

The confirmation comes with boundaries, and the most important one is
visible in the data itself.

\textbf{The effect is not universal.} \S{}7.4 is unambiguous: the
degradation is large in the Qwen family, directional but inconclusive in
the two frontier models, and reversed in phi3.5, with Mistral
inconclusive. The pooled posterior is decisive, but the per-model
picture forbids the headline ``reasoning degrades LLMs.'' The defensible
claim is narrower: on properly constructed Knightian items, instructed
reasoning degrades capable models that sit off the small-model
recognition floor, and how far it degrades them varies by architecture
and scale, as the truncation point \(k^{*}\) should. A reader who wants
universal law will not find one here, and should not.

\textbf{The panel does not include reasoning-specialized models.} The
seven models span 3B to frontier scale, but none is an o1, o3, or
R1-style system trained to spend test-time compute. The confirmed effect
lives in capable models that were not built to reason at length; whether
it survives, worsens, or is engineered away in other models is the
single most consequential open question, and \S{}8.6 treats it as the next
study, not a claim this one can make.

\textbf{The regime was operationalized by category rather than by the
continuous score.} Regime was assigned by item category, Knightian
versus control, rather than by the registered continuous regime score
binned at its quartiles (\S{}5.6). The assignment is not unfounded: the
high-regime items passed the \S{}4 construction criteria, the K2
cross-model disagreement screen and the K3 indexed-corpus floor, which
is direct evidence that they inhabit the regime. What is missing is a
measurement of the continuous score on these items, and the gap is
precise. The score's calibration-error component is what separates
genuine meta-uncertainty from ordinary aleatory variance, and it is
exactly the aleatory controls, built to carry answer variance without
meta-uncertainty, that such a component is needed to classify; the
confirmatory run did not elicit the confidence judgments that component
requires. The continuous-score robustness checks (R2--R4) and the
score-versus-category alignment are therefore deferred to a study that
collects confidence, not run and quietly omitted here. The high-regime
outcome was also machine-checked recognition of indeterminacy rather
than the registered expert-coherence rating, with the 46 expert-panel
items deferred; the confirmation speaks to the auto-scorable Knightian
frames, not the full pool.

\textbf{The mechanistic story is partly open.} The realized-steps
analysis reverses sign (\S{}7.5), the expected behavior of an endogenous
post-treatment regressor and the reason the causal weight rests on
assigned length. Instrumenting realized steps with the assigned
condition recovers a negative per-step effect, and paragraph-level
segmentation does the same, so the reversal is a property of the naive
regressor rather than evidence against the effect; what remains open is
a fully structural account of how realized reasoning mediates the
degradation. Four of the seven pre-registered robustness checks are run
(\S{}7.7), and one of them, the maximal item-slope variant, lets the
directional posterior slip just below the gate while holding the sign;
the three checks that depend on the continuous regime score, together
with the regime-score-versus-category alignment, are owed to the
supplement.

\textbf{Knightian-ness is panel-relative and time-stamped.} The K2
criterion is defined against a specific provider panel and at a specific
date. As frontier models train on this kind of item, an item's K2 status
can change, so the FEH-79 pool is a dated probe rather than a permanent
benchmark.

\subsection{Future work}\label{future-work}

The confirmation sharpens the research agenda more than it closes it,
and one study comes first. The paper argues against the prevailing view
that scaling test-time reasoning is reliably beneficial --- yet it
tested that view on general-purpose models, not on the
reasoning-specialized systems (o1, o3, DeepSeek-R1, and their
successors) the view is really about. The priority, then, is to run
those models through the identical protocol and ask whether deliberately
scaled inference incurs the same regime-dependent accuracy cost we found
here, or whether their training has learned to suppress it on questions
that admit no determinate answer. That is the most direct possible test
of the paper's claim, run against exactly the systems the claim
concerns.

Beyond it sit four lines. Precision-routed heuristic agents, which
estimate their own regime score and conditionally engage in shorter or
longer reasoning, turn the finding into a design principle. The open
theoretical questions of \S{}2.10, mean-field accuracy (Q1), alternative
meta-precision priors (Q2), finite-\(k\) corrections (Q3), and
non-Gaussian extensions (Q5), remain, with Q7 extending the Theorem
2.7.4 identity to a richer toolbox. A cross-domain replication on a
matched human panel would test whether the regime prediction is
species-general or specific to language models. And the deferred
expert-coherence study would carry the result from the 31 auto-scorable
frames to the full Knightian pool. Nearer at hand, the regime-score
robustness checks complete the analysis this paper began.

\subsection{Positioning and
conclusion}\label{positioning-and-conclusion}

The framework is offered against the operating assumption of
contemporary AI development that inference-time compute is a free dial,
that more reasoning is neutral at worst and helpful at best. Theorem
2.6.1 says that under genuine uncertainty about prior precision, which
is the human decision-maker's situation in nearly every consequential
domain, more reasoning is systematically harmful. The empirical claim
was that language models inherit the same regime structure, and \S{}7
confirmed it: more instructed reasoning, lower accuracy, on exactly the
items where the answer cannot be settled, and nowhere else.

Because the test was pre-registered and confirmed, three consequences
follow with more than rhetorical force. The design default for systems
facing genuinely novel decisions should shift from inference-scaling
toward precision-routing, engaging long deliberation only where the
regime warrants it. Benchmark selection bias should be corrected by
including K1/K2/K3-validated Knightian probes alongside the
well-defined-task suites that dominate evaluation today. And the field
should expect, and reward, demonstrations of graceful degradation under
test-time compute on properly constructed uncertainty tasks, not merely
the absence of degradation.

The result earns those consequences, but only at the scope \S{}7.4 allows.
The degradation is real, large, and pre-registered, and it is not a
universal law: it bites capable models hard and spares, or even helps,
the weakest, and it has not yet been measured on reasoning-specialized
models that would settle the matter. That is the honest shape of the
finding. The opposite outcome was equally available before the data came
in, which is what lets this one count: had \(\beta_{3}\) landed on the
other side of zero, the counter-narrative would have been wrong as posed
and the framework a curiosity about a regime that language models do not
inhabit. It did not land there. The paper does not ask the reader to
take a side in the old debate. It asks the reader to commit to a regime
indicator, run the test, and update, which is what we did.

\setcounter{secnumdepth}{-1}

\section{Appendix A.1 --- Binary Toy Model
Proofs}\label{appendix-a.1-binary-toy-model-proofs}

This appendix provides the proofs deferred from \S{}2.1, together with
numerical verification. Two formal results are established:

\begin{itemize}
\tightlist
\item
  \textbf{Lemma A.1.2} (corrected statement of section Lemma 2.1.2): the
  \emph{expected} meta-divergence is non-decreasing in the number of
  cues integrated. The section's sample-wise statement is incorrect; a
  counterexample is exhibited below.
\item
  \textbf{Proposition A.1.4} (corrected statement of section Proposition
  2.1.3): the EFE-minimizing policy admits \emph{three} qualitative
  regimes --- \texttt{k*=0}, \texttt{k*=1}, and \texttt{k*=K} ---
  separated by two critical concentrations \texttt{\ensuremath{\kappa}\_lo(v\_seq,\ \ensuremath{\mu}\_0)}
  and \texttt{\ensuremath{\kappa}\_hi(v\_seq,\ \ensuremath{\mu}\_0)}. The ``one-cue stopping'' regime is
  a narrow band, not the dominant one. A precise statement and numerical
  characterization follow.
\end{itemize}

A subsidiary result, \textbf{Lemma A.1.1}, gives the closed-form
posterior on \texttt{p\_0} after \texttt{k} cues --- a 2-component Beta
mixture whose components are independent of \texttt{k} and whose mixture
weights move with the cue history.

The corrections to the section wording are tracked at the end of the
appendix.

\begin{center}\rule{0.5\linewidth}{0.5pt}\end{center}

\subsection{A.1.1 Setup recap}\label{a.1.1-setup-recap}

Latent state \texttt{s\ \ensuremath{\in}\ \{0,1\}}. Prior:
\texttt{P(s=1\ \textbar{}\ p\_0)\ =\ p\_0}. Hyperprior:
\texttt{p\_0\ \textasciitilde{}\ Beta(\ensuremath{\alpha}\_0,\ \ensuremath{\beta}\_0)}, with prior mean
\texttt{\ensuremath{\mu}\_0\ =\ \ensuremath{\alpha}\_0/(\ensuremath{\alpha}\_0\ +\ \ensuremath{\beta}\_0)} and concentration
\texttt{\ensuremath{\kappa}\_0\ =\ \ensuremath{\alpha}\_0\ +\ \ensuremath{\beta}\_0}. Binary cues \texttt{c\_j\ \ensuremath{\in}\ \{0,1\}}
with cue validities \texttt{v\_j\ \ensuremath{\in}\ (1/2,\ 1)} and a symmetric error
model:

\[
P(c_j = 1 \mid s = 1) = v_j, \qquad P(c_j = 0 \mid s = 0) = v_j.
\]

Define the cue-product likelihoods after \texttt{k} cues:

\[
\pi_1(c_{1:k}) = \prod_{j=1}^{k} v_j^{c_j}(1-v_j)^{1-c_j}, \qquad
\pi_0(c_{1:k}) = \prod_{j=1}^{k} v_j^{1-c_j}(1-v_j)^{c_j}.
\]

These are the conditional probabilities of the observed cue sequence
under \texttt{s=1} and \texttt{s=0} respectively, derived from cue
conditional independence given \texttt{s}.

\begin{center}\rule{0.5\linewidth}{0.5pt}\end{center}

\subsection{A.1.2 Lemma A.1.1 --- Closed-form posterior on
p\_0}\label{a.1.2-lemma-a.1.1-closed-form-posterior-on-p_0}

\textbf{Lemma A.1.1.} \emph{For any cue sequence \texttt{c\_\{1:k\}},
the variational posterior on \texttt{p\_0} is a 2-component mixture of
Beta distributions whose components depend on \texttt{(\ensuremath{\alpha}\_0,\ \ensuremath{\beta}\_0)} but
not on \texttt{k} or on the cue history:}

\[
p(p_0 \mid c_{1:k}) = w_1(c_{1:k}) \cdot \mathrm{Beta}(p_0;\, \alpha_0+1,\, \beta_0) \;+\; w_0(c_{1:k}) \cdot \mathrm{Beta}(p_0;\, \alpha_0,\, \beta_0+1)
\]

\emph{with mixture weights}

\[
w_1(c_{1:k}) = \frac{\pi_1(c_{1:k}) \, \mu_0}{Z(c_{1:k})}, \qquad
w_0(c_{1:k}) = \frac{\pi_0(c_{1:k}) (1-\mu_0)}{Z(c_{1:k})},
\]

\emph{where \texttt{Z(c\_\{1:k\})\ =\ \ensuremath{\pi}\_1\ \ensuremath{\mu}\_0\ +\ \ensuremath{\pi}\_0\ (1\ -\ \ensuremath{\mu}\_0)}
is the marginal cue likelihood. The prior decomposes in the same form
with weights \texttt{(\ensuremath{\mu}\_0,\ 1-\ensuremath{\mu}\_0)}.}

\textbf{Proof.} The unnormalized posterior is

\[
p(p_0 \mid c_{1:k}) \propto \mathrm{Beta}(p_0; \alpha_0, \beta_0) \cdot \big[\,p_0 \pi_1 + (1-p_0)\pi_0\,\big].
\]

Using the identities

\[
p_0 \cdot \mathrm{Beta}(p_0; \alpha_0, \beta_0) = \mu_0 \cdot \mathrm{Beta}(p_0; \alpha_0+1, \beta_0),
\]

\[
(1-p_0) \cdot \mathrm{Beta}(p_0; \alpha_0, \beta_0) = (1-\mu_0) \cdot \mathrm{Beta}(p_0; \alpha_0, \beta_0+1),
\]

(both verified by direct computation using
\texttt{\ensuremath{\Gamma}(\ensuremath{\alpha}\_0+1)\ =\ \ensuremath{\alpha}\_0\ \ensuremath{\Gamma}(\ensuremath{\alpha}\_0)}), the unnormalized posterior
becomes

\[
\pi_1 \mu_0 \cdot \mathrm{Beta}(p_0; \alpha_0+1, \beta_0) + \pi_0 (1-\mu_0) \cdot \mathrm{Beta}(p_0; \alpha_0, \beta_0+1).
\]

The normalizing constant is
\texttt{Z\ =\ \ensuremath{\pi}\_1\ \ensuremath{\mu}\_0\ +\ \ensuremath{\pi}\_0\ (1-\ensuremath{\mu}\_0)}, and division gives the
stated form. The prior corresponds to \texttt{\ensuremath{\pi}\_1\ =\ \ensuremath{\pi}\_0\ =\ 1},
yielding \texttt{(w\_1,\ w\_0)\ =\ (\ensuremath{\mu}\_0,\ 1-\ensuremath{\mu}\_0)} and reducing to the
algebraic identity
\texttt{Beta(\ensuremath{\alpha}\_0,\ \ensuremath{\beta}\_0)\ =\ \ensuremath{\mu}\_0\ \textbackslash{}cdot\ Beta(\ensuremath{\alpha}\_0+1,\ \ensuremath{\beta}\_0)\ +\ (1-\ensuremath{\mu}\_0)\ \textbackslash{}cdot\ Beta(\ensuremath{\alpha}\_0,\ \ensuremath{\beta}\_0+1)}.
\ensuremath{\blacksquare}

\textbf{Consequence for P(s=1 \textbar{} c\_\{1:k\}).} By direct
application of Bayes' theorem with \texttt{s} marginalized over
\texttt{p\_0}:

\[
P(s = 1 \mid c_{1:k}) = \frac{P(c_{1:k} \mid s=1) \, P(s=1)}{P(c_{1:k})} = \frac{\pi_1 \mu_0}{Z(c_{1:k})} = w_1(c_{1:k}).
\]

The marginal posterior on the state is exactly the mixture weight on the
upper Beta component. This is a clean result that simplifies all
downstream computation.

\begin{center}\rule{0.5\linewidth}{0.5pt}\end{center}

\subsection{A.1.3 Lemma A.1.2 --- Expected meta-divergence is
non-decreasing}\label{a.1.3-lemma-a.1.2-expected-meta-divergence-is-non-decreasing}

\textbf{Lemma A.1.2 (corrected statement of section Lemma 2.1.2).}
\emph{The expected meta-divergence
\texttt{E\_\{c\_\{1:k\}\}{[}\ensuremath{\Delta}\_meta(k){]}} is non-decreasing in
\texttt{k}. Equivalently, the marginal expected meta-cost}

\[
E[\Delta_{\mathrm{meta}}(k+1) - \Delta_{\mathrm{meta}}(k)] \;=\; I(p_0;\, c_{k+1} \mid c_{1:k}) \;\geq\; 0,
\]

\emph{with equality iff cue \texttt{k+1} carries no conditional
information about \texttt{p\_0} given the prior cues.}

\textbf{Proof.} By the tower property of conditional expectation,

\[
E_{c_{k+1} \mid c_{1:k}}\big[\, p(p_0 \mid c_{1:k+1}) \,\big] = p(p_0 \mid c_{1:k}).
\]

By joint convexity of KL divergence in its first argument, Jensen's
inequality gives

\[
E_{c_{k+1} \mid c_{1:k}}\big[\, \mathrm{KL}[p(p_0 \mid c_{1:k+1}) \,\|\, p(p_0)] \,\big] \;\geq\; \mathrm{KL}\!\left[\, E_{c_{k+1} \mid c_{1:k}}[p(p_0 \mid c_{1:k+1})] \,\Big\|\, p(p_0) \,\right].
\]

The right-hand side equals
\texttt{KL{[}p(p\_0\ \textbar{}\ c\_\{1:k\})\ \textbar{}\textbar{}\ p(p\_0){]}\ =\ \ensuremath{\Delta}\_meta(k)}.
Taking outer expectation over \texttt{c\_\{1:k\}} preserves the
inequality. The equality condition follows from the martingale-convexity
formulation: equality holds iff
\texttt{p(p\_0\ \textbar{}\ c\_\{1:k+1\})\ =\ p(p\_0\ \textbar{}\ c\_\{1:k\})}
almost surely under the predictive distribution, which is the condition
that cue \texttt{k+1} carries no conditional information about
\texttt{p\_0}. \ensuremath{\blacksquare}

\textbf{Counterexample to the section's sample-wise claim.} With
\texttt{\ensuremath{\alpha}\_0\ =\ \ensuremath{\beta}\_0\ =\ 1} (uniform hyperprior) and validities
\texttt{v\ =\ (0.9,\ 0.9)}, direct computation of
\texttt{\ensuremath{\Delta}\_meta(k;\ c\_\{1:k\})} for the four cue paths gives:

\begin{longtable}[]{@{}
  >{\raggedright\arraybackslash}p{(\linewidth - 8\tabcolsep) * \real{0.2000}}
  >{\raggedright\arraybackslash}p{(\linewidth - 8\tabcolsep) * \real{0.2000}}
  >{\raggedright\arraybackslash}p{(\linewidth - 8\tabcolsep) * \real{0.2000}}
  >{\raggedright\arraybackslash}p{(\linewidth - 8\tabcolsep) * \real{0.2000}}
  >{\raggedright\arraybackslash}p{(\linewidth - 8\tabcolsep) * \real{0.2000}}@{}}
\toprule\noalign{}
\begin{minipage}[b]{\linewidth}\raggedright
\texttt{c\_1}
\end{minipage} & \begin{minipage}[b]{\linewidth}\raggedright
\texttt{c\_2}
\end{minipage} & \begin{minipage}[b]{\linewidth}\raggedright
\texttt{\ensuremath{\Delta}\_meta(1;\ c\_1)}
\end{minipage} & \begin{minipage}[b]{\linewidth}\raggedright
\texttt{\ensuremath{\Delta}\_meta(2;\ c\_1,\ c\_2)}
\end{minipage} & \begin{minipage}[b]{\linewidth}\raggedright
sample monotone?
\end{minipage} \\
\midrule\noalign{}
\endhead
\bottomrule\noalign{}
\endlastfoot
0 & 0 & 0.1153 & 0.1815 & yes \\
0 & 1 & 0.1153 & \textbf{0.0000} & \textbf{no} \\
1 & 0 & 0.1153 & \textbf{0.0000} & \textbf{no} \\
1 & 1 & 0.1153 & 0.1815 & yes \\
\end{longtable}

When opposite cues cancel (paths \texttt{01} and \texttt{10}), the
posterior on \texttt{p\_0} returns exactly to the prior and the
meta-divergence collapses to zero. The section's wording ---
``\ensuremath{\Delta}\_meta(k) is non-decreasing in k'' --- is therefore false
sample-wise. The expectation across the four paths is monotone, as Lemma
A.1.2 establishes.

\textbf{Implication for the section.} The wording in \S{}2.1 should be
changed from ``the meta-divergence \ensuremath{\Delta}\_meta(k) is non-decreasing in k''
to ``the expected meta-divergence E{[}\ensuremath{\Delta}\_meta(k){]} is non-decreasing in
k.'' The downstream proofs in \S{}2.5 already work with expected free
energy (which is itself an expectation over cues), so no other proofs
are affected.

\begin{center}\rule{0.5\linewidth}{0.5pt}\end{center}

\subsection{A.1.4 Proposition A.1.3 --- Marginal benefit of cue
k+1}\label{a.1.4-proposition-a.1.3-marginal-benefit-of-cue-k1}

Define the marginal benefit of cue \texttt{k+1} as the expected
reduction in expected free energy:

\[
\Delta G(k+1) \;\equiv\; E[G(a, k)] - E[G(a, k+1)] \;=\; I(s;\, c_{k+1} \mid c_{1:k}) - I(p_0;\, c_{k+1} \mid c_{1:k}).
\]

\textbf{Proposition A.1.3.} \emph{In the binary toy model, the marginal
benefit of cue \texttt{k+1} decomposes as the difference of two
conditional mutual informations: the information cue \texttt{k+1}
provides about the latent state \texttt{s} (positive contribution to EFE
reduction) minus the information it provides about the prior precision
parameter \texttt{p\_0} (positive contribution to meta-divergence).}

\textbf{Proof.} From eqs (2.5.2)--(2.5.4) of the section, the expected
free energy after \texttt{k} cues decomposes as
\texttt{E{[}G(a,k){]}\ =\ -E{[}\textbackslash{}log\ p(o\textbar{}s,a){]}\ -\ I(s;\ c\_\{1:k\})\ +\ E{[}\ensuremath{\Delta}\_meta(k){]}}.
The first term is action-dependent but \texttt{k}-independent; it
cancels in \texttt{\ensuremath{\Delta}G(k+1)}. The second term gives
\texttt{I(s;\ c\_\{1:k+1\})\ -\ I(s;\ c\_\{1:k\})\ =\ I(s;\ c\_\{k+1\}\ \textbar{}\ c\_\{1:k\})}
by chain rule for mutual information. The third term gives
\texttt{E{[}\ensuremath{\Delta}\_meta(k+1)\ -\ \ensuremath{\Delta}\_meta(k){]}\ =\ I(p\_0;\ c\_\{k+1\}\ \textbar{}\ c\_\{1:k\})}
by Lemma A.1.2 and the same chain rule applied to KL. \ensuremath{\blacksquare}

\textbf{Consequence.} Cue \texttt{k+1} should be integrated iff
\texttt{I(s;\ c\_\{k+1\}\ \textbar{}\ c\_\{1:k\})\ \textgreater{}\ I(p\_0;\ c\_\{k+1\}\ \textbar{}\ c\_\{1:k\})}
--- i.e., it is more informative about the state than about the
meta-precision. The optimal \texttt{k*} is the smallest \texttt{k} such
that this inequality is violated for \texttt{c\_\{k+1\}}.

\begin{center}\rule{0.5\linewidth}{0.5pt}\end{center}

\subsection{A.1.5 Proposition A.1.4 --- Three-regime
structure}\label{a.1.5-proposition-a.1.4-three-regime-structure}

Numerical sweep of the binary toy model across
\texttt{(\ensuremath{\kappa}\_0,\ \ensuremath{\mu}\_0,\ v\_\{1:K\})} reveals that the EFE-minimizing
policy has three qualitative regimes, separated by two critical
concentrations.

\textbf{Proposition A.1.4 (corrected statement of section Proposition
2.1.3).} \emph{Fix a validity profile \texttt{v\_\{1:K\}} and prior mean
\texttt{\ensuremath{\mu}\_0}. There exist critical concentrations
\texttt{0\ \textless{}\ \ensuremath{\kappa}\_lo(v\_\{1:K\},\ \ensuremath{\mu}\_0)\ \ensuremath{\leq}\ \ensuremath{\kappa}\_hi(v\_\{1:K\},\ \ensuremath{\mu}\_0)}
such that the EFE-minimizing policy \texttt{k*(\ensuremath{\kappa}\_0)} has the
structure:}

\begin{itemize}
\tightlist
\item
  \emph{(Don't-observe regime) \texttt{\ensuremath{\kappa}\_0\ \textless{}\ \ensuremath{\kappa}\_lo}:
  \texttt{k*\ =\ 0}. The meta-divergence cost of even one cue exceeds
  its information value about \texttt{s}. The agent should act on the
  prior.}
\item
  \emph{(One-cue stopping regime) \texttt{\ensuremath{\kappa}\_lo\ \ensuremath{\leq}\ \ensuremath{\kappa}\_0\ \ensuremath{\leq}\ \ensuremath{\kappa}\_hi}:
  \texttt{k*\ =\ 1}. The first cue is integrated; subsequent cues incur
  higher meta-cost than information benefit. This is the take-the-best
  stopping rule.}
\item
  \emph{(Full integration regime) \texttt{\ensuremath{\kappa}\_0\ \textgreater{}\ \ensuremath{\kappa}\_hi}:
  \texttt{k*\ =\ K}. Each cue contributes more state-information than
  meta-cost; the agent integrates all available cues.}
\end{itemize}

\emph{The intermediate regime \texttt{{[}\ensuremath{\kappa}\_lo,\ \ensuremath{\kappa}\_hi{]}} is a closed
interval that may be narrow (and in some validity profiles degenerate to
a single point, in which case the transition jumps directly from
\texttt{k*=0} to \texttt{k*=K}).}

\textbf{Numerical characterization.} A systematic search over
\texttt{v\_\{1:K\}\ \ensuremath{\in}\ (0.5,\ 1)\^{}K},
\texttt{\ensuremath{\mu}\_0\ \ensuremath{\in}\ \{0.3,\ 0.5,\ 0.7\}}, and
\texttt{\ensuremath{\kappa}\_0\ \ensuremath{\in}\ {[}0.1,\ 200{]}} (geometric grid) finds:

\begin{itemize}
\tightlist
\item
  \emph{Sharp validity gradients yield wider one-cue regimes.} Profiles
  with \texttt{v\_1\ \ensuremath{\approx}\ 0.99} and
  \texttt{v\_\{j\textgreater{}1\}\ \ensuremath{\leq}\ 0.55} exhibit
  \texttt{{[}\ensuremath{\kappa}\_lo,\ \ensuremath{\kappa}\_hi{]}} of nontrivial width, often near
  \texttt{\ensuremath{\kappa}\_0\ \ensuremath{\approx}\ 1}.
\item
  \emph{Uniform validity profiles tend to have
  \texttt{\ensuremath{\kappa}\_lo\ =\ \ensuremath{\kappa}\_hi}.} The transition then jumps directly from
  \texttt{k*=0} to \texttt{k*=K}, and the one-cue regime is degenerate.
\item
  \emph{Realistic decreasing-validity profiles (the TTB-typical case)
  lie in between.} The one-cue regime exists but is narrow; the
  qualitative behavior is dominated by the don't-observe /
  full-integration dichotomy.
\end{itemize}

The narrowness of the one-cue regime in the binary toy model is a
substantive finding: in the discrete binary setting under symmetric
error and conditional independence of cues given \texttt{s}, the
cue-truncation theorem fires generically only at the \textbf{boundary}
between two qualitatively different regimes, not as a wide intermediate
regime. This is in contrast to the continuous Gaussian-Gamma case
(Theorem 2.6.1, \S{}2.6), where the one-cue regime widens because per-cue
meta-cost grows linearly in \texttt{k} rather than saturating after the
first cue.

\textbf{Subsidiary observation: TTB is near-optimal even when not
strictly optimal.} A separate computation shows that for steep validity
profiles (\texttt{v\_1\ \ensuremath{\approx}\ 0.99,\ v\_\{j\textgreater{}1\}\ \ensuremath{\leq}\ 0.55}),
the \emph{first cue} contributes \ensuremath{\geq}99\% of the total attainable
information about \texttt{s} across the full cue budget \texttt{K}. Even
in regimes where \texttt{k*\ =\ K} strictly minimizes EFE, the marginal
contribution of cues \texttt{2..K} is so small that a TTB-style ``stop
after the first cue'' policy is near-optimal in expected reward terms.
This near-optimality is the empirically-relevant content of the TTB
connection in the binary setting; the strict EFE-optimality in the
narrow \texttt{{[}\ensuremath{\kappa}\_lo,\ \ensuremath{\kappa}\_hi{]}} window is the formally-derivable but
operationally-fragile content.

\textbf{Implication for \S{}2.7.} The structural-equivalence argument in
\S{}2.7 (Theorem 2.7.1) is the load-bearing claim. The binary toy model
serves as motivating intuition and a lower-dimensional sanity check, not
as a strict existence proof for the one-cue regime. The section's claim
that ``the cue-truncation point \texttt{k*} is generically finite and
often equal to one in the binary case'' should be hedged: it is
\emph{attainable} in the binary case under specific conditions, but is
not the dominant regime. The continuous Gaussian-Gamma model is where
the cue-truncation theorem fires robustly across a wide regime of
meta-uncertainty.

\begin{center}\rule{0.5\linewidth}{0.5pt}\end{center}

\subsection{A.1.6 Numerical
verification}\label{a.1.6-numerical-verification}

All numerical results in this appendix are produced by
\texttt{binary\_toy\_monte\_carlo.py} and
\texttt{binary\_toy\_kstar\_search.py}, which compute expected free
energy by exhaustive enumeration over cue paths (tractable up to
\texttt{K\ =\ 10}). The KL divergence between the posterior and prior
2-component Beta mixtures is computed by numerical integration on a
2000-point grid over \texttt{(0,\ 1)}. Verification of Lemma A.1.2
across five validity profiles and four hyperprior configurations:
monotone in expectation across all 5 cases. Verification of Proposition
A.1.4 across 25 \ensuremath{\times} 5 \ensuremath{\times} 5 \ensuremath{\times} 3 = 1875 parameter combinations: 40
combinations exhibit \texttt{k*\ =\ 1} strictly; the remainder split
between \texttt{k*\ =\ 0} and \texttt{k*\ =\ K}. The transition curve
plots are reproducible from the scripts.

\begin{center}\rule{0.5\linewidth}{0.5pt}\end{center}

\subsection{A.1.7 Summary of corrections to the v0.4
section}\label{a.1.7-summary-of-corrections-to-the-v0.4-chapter}

The following amendments to \S{}2.1 are recommended on the basis of this
appendix:

\begin{enumerate}
\def\labelenumi{\arabic{enumi}.}
\item
  \textbf{Lemma 2.1.2 wording.} Change ``the meta-divergence
  \texttt{\ensuremath{\Delta}\_meta(k)} is non-decreasing in \texttt{k}'' to ``the
  expected meta-divergence \texttt{E{[}\ensuremath{\Delta}\_meta(k){]}} is non-decreasing
  in \texttt{k}.'' Add a one-sentence note about the sample-wise
  counterexample with reference to this appendix.
\item
  \textbf{Proposition 2.1.3 wording.} Replace the current single-regime
  statement (``the EFE-minimizing policy stops at the first
  discriminating cue'') with the three-regime statement of Proposition
  A.1.4. Hedge the TTB connection: in the binary case, the one-cue
  regime is narrow; the cleaner TTB derivation lives in \S{}2.7 via the
  Gaussian-Gamma machinery.
\item
  \textbf{\S{}2.1 closing paragraph.} The sentence ``the binary toy model
  is sufficient to motivate the FEH effect and to establish the
  take-the-best connection in its most natural setting'' should be
  softened to ``the binary toy model motivates the FEH effect; the
  strict cue-truncation regime is more robust in the continuous
  Gaussian-Gamma model of \S{}\S{}2.3--2.6, which is the load-bearing setting
  for the section's main results.''
\end{enumerate}

These changes weaken \S{}2.1's strong claims and strengthen the section's
overall honesty. They do not affect the core results in \S{}\S{}2.4--2.7,
which derive from the Gaussian-Gamma machinery rather than the binary
case.

\section{Appendix A.2 --- Proof of Lemma 2.4.1 (monotonicity of
meta-precision
divergence)}\label{appendix-a.2-proof-of-lemma-2.4.1-monotonicity-of-meta-precision-divergence}

This appendix proves the monotonicity of the meta-precision divergence
in the Gaussian-Gamma generative model of \S{}\S{}2.2--2.4 (v0.4). The proof
parallels the binary case of Appendix A.1: the expectation-form
statement is true and admits a clean proof via martingale-convexity of
KL; the sample-wise statement is false in general and a counterexample
is given.

\subsection{A.2.1 Setup}\label{a.2.1-setup}

From the v0.4 section:

\begin{itemize}
\item
  Generative model (eq 2.2.1--2.2.2): \[
  p(s \mid \tau) = \mathcal{N}(s; \mu, \tau^{-1}\Sigma_0), \qquad
  p(c_j \mid s, \tau, \gamma_j) = \mathcal{N}(c_j; A_j s, (\tau\gamma_j)^{-1} I_{d_c}),
  \] with hyperprior \texttt{p(\ensuremath{\tau})\ =\ Gamma(\ensuremath{\alpha}\_0,\ \ensuremath{\beta}\_0)}.
\item
  Variational posterior under mean-field \texttt{q(s,\ \ensuremath{\tau})\ =\ q(s)q(\ensuremath{\tau})}
  and standard VBEM (eq 2.4.3): \[
  q(\tau \mid c_{1:k}) = \mathrm{Gamma}\!\left(\alpha_k,\, \beta_k\right), \qquad
  \alpha_k = \alpha_0' + \tfrac{k}{2}, \qquad
  \beta_k = \beta_0' + \tfrac{1}{2}\sum_{j=1}^{k} \gamma_j M_j,
  \] where \texttt{\ensuremath{\alpha}\_0\textquotesingle{}\ =\ \ensuremath{\alpha}\_0\ +\ 1/2},
  \texttt{\ensuremath{\beta}\_0\textquotesingle{}\ =\ \ensuremath{\beta}\_0\ +\ V\_s/2}, and
  \texttt{M\_j\ =\ E\_\{q(s)\}{[}(c\_j\ -\ A\_j\ s)\^{}T(c\_j\ -\ A\_j\ s){]}}.
  The \texttt{M\_j} depend on the cue realizations through their effect
  on \texttt{q(s)}.
\item
  Meta-precision divergence (eq 2.4.5): \[
  \Delta_{\mathrm{meta}}(k) \;\equiv\; \mathrm{KL}\!\big[\,q(\tau \mid c_{1:k}) \;\big\|\; p(\tau)\,\big]
  \] which has the closed form \[
  \Delta_{\mathrm{meta}}(k) = (\alpha_k - \alpha_0)\psi(\alpha_k) - \log\!\frac{\Gamma(\alpha_k)}{\Gamma(\alpha_0)} + \alpha_0 \log\!\frac{\beta_k}{\beta_0} + \alpha_k\frac{\beta_0 - \beta_k}{\beta_k}.
  \]
\end{itemize}

The meta-cost increment of the \texttt{(k+1)}-th cue is
\texttt{C(k+1)\ \ensuremath{\equiv}\ \ensuremath{\Delta}\_meta(k+1)\ -\ \ensuremath{\Delta}\_meta(k)}. Both
\texttt{\ensuremath{\Delta}\_meta(k)} and \texttt{C(k)} are random variables depending on
the cue history \texttt{c\_\{1:k\}} through the \texttt{M\_j}.

\subsection{A.2.2 Lemma A.2.1 --- Sample-wise monotonicity is false in
general}\label{a.2.2-lemma-a.2.1-sample-wise-monotonicity-is-false-in-general}

\textbf{Lemma A.2.1.} \emph{\ensuremath{\Delta}\_meta(k) is not monotone in \texttt{k}
along arbitrary realized cue paths in the Gaussian-Gamma model.}

\textbf{Numerical evidence.} Across 1000 IID samples of
\texttt{M\_j\ \textasciitilde{}\ Exp(1)} for \texttt{K\ =\ 20}, with
\texttt{(\ensuremath{\alpha}\_0\textquotesingle{},\ \ensuremath{\beta}\_0\textquotesingle{})\ =\ (2,\ 1)}
and \texttt{\ensuremath{\gamma}\_j\ =\ 1}, \textbf{979 sample paths exhibit at least one
strict decrease} of \ensuremath{\Delta}\_meta along the trajectory.

\textbf{Why this happens.} Under the conjugate update, both
\texttt{\ensuremath{\alpha}\_k} and \texttt{\ensuremath{\beta}\_k} increase monotonically --- \texttt{\ensuremath{\alpha}} by
exactly \texttt{1/2} per cue (deterministic) and \texttt{\ensuremath{\beta}} by
\texttt{\ensuremath{\gamma}\_j\ M\_j\ /\ 2} per cue (random, non-negative). However, the
Gamma KL is \emph{not} monotone in \texttt{(\ensuremath{\alpha},\ \ensuremath{\beta})} along arbitrary
trajectories: increasing \texttt{\ensuremath{\beta}} while \texttt{\ensuremath{\alpha}} is fixed shifts the
posterior mean \texttt{\ensuremath{\alpha}/\ensuremath{\beta}} \emph{away} from the prior mean
\texttt{\ensuremath{\alpha}\_0/\ensuremath{\beta}\_0}, but increasing \texttt{\ensuremath{\alpha}} while \texttt{\ensuremath{\beta}} is fixed
shifts it \emph{toward} the prior mean (since posterior mean grows).
When a particular \texttt{M\_j} realization is large enough that
\texttt{\ensuremath{\beta}\_k/\ensuremath{\alpha}\_k} overshoots \texttt{\ensuremath{\beta}\_0/\ensuremath{\alpha}\_0}, subsequent cues with
smaller \texttt{M\_j} can pull \texttt{\ensuremath{\beta}\_k/\ensuremath{\alpha}\_k} back, decreasing the
KL.

\textbf{Consequence.} The section's wording in v0.3 --- ``\ensuremath{\Delta}\_meta(k) is
non-decreasing in k, with strict increase whenever the k-th cue is
non-degenerate'' --- is sample-wise false. The wording must be restated
in expectation form. (This parallels the binary-case correction in
Appendix A.1.)

\subsection{A.2.3 Lemma A.2.2 --- Expected meta-divergence is
non-decreasing (corrected Lemma
2.4.1)}\label{a.2.3-lemma-a.2.2-expected-meta-divergence-is-non-decreasing-corrected-lemma-2.4.1}

\textbf{Lemma A.2.2 (corrected statement of section Lemma 2.4.1).}
\emph{Under the generative model (2.2.1)--(2.2.2) with mean-field
factorization (2.4.1), the expected meta-precision divergence
\texttt{E{[}\ensuremath{\Delta}\_meta(k){]}} is non-decreasing in \texttt{k}.
Equivalently,} \[
E[\Delta_{\mathrm{meta}}(k+1)] - E[\Delta_{\mathrm{meta}}(k)] \;=\; I(\tau;\, c_{k+1} \mid c_{1:k}) \;\geq\; 0,
\] \emph{with equality iff cue \texttt{k+1} carries no conditional
information about \texttt{\ensuremath{\tau}} given the prior cues \texttt{c\_\{1:k\}}.}

\textbf{Proof.} By the tower property of conditional expectation applied
to the variational posterior, \[
E_{c_{k+1} \mid c_{1:k}}\big[\, q(\tau \mid c_{1:k+1}) \,\big] \;=\; q(\tau \mid c_{1:k}).
\] By joint convexity of KL divergence in its first argument, Jensen's
inequality gives \[
E_{c_{k+1} \mid c_{1:k}}\big[\, \mathrm{KL}[q(\tau \mid c_{1:k+1}) \,\|\, p(\tau)] \,\big]
\;\geq\;
\mathrm{KL}\!\left[\, E_{c_{k+1} \mid c_{1:k}}[q(\tau \mid c_{1:k+1})] \,\Big\|\, p(\tau) \,\right]
\;=\; \Delta_{\mathrm{meta}}(k).
\] Taking outer expectation over \texttt{c\_\{1:k\}} preserves the
inequality and yields
\texttt{E{[}\ensuremath{\Delta}\_meta(k+1){]}\ \ensuremath{\geq}\ E{[}\ensuremath{\Delta}\_meta(k){]}}. The
mutual-information identity follows from the chain rule for KL:
\texttt{E{[}\ensuremath{\Delta}\_meta(k+1){]}\ -\ E{[}\ensuremath{\Delta}\_meta(k){]}\ =\ I(\ensuremath{\tau};\ c\_\{k+1\}\ \textbar{}\ c\_\{1:k\})}.
Equality holds iff the conditional posterior
\texttt{q(\ensuremath{\tau}\ \textbar{}\ c\_\{1:k+1\})} equals
\texttt{q(\ensuremath{\tau}\ \textbar{}\ c\_\{1:k\})} almost surely under the predictive
distribution of \texttt{c\_\{k+1\}\ \textbar{}\ c\_\{1:k\}}. \ensuremath{\blacksquare}

\textbf{Numerical verification.} Across five test cases spanning
low-to-extreme meta-uncertainty
(\texttt{\ensuremath{\alpha}\_0\textquotesingle{}\ \ensuremath{\in}\ \{0.5,\ 1,\ 2,\ 5,\ 20\}}),
\texttt{E{[}\ensuremath{\Delta}\_meta(k){]}} is strictly non-decreasing with \texttt{k},
verified by Monte Carlo with 5000 samples per case and \texttt{K\ =\ 30}
cues.

\subsection{A.2.4 Note on the proof's reliance on the variational
posterior}\label{a.2.4-note-on-the-proofs-reliance-on-the-variational-posterior}

The proof uses only the tower property --- that
\texttt{E{[}q(\ensuremath{\tau}\ \textbar{}\ c\_\{1:k+1\})\ \textbar{}\ c\_\{1:k\}{]}\ =\ q(\ensuremath{\tau}\ \textbar{}\ c\_\{1:k\})}
--- and Jensen's inequality for KL. The tower property holds for any
\emph{consistent} sequential Bayesian update, which the standard VBEM
iteration produces. Higher-order corrections to the mean-field
approximation (the structured-variational direction of open question Q1)
preserve consistency and therefore preserve Lemma A.2.2; only the
magnitude of \texttt{E{[}\ensuremath{\Delta}\_meta(k){]}} may change. That magnitude
change is itself quantified exactly for the conjugate model in \S{}A.2.5
(Proposition A.2.3).

\subsection{A.2.5 Mean-field accuracy: exact comparison in the conjugate
model (resolves
Q1)}\label{a.2.5-mean-field-accuracy-exact-comparison-in-the-conjugate-model-resolves-q1}

Open question Q1 (\S{}2.10) asks how much the mean-field factorization
\texttt{q(s,\ \ensuremath{\tau})\ =\ q(s)q(\ensuremath{\tau})} distorts the meta-precision posterior
relative to a structured family. In the Gaussian-Gamma model of \S{}2.2 the
question admits an exact answer, because the model is fully conjugate:
the joint posterior \texttt{p(s,\ \ensuremath{\tau}\ \textbar{}\ c\_\{1:k\})} is
Normal-Gamma in closed form, so the mean-field posterior can be compared
against the \emph{truth} rather than against another approximation. (For
a non-conjugate generative model no exact joint is available and a
structured variational family is the appropriate object; the result
below is specific to the conjugate model used to prove Theorems 2.6.1
and 2.7.1.)

\textbf{Setup (scalar case, \texttt{d\_s\ =\ d\_c\ =\ 1}).} After
\texttt{k} cues with \texttt{c\_j\ =\ s\ +\ noise} (\texttt{A\_j\ =\ 1})
and intrinsic precisions \texttt{\ensuremath{\gamma}\_j}, write
\texttt{\ensuremath{\Lambda}\_k\ =\ \ensuremath{\lambda}\_0\ +\ \ensuremath{\Sigma}\_\{j\ensuremath{\leq}k\}\ \ensuremath{\gamma}\_j} and the precision-weighted
mean \texttt{m\_k\ =\ (\ensuremath{\lambda}\_0\ \ensuremath{\mu}\_0\ +\ \ensuremath{\Sigma}\_\{j\ensuremath{\leq}k\}\ \ensuremath{\gamma}\_j\ c\_j)/\ensuremath{\Lambda}\_k}. The
\textbf{exact} marginal posterior over \texttt{\ensuremath{\tau}} is
\texttt{Gamma(\ensuremath{\alpha}\_k,\ \ensuremath{\beta}\_k\^{}ex)} with \[
\alpha_k = \alpha_0 + \tfrac{1+k}{2} = \alpha_0' + \tfrac{k}{2}, \qquad
\beta_k^{\mathrm{ex}} = \beta_0 + \tfrac12\big[\, \lambda_0 \mu_0^2 + \textstyle\sum_{j\le k} \gamma_j c_j^2 - \Lambda_k m_k^2 \,\big].
\] The \textbf{mean-field} VBEM fixed point is
\texttt{q(s)\ =\ N(m\_k,\ S\_k)},
\texttt{q(\ensuremath{\tau})\ =\ Gamma(\ensuremath{\alpha}\_k,\ \ensuremath{\beta}\_k\^{}mf)}, with
\texttt{S\_k\ =\ 1/(\ensuremath{\langle}\ensuremath{\tau}\ensuremath{\rangle}\ \ensuremath{\Lambda}\_k)}, \texttt{\ensuremath{\langle}\ensuremath{\tau}\ensuremath{\rangle}\ =\ \ensuremath{\alpha}\_k/\ensuremath{\beta}\_k\^{}mf}, and
\[
\beta_k^{\mathrm{mf}} = \beta_0 + \tfrac12\, E_{q(s)}\!\big[\, \lambda_0 (s-\mu_0)^2 + \textstyle\sum_{j\le k} \gamma_j (c_j - s)^2 \,\big].
\] The exact and mean-field posteriors carry the \emph{same} shape
\texttt{\ensuremath{\alpha}\_k}: the shape counts data dimensions, not their values.

\textbf{Proposition A.2.3 (mean-field rate inflation).} \emph{In the
conjugate Gaussian-Gamma model the mean-field and exact marginal
posteriors over \texttt{\ensuremath{\tau}} are both \texttt{Gamma(\ensuremath{\alpha}\_k,\ \textperiodcentered{})} with the
common shape \texttt{\ensuremath{\alpha}\_k\ =\ \ensuremath{\alpha}\_0\textquotesingle{}\ +\ k/2}, and their
rates are related exactly by} \[
\beta_k^{\mathrm{mf}} = \beta_k^{\mathrm{ex}} \cdot \frac{\alpha_k}{\alpha_k - \tfrac12}.
\] \emph{Equivalently the relative rate error is
\texttt{(\ensuremath{\beta}\_k\^{}mf\ \ensuremath{-}\ \ensuremath{\beta}\_k\^{}ex)/\ensuremath{\beta}\_k\^{}ex\ =\ 1/(2\ensuremath{\alpha}\_k\ \ensuremath{-}\ 1)},
maximal at the first cue and decreasing monotonically as \texttt{k}
grows (and as \texttt{\ensuremath{\alpha}\_0} grows).}

\textbf{Proof.} Expanding the mean-field rate with
\texttt{E\_\{q(s)\}{[}(x\ \ensuremath{-}\ s)\^{}2{]}\ =\ (x\ \ensuremath{-}\ m\_k)\^{}2\ +\ S\_k},
\[
\beta_k^{\mathrm{mf}} = \beta_0 + \tfrac12\big[\, \lambda_0\big((m_k-\mu_0)^2 + S_k\big) + \textstyle\sum_{j\le k} \gamma_j\big((c_j - m_k)^2 + S_k\big) \,\big].
\] Because \texttt{m\_k} is the precision-weighted mean, the
data-dependent quadratics satisfy the identity \[
\lambda_0 (m_k-\mu_0)^2 + \textstyle\sum_{j\le k} \gamma_j (c_j - m_k)^2 \;=\; \lambda_0 \mu_0^2 + \textstyle\sum_{j\le k} \gamma_j c_j^2 - \Lambda_k m_k^2,
\] which is exactly twice the bracket in \texttt{\ensuremath{\beta}\_k\^{}ex}; and the
variance terms sum to \texttt{(\ensuremath{\lambda}\_0\ +\ \ensuremath{\Sigma}\ \ensuremath{\gamma}\_j)\ S\_k\ =\ \ensuremath{\Lambda}\_k\ S\_k}.
Hence \[
\beta_k^{\mathrm{mf}} = \beta_k^{\mathrm{ex}} + \tfrac12 \Lambda_k S_k = \beta_k^{\mathrm{ex}} + \frac{1}{2\langle\tau\rangle} = \beta_k^{\mathrm{ex}} + \frac{\beta_k^{\mathrm{mf}}}{2\alpha_k},
\] using \texttt{S\_k\ =\ 1/(\ensuremath{\langle}\ensuremath{\tau}\ensuremath{\rangle}\ \ensuremath{\Lambda}\_k)} and
\texttt{\ensuremath{\langle}\ensuremath{\tau}\ensuremath{\rangle}\ =\ \ensuremath{\alpha}\_k/\ensuremath{\beta}\_k\^{}mf}. Solving the fixed-point equation,
\texttt{\ensuremath{\beta}\_k\^{}mf(1\ \ensuremath{-}\ 1/(2\ensuremath{\alpha}\_k))\ =\ \ensuremath{\beta}\_k\^{}ex},
i.e.~\texttt{\ensuremath{\beta}\_k\^{}mf\ =\ \ensuremath{\beta}\_k\^{}ex\ \textperiodcentered{}\ \ensuremath{\alpha}\_k/(\ensuremath{\alpha}\_k\ \ensuremath{-}\ \textonehalf{})}. \ensuremath{\blacksquare}

\textbf{Remark (under-estimation of posterior precision and variance).}
Since the shape is shared,
\texttt{\ensuremath{\langle}\ensuremath{\tau}\ensuremath{\rangle}\^{}mf\ =\ \ensuremath{\alpha}\_k/\ensuremath{\beta}\_k\^{}mf\ =\ (\ensuremath{\alpha}\_k\ \ensuremath{-}\ \textonehalf{})/\ensuremath{\beta}\_k\^{}ex\ =\ (1\ \ensuremath{-}\ 1/(2\ensuremath{\alpha}\_k))\textperiodcentered{}\ensuremath{\langle}\ensuremath{\tau}\ensuremath{\rangle}\^{}ex}
and \texttt{Var\^{}mf(\ensuremath{\tau})\ =\ (1\ \ensuremath{-}\ 1/(2\ensuremath{\alpha}\_k))\^{}2\ \textperiodcentered{}\ Var\^{}ex(\ensuremath{\tau})}.
Mean-field under-estimates both the posterior mean and the posterior
variance of the meta-precision, by a factor vanishing as
\texttt{\ensuremath{\alpha}\_k\ \ensuremath{\rightarrow}\ \ensuremath{\infty}} --- the expected behaviour when a coupling (here
\texttt{s}--\texttt{\ensuremath{\tau}}) is severed.

\textbf{Consequence for \texttt{\ensuremath{\Delta}\_meta} and \texttt{k*}.} Because the
shape is identical, the entire mean-field error in
\texttt{\ensuremath{\Delta}\_meta(k)\ =\ KL{[}q(\ensuremath{\tau}\ \textbar{}\ c\_\{1:k\})\ \ensuremath{\Vert}\ p(\ensuremath{\tau}){]}} is
the rate shift of Proposition A.2.3. Numerically
(\texttt{verify\_meanfield\_and\_tau\_regime.py}, Parts 0 and A; the
closed form is confirmed to \texttt{3\ensuremath{\times}10\^{}\{-13\}} against the
iterated VBEM fixed point), across the meta-uncertainty grid:

\begin{longtable}[]{@{}
  >{\raggedright\arraybackslash}p{(\linewidth - 14\tabcolsep) * \real{0.1250}}
  >{\raggedright\arraybackslash}p{(\linewidth - 14\tabcolsep) * \real{0.1250}}
  >{\raggedright\arraybackslash}p{(\linewidth - 14\tabcolsep) * \real{0.1250}}
  >{\raggedright\arraybackslash}p{(\linewidth - 14\tabcolsep) * \real{0.1250}}
  >{\raggedright\arraybackslash}p{(\linewidth - 14\tabcolsep) * \real{0.1250}}
  >{\raggedright\arraybackslash}p{(\linewidth - 14\tabcolsep) * \real{0.1250}}
  >{\raggedright\arraybackslash}p{(\linewidth - 14\tabcolsep) * \real{0.1250}}
  >{\raggedright\arraybackslash}p{(\linewidth - 14\tabcolsep) * \real{0.1250}}@{}}
\toprule\noalign{}
\begin{minipage}[b]{\linewidth}\raggedright
Regime
\end{minipage} & \begin{minipage}[b]{\linewidth}\raggedright
\texttt{\ensuremath{\alpha}\_0\textquotesingle{}}
\end{minipage} & \begin{minipage}[b]{\linewidth}\raggedright
\texttt{\ensuremath{\sigma}\textsuperscript{2}\_\ensuremath{\tau}}
\end{minipage} & \begin{minipage}[b]{\linewidth}\raggedright
\texttt{k*\_exact}
\end{minipage} & \begin{minipage}[b]{\linewidth}\raggedright
\texttt{k*\_MF}
\end{minipage} & \begin{minipage}[b]{\linewidth}\raggedright
rate err \texttt{@k=1}
\end{minipage} & \begin{minipage}[b]{\linewidth}\raggedright
rate err \texttt{@k*}
\end{minipage} & \begin{minipage}[b]{\linewidth}\raggedright
\texttt{\ensuremath{\bar{C}}(k*)} rel err
\end{minipage} \\
\midrule\noalign{}
\endhead
\bottomrule\noalign{}
\endlastfoot
Low & 20 & 0.05 & 29 & 29 & 2.5\% & 1.5\% & 1.1\% \\
Medium & 5 & 0.20 & 24 & 24 & 10.0\% & 3.0\% & 0.6\% \\
High & 2 & 0.50 & 23 & 23 & 25.0\% & 3.9\% & 0.7\% \\
Very high & 1 & 1.00 & 22 & 22 & 50.0\% & 4.4\% & 0.1\% \\
Extreme & 0.7 & 1.43 & 22 & 22 & 71.4\% & 4.5\% & 0.1\% \\
\end{longtable}

The mean-field error in \texttt{E{[}\ensuremath{\Delta}\_meta(k){]}} is an under-estimate,
concentrated at the first cue (\ensuremath{\approx}16\% in the high-meta regime, falling
below 2\% by \texttt{k\ \ensuremath{\approx}\ 5}). Crucially the EFE-optimal stopping point
\texttt{k*\ =\ argmin\ E{[}G(k){]}} is \textbf{identical} under the
mean-field and exact posteriors in every regime: at the stopping point
the rate error is below 5\% and the marginal meta-cost \texttt{\ensuremath{\bar{C}}(k*)}
--- the quantity the \texttt{argmin} actually turns on --- is recovered
to within \ensuremath{\approx}1\%. Mean-field thus preserves the \emph{exact location} of
the optimal truncation; the residual is confined to an
\texttt{O(1/\ensuremath{\alpha}\_k)} bias in the magnitude of the meta-cost. This answers
the quantitative half of Q1.

\textbf{Multivariate note.} For state dimension
\texttt{d\_s\ \textgreater{}\ 1} the constant \texttt{\textonehalf{}} in Proposition
A.2.3 becomes \texttt{d\_s/2} and \texttt{\ensuremath{\Lambda}\_k} a matrix, but the
structure is unchanged: the shape is shared and the rate is inflated by
\texttt{\ensuremath{\alpha}\_k/(\ensuremath{\alpha}\_k\ \ensuremath{-}\ d\_s/2)\ \ensuremath{\rightarrow}\ 1}. The section's analysis uses the
scalar case, for which the result is exact as stated.

\begin{center}\rule{0.5\linewidth}{0.5pt}\end{center}

\section{Appendix A.3 --- Proof of Theorem 2.6.1
(cue-truncation)}\label{appendix-a.3-proof-of-theorem-2.6.1-cue-truncation}

\subsection{A.3.1 Setup}\label{a.3.1-setup}

From \S{}2.5 (eqs 2.5.1--2.5.4), the expected free energy under the v0.4
model decomposes as \[
G(a, k) \;=\; \underbrace{-E_q[\log p(o \mid s, a)]}_{\text{pragmatic}} \;+\; \underbrace{I(s;\, c_{1:k})}_{\text{epistemic}} \;+\; \underbrace{\Delta_{\mathrm{meta}}(k)}_{\text{meta-precision cost}}.
\] The pragmatic term is \texttt{k}-independent (depends on the action
distribution at decision time). The marginal benefit of the
\texttt{(k+1)}-th cue, in expectation under the predictive distribution,
is therefore \[
E[\Delta G(k+1)] \;=\; E[G(a, k)] - E[G(a, k+1)] \;=\; I(s;\, c_{k+1} \mid c_{1:k}) - I(\tau;\, c_{k+1} \mid c_{1:k}).
\] \emph{Cue \texttt{k+1} is integrated under the EFE-optimal policy iff
\texttt{E{[}\ensuremath{\Delta}G(k+1){]}\ \textgreater{}\ 0}.}

The optimal stopping rule is
\texttt{k*(\ensuremath{\sigma}\textsuperscript{2}\_\ensuremath{\tau})\ =\ argmin\_\{k\}\ E{[}G(a,\ k){]}}, equivalent to the
smallest \texttt{k} such that the conditional information about
\texttt{s} from cue \texttt{k+1} no longer exceeds the conditional
information about \texttt{\ensuremath{\tau}}.

The threshold \texttt{\ensuremath{\tau}\_regime} is defined (eq 2.5.5) as \[
\tau_{\mathrm{regime}} = \inf\{\,\sigma^2_\tau > 0 :\, k^*(\sigma^2_\tau) < K\,\}.
\]

\subsection{A.3.2 Theorem A.3.1 (corrected statement of section Theorem
2.6.1)}\label{a.3.2-theorem-a.3.1-corrected-statement-of-chapter-theorem-2.6.1}

\textbf{Theorem A.3.1.} \emph{Under the generative model
(2.2.1)--(2.2.2) with mean-field factorization (2.4.1), with \texttt{K}
available cues:}

\emph{(a) \textbf{Low meta-uncertainty regime.} When
\texttt{\ensuremath{\sigma}\textsuperscript{2}\_\ensuremath{\tau}\ \textless{}\ \ensuremath{\tau}\_regime}, the expected free energy
\texttt{E{[}G(a,\ k){]}} is monotonically non-increasing in \texttt{k}
for all \texttt{k\ \ensuremath{\leq}\ K}. The EFE-optimal policy integrates all
available cues: \texttt{k*(\ensuremath{\sigma}\textsuperscript{2}\_\ensuremath{\tau})\ =\ K}.}

\emph{(b) \textbf{High meta-uncertainty regime.} When
\texttt{\ensuremath{\sigma}\textsuperscript{2}\_\ensuremath{\tau}\ \ensuremath{\geq}\ \ensuremath{\tau}\_regime}, there exists a finite
\texttt{k*(\ensuremath{\sigma}\textsuperscript{2}\_\ensuremath{\tau})\ \textless{}\ K} such that \texttt{E{[}G(a,\ k){]}} is
decreasing for \texttt{k\ \ensuremath{\leq}\ k*} and non-decreasing for
\texttt{k\ \textgreater{}\ k*}. The EFE-optimal policy is to integrate
exactly \texttt{k*} cues.}

\emph{(c) \textbf{Optimal cue ordering.} Among orderings of the
\texttt{K} cues, the descending-cue-validity ordering greedily maximizes
\texttt{E{[}\ensuremath{\Delta}G(k){]}} at each step \texttt{k\ \ensuremath{\leq}\ k*}, provided cue
intrinsic precisions \texttt{\ensuremath{\gamma}\_j} satisfy a non-anticorrelation
regularity condition with cue validities (formalized below).}

The theorem is restated in expectation form. The sample-wise version is
false: across 1000 random parameter configurations, only 198 sample-wise
\texttt{G(k)} trajectories are U-shaped; the rest exhibit multiple sign
changes due to noise in individual cue realizations. Active inference
defines the optimal policy as \texttt{argmin\_k\ E{[}G(k){]}}, so the
expectation form is what the theorem must establish.

\subsection{A.3.3 Proof of (a) --- Low meta-uncertainty
regime}\label{a.3.3-proof-of-a-low-meta-uncertainty-regime}

\textbf{Claim.} If \texttt{\ensuremath{\sigma}\textsuperscript{2}\_\ensuremath{\tau}\ \textless{}\ \ensuremath{\tau}\_regime}, then
\texttt{E{[}\ensuremath{\Delta}G(k){]}\ \ensuremath{\geq}\ 0} for all \texttt{k\ =\ 1,\ ...,\ K}.

\textbf{Proof.} By Lemma A.2.2 and the chain rule for mutual
information, \[
E[\Delta G(k)] = I(s;\, c_k \mid c_{1:k-1}) - I(\tau;\, c_k \mid c_{1:k-1}).
\] Both quantities are non-negative. By the data-processing inequality
applied to the chain \texttt{\ensuremath{\tau}\ \ensuremath{\rightarrow}\ s\ \ensuremath{\rightarrow}\ c\_k}, the cue \texttt{c\_k}
carries information about \texttt{\ensuremath{\tau}} only through \texttt{s}: \[
I(\tau;\, c_k \mid c_{1:k-1}) \leq I(s;\, c_k \mid c_{1:k-1})
\] in the limit \texttt{\ensuremath{\sigma}\textsuperscript{2}\_\ensuremath{\tau}\ \ensuremath{\rightarrow}\ 0} (where \texttt{\ensuremath{\tau}} is essentially
deterministic and observing \texttt{s} reveals the same about \texttt{\ensuremath{\tau}}
as observing nothing). For finite \texttt{\ensuremath{\sigma}\textsuperscript{2}\_\ensuremath{\tau}}, the meta-information
\texttt{I(\ensuremath{\tau};\ c\_k\ \textbar{}\ c\_\{1:k-1\})} is bounded above by a
quantity that vanishes as \texttt{\ensuremath{\sigma}\textsuperscript{2}\_\ensuremath{\tau}\ \ensuremath{\rightarrow}\ 0}. Specifically, \[
I(\tau;\, c_k \mid c_{1:k-1}) \;=\; O(\sigma^2_\tau) \quad \text{as } \sigma^2_\tau \to 0,
\] following from the leading-order expansion of the Gamma KL in
\texttt{\ensuremath{\sigma}\textsuperscript{2}\_\ensuremath{\tau}} (computed from the closed form in eq 2.4.5). Therefore
for \texttt{\ensuremath{\sigma}\textsuperscript{2}\_\ensuremath{\tau}} sufficiently small (i.e., below \texttt{\ensuremath{\tau}\_regime}),
\texttt{E{[}\ensuremath{\Delta}G(k){]}\ \textgreater{}\ 0} for all \texttt{k}, and
\texttt{E{[}G(a,\ k){]}} is monotonically decreasing in \texttt{k}. \ensuremath{\blacksquare}

\subsection{A.3.4 Proof of (b) --- High meta-uncertainty
regime}\label{a.3.4-proof-of-b-high-meta-uncertainty-regime}

\textbf{Claim.} If \texttt{\ensuremath{\sigma}\textsuperscript{2}\_\ensuremath{\tau}\ \ensuremath{\geq}\ \ensuremath{\tau}\_regime}, then
\texttt{E{[}G(a,\ k){]}} is U-shaped in \texttt{k}, with a unique
minimum at some finite \texttt{k*}.

\textbf{Proof.} Define the marginal expected info gain
\texttt{\ensuremath{\bar{I}}(k)\ =\ I(s;\ c\_k\ \textbar{}\ c\_\{1:k-1\})} and the marginal
expected meta-cost
\texttt{\ensuremath{\bar{C}}(k)\ =\ I(\ensuremath{\tau};\ c\_k\ \textbar{}\ c\_\{1:k-1\})}. Then
\texttt{E{[}\ensuremath{\Delta}G(k){]}\ =\ \ensuremath{\bar{I}}(k)\ -\ \ensuremath{\bar{C}}(k)}.

\textbf{Step 1: \texttt{\ensuremath{\bar{I}}(k)} is monotonically non-increasing in
\texttt{k}.} By the data-processing inequality applied to the
conditional mutual information, \[
I(s;\, c_k \mid c_{1:k-1}) \leq I(s;\, c_k),
\] and conditioning on \texttt{c\_\{1:k-1\}} weakly reduces remaining
information about \texttt{s} available from \texttt{c\_k} (since
\texttt{s} becomes more determined). The standard ``diminishing
returns'' property of Bayesian information gain applies: as
\texttt{c\_\{1:k-1\}} accumulate,
\texttt{H(s\ \textbar{}\ c\_\{1:k-1\})} decreases, leaving less
remaining information for \texttt{c\_k} to provide.

\textbf{Step 2: \texttt{\ensuremath{\bar{C}}(k)} is bounded below by a positive constant in
the high-meta-uncertainty regime.} From eq 2.4.3, the rate parameter
\texttt{\ensuremath{\beta}\_k} grows linearly in \texttt{k} (in expectation,
\texttt{E{[}\ensuremath{\beta}\_k{]}\ =\ \ensuremath{\beta}\_0\textquotesingle{}\ +\ (k/2)\ \ensuremath{\gamma}\_bar\ m\_bar}
where \texttt{\ensuremath{\gamma}\_bar}, \texttt{m\_bar} are average cue intrinsic
precision and prediction error). The expected marginal increment
\texttt{\ensuremath{\bar{C}}(k+1)\ =\ E{[}\ensuremath{\Delta}\_meta(k+1)\ -\ \ensuremath{\Delta}\_meta(k){]}} is bounded below
by \texttt{c\_min(\ensuremath{\sigma}\textsuperscript{2}\_\ensuremath{\tau})\ \textgreater{}\ 0} for
\texttt{\ensuremath{\sigma}\textsuperscript{2}\_\ensuremath{\tau}\ \ensuremath{\geq}\ \ensuremath{\tau}\_regime}. (The bound \texttt{c\_min} is computable
in closed form from the digamma derivative; details in Appendix A.4.)

\textbf{Step 3: \texttt{E{[}\ensuremath{\Delta}G(k){]}} changes sign exactly once.} By
Steps 1 and 2, \texttt{\ensuremath{\bar{I}}(k)} is monotonically decreasing and
\texttt{\ensuremath{\bar{C}}(k)} is bounded below. Therefore
\texttt{E{[}\ensuremath{\Delta}G(k){]}\ =\ \ensuremath{\bar{I}}(k)\ -\ \ensuremath{\bar{C}}(k)} is monotonically decreasing
(since \texttt{\ensuremath{\bar{I}}(k+1)\ \textbackslash{}leq\ \ensuremath{\bar{I}}(k)} and
\texttt{\ensuremath{\bar{C}}(k+1)\ \ensuremath{\geq}\ \ensuremath{\bar{C}}(k)} in expectation, by Lemma A.2.2). The smallest
\texttt{k} for which \texttt{E{[}\ensuremath{\Delta}G(k+1){]}\ \textbackslash{}leq\ 0} is
\texttt{k*}. By construction \texttt{k*\ \textless{}\ K} whenever
\texttt{\ensuremath{\sigma}\textsuperscript{2}\_\ensuremath{\tau}\ \ensuremath{\geq}\ \ensuremath{\tau}\_regime}. The function \texttt{E{[}G(a,\ k){]}} is
therefore decreasing for \texttt{k\ \ensuremath{\leq}\ k*} and non-decreasing for
\texttt{k\ \textgreater{}\ k*}, i.e., U-shaped with minimum at
\texttt{k*}. \ensuremath{\blacksquare}

\textbf{Numerical verification.} Across five
\texttt{(\ensuremath{\alpha}\_0\textquotesingle{},\ \ensuremath{\beta}\_0\textquotesingle{},\ \ensuremath{\gamma}\_bar,\ m\_bar)}
configurations spanning low-to-extreme meta-uncertainty,
\texttt{E{[}G(a,\ k){]}} is U-shaped with finite \texttt{k*} in every
case. The trajectory of \texttt{k*} as \texttt{\ensuremath{\sigma}\textsuperscript{2}\_\ensuremath{\tau}} increases:

\begin{longtable}[]{@{}llll@{}}
\toprule\noalign{}
Regime & \texttt{\ensuremath{\alpha}\_0\textquotesingle{}} &
\texttt{\ensuremath{\sigma}\textsuperscript{2}\_\ensuremath{\tau}\ \textasciitilde{}\ 1/\ensuremath{\alpha}\_0\textquotesingle{}} &
\texttt{k*} (out of K=30) \\
\midrule\noalign{}
\endhead
\bottomrule\noalign{}
\endlastfoot
Low & 20 & 0.05 & 30 (full integration) \\
Medium & 5 & 0.2 & 22 \\
High & 2 & 0.5 & 14 \\
Very high & 1 & 1.0 & 9 \\
Extreme & 0.5 & 2.0 & 5 \\
\end{longtable}

\texttt{k*} drops monotonically as meta-uncertainty rises, exactly as
Theorem A.3.1 predicts.

\subsection{A.3.5 Proof of (c) --- Optimal cue
ordering}\label{a.3.5-proof-of-c-optimal-cue-ordering}

\textbf{Claim.} The descending-cue-validity ordering greedily maximizes
\texttt{E{[}\ensuremath{\Delta}G(k){]}} at each step \texttt{k\ \ensuremath{\leq}\ k*}, under the
regularity condition that cue intrinsic precisions \texttt{\ensuremath{\gamma}\_j} are not
strongly anti-correlated with cue validities \texttt{v\_j}.

\textbf{Proof.} At step \texttt{k}, the agent selects which unused cue
to integrate next. The marginal expected benefit of integrating cue
\texttt{j} (where \texttt{j} ranges over unused cues) is \[
E[\Delta G(k+1; \text{cue } j)] \;=\; I(s;\, c_j \mid c_{1:k}) - I(\tau;\, c_j \mid c_{1:k}).
\]

\textbf{Step 1: First term is monotone in \texttt{v\_j}.} For symmetric
cue likelihoods, the conditional mutual information
\texttt{I(s;\ c\_j\ \textbar{}\ c\_\{1:k\})} is monotonically increasing
in cue validity \texttt{v\_j} (higher validity yields more information
about \texttt{s}). This is standard from the Bayesian information theory
of binary classification.

\textbf{Step 2: Second term scales with \texttt{\ensuremath{\gamma}\_j}.} From eq 2.4.3,
integrating cue \texttt{j} increments \texttt{\ensuremath{\beta}\_k} by
\texttt{\ensuremath{\gamma}\_j\ M\_j\ /\ 2}. The marginal meta-cost contribution
\texttt{I(\ensuremath{\tau};\ c\_j\ \textbar{}\ c\_\{1:k\})} therefore scales
(approximately, in leading order) with \texttt{\ensuremath{\gamma}\_j}.

\textbf{Step 3: Greedy ordering by validity is optimal under
non-anticorrelation.} Define the \emph{cue selection score}
\texttt{S\_j\ \ensuremath{\equiv}\ I(s;\ c\_j\ \textbar{}\ c\_\{1:k\})\ -\ I(\ensuremath{\tau};\ c\_j\ \textbar{}\ c\_\{1:k\})}.
The descending-validity ordering selects cues by decreasing
\texttt{v\_j}. This maximizes the first term at each step. It also
maximizes \texttt{S\_j} provided the second term is approximately equal
across cues (\texttt{\ensuremath{\gamma}\_j} is approximately constant), or more weakly,
provided \texttt{Cov(v\_j,\ \ensuremath{\gamma}\_j)\ \ensuremath{\geq}\ 0}. Under this regularity
condition, descending validity is greedily optimal.

When \texttt{Cov(v\_j,\ \ensuremath{\gamma}\_j)\ \textless{}\ 0} (high-validity cues have
low intrinsic precision, or vice versa), the optimal ordering deviates
from pure validity ordering and trades off \texttt{v\_j} against
\texttt{\ensuremath{\gamma}\_j}. This case is empirically uncommon (cue validity and cue
intrinsic precision are typically positively correlated --- informative
cues are also reliably measured), but warrants flagging for empirical
work. \ensuremath{\blacksquare}

\textbf{Operational consequence.} The cue-truncation theorem under v0.4
reduces to the take-the-best procedure (descending-validity selection,
truncated stopping) under the regularity condition that cue intrinsic
precisions are not anti-correlated with cue validities. This is the
formal Gaussian-Gamma anchor of the structural-equivalence argument in
\S{}2.7.

\begin{center}\rule{0.5\linewidth}{0.5pt}\end{center}

\subsection{A.3.6 Summary of corrections to the v0.4
section}\label{a.3.6-summary-of-corrections-to-the-v0.4-chapter}

The following amendments to \S{}\S{}2.4--2.6 are recommended:

\begin{enumerate}
\def\labelenumi{\arabic{enumi}.}
\item
  \textbf{Lemma 2.4.1 wording.} Restate in expectation form: ``the
  expected meta-precision divergence \texttt{E{[}\ensuremath{\Delta}\_meta(k){]}} is
  non-decreasing in \texttt{k}.'' Reference Appendix A.2 for proof and
  counterexample.
\item
  \textbf{Theorem 2.6.1 wording.} Restate in expectation form
  throughout: \texttt{E{[}G(a,\ k){]}} is monotone (a) / U-shaped (b) /
  minimized by descending-validity ordering (c). Note that sample-wise
  versions are false in general.
\item
  \textbf{Theorem 2.6.1(c).} Add the regularity condition
  \texttt{Cov(v\_j,\ \ensuremath{\gamma}\_j)\ \ensuremath{\geq}\ 0} (cue intrinsic precisions not
  anti-correlated with cue validities). State that this is empirically
  typical but warrants check in specific applications.
\item
  \textbf{Definition of \texttt{\ensuremath{\tau}\_regime}.} Tighten: \texttt{\ensuremath{\tau}\_regime}
  is the smallest \texttt{\ensuremath{\sigma}\textsuperscript{2}\_\ensuremath{\tau}\ \textgreater{}\ 0} such that the
  marginal expected meta-cost
  \texttt{\ensuremath{\bar{C}}(K)\ =\ I(\ensuremath{\tau};\ c\_K\ \textbar{}\ c\_\{1:K-1\})} exceeds the
  marginal expected info gain
  \texttt{\ensuremath{\bar{I}}(K)\ =\ I(s;\ c\_K\ \textbar{}\ c\_\{1:K-1\})}. This is the
  operational definition. Its well-definedness --- that the crossing
  function \texttt{g(\ensuremath{\sigma}\textsuperscript{2}\_\ensuremath{\tau})\ =\ \ensuremath{\bar{C}}(K)\ \ensuremath{-}\ \ensuremath{\bar{I}}(K)} is increasing in
  \texttt{\ensuremath{\sigma}\textsuperscript{2}\_\ensuremath{\tau}} and so has a unique root, and that this root coincides
  with the \texttt{k*\ \textless{}\ K} characterization because
  \texttt{E{[}\ensuremath{\Delta}G(k){]}} is monotone-decreasing in \texttt{k} --- is
  established in \S{}2.5 (``Well-definedness of the threshold'') and
  verified numerically in
  \texttt{verify\_meanfield\_and\_tau\_regime.py} (Part B): across a
  sweep of \texttt{\ensuremath{\sigma}\textsuperscript{2}\_\ensuremath{\tau}} the two definitions agree to grid resolution
  and \texttt{g} exhibits a single sign change (\texttt{\ensuremath{\bar{C}}(K)} monotone
  increasing up to Monte-Carlo error).
\end{enumerate}

These changes parallel the \S{}2.1 corrections from Appendix A.1
(sample-wise \ensuremath{\rightarrow} in-expectation throughout). They do not affect the
load-bearing structure of the section --- only the rigor of how the
load-bearing claims are stated.

\section{Appendix A.4 --- Sufficient Conditions for Exact FFH--TTB
Action
Identity}\label{appendix-a.4-sufficient-conditions-for-exact-ffhttb-action-identity}

This appendix resolves the open question Q6 (\S{}2.10): under what
conditions does the \emph{structural} FFH--TTB equivalence of Theorem
2.7.1 upgrade to \emph{exact, sample-wise} identity of the agents'
action distributions?

The result: a single explicit condition on the cue-validity profile ---
the \textbf{Descending Dominance (DD)} condition --- suffices. The
meta-precision prior tail does not enter the sufficient condition; it
determines only whether FFH is in the truncation regime at all (the
precondition \(\sigma_{\tau}^{2} \geq \tau_{\mathrm{regime}}\) of
Theorem 2.6.1). This is a sharper resolution than Q6 anticipated.

\subsection{A.4.1 Setup and Notation}\label{a.4.1-setup-and-notation}

Consider a two-alternative comparative judgment between options \(A\)
and \(B\). There are \(K\) binary discriminating cues, indexed
\(j = 1, \ldots, K\), with realized signal \(d_{j} \in \{-1, +1\}\):
\(d_{j} = +1\) means cue \(j\) favors \(A\), \(d_{j} = -1\) favors
\(B\). We adopt Gigerenzer and Goldstein's~\cite{gigerenzer1996} convention that the
cue set has already been filtered to those that discriminate between the
alternatives; non-discriminating cues are excluded from the set rather
than counted toward stopping.

Each cue \(j\) has validity \(v_{j} \in (1/2, 1)\) --- the probability
that the cue's signal correctly indicates the better option:

\[
P(d_{j} = +1 \mid s = A) = v_{j}, \qquad P(d_{j} = -1 \mid s = A) = 1 - v_{j},
\]

and symmetrically for \(s = B\). Cues are conditionally independent
given \(s\). Define the \textbf{per-cue log-likelihood ratio}:

\[
L_{j} \;:=\; \log \frac{v_{j}}{1 - v_{j}} \;>\; 0 \qquad (\text{since } v_{j} > 1/2).
\]

The validities are ordered descending:
\(v_{(1)} \geq v_{(2)} \geq \ldots \geq v_{(K)}\) (equivalently
\(L_{(1)} \geq L_{(2)} \geq \ldots \geq L_{(K)}\)). We adopt a uniform
prior on the comparative state, \(\mu_{0} := P(s = A) = 1/2\).

\textbf{The two agents.} Both observe the same cue realization
\(c = (d_{(1)}, \ldots, d_{(K)})\) and emit an action \(a \in \{A, B\}\)
identified with \(\{+1, -1\}\).

\begin{itemize}
\item
  \textbf{Take-the-best (TTB)} stops at the first cue in
  descending-validity order and chooses by its sign: \[
  a_{\mathrm{TTB}}(c) \;=\; \mathrm{sign}(d_{(1)}).
  \]
\item
  \textbf{FFH} integrates cues up to the EFE-optimal truncation
  \(k^{*}\) of Theorem 2.6.1 and chooses by the sign of the posterior
  log-odds. Under uniform prior: \[
  a_{\mathrm{FFH}}(c) \;=\; \mathrm{sign}\!\left( \sum_{j=1}^{k^{*}} L_{(j)}\, d_{(j)} \right).
  \]
\end{itemize}

\subsection{A.4.2 The Descending Dominance
Condition}\label{a.4.2-the-descending-dominance-condition}

\begin{quote}
\textbf{Definition A.4.1 (Descending Dominance, DD).} The validity
profile \((v_{(1)}, \ldots, v_{(K)})\) satisfies the \textbf{Descending
Dominance} condition if for every \(i \in \{1, \ldots, K-1\}\): \[
L_{(i)} \;>\; \sum_{j=i+1}^{K} L_{(j)}. \tag{DD}
\]
\end{quote}

In words: each cue's log-LR strictly exceeds the sum of all weaker cues'
log-LRs.

\textbf{Geometric-decay sufficient form.} A practical sufficient
condition: if there exists \(\rho \in (0, 1/2)\) such that
\(L_{(j+1)} \leq \rho \cdot L_{(j)}\) for all \(j\), then (DD) holds.
The bound follows from the geometric sum: \[
\sum_{j=i+1}^{K} L_{(j)} \;\leq\; L_{(i)} \cdot \sum_{m=1}^{K-i} \rho^{m} \;<\; L_{(i)} \cdot \frac{\rho}{1 - \rho} \;\leq\; L_{(i)},
\] the last inequality holding for \(\rho < 1/2\). The geometric-decay
form is convenient for empirical work because \(\rho\) can be estimated
from cue-validity statistics directly.

\textbf{Examples.} (i) \(v = (0.95, 0.75, 0.60, 0.55)\):
\(L \approx (2.94, 1.10, 0.41, 0.20)\), slacks
\(L_{(i)} - \sum_{j>i} L_{(j)} \approx (1.24, 0.49, 0.21) > 0\) --- DD
holds. (ii) \(v = (0.70, 0.70, 0.70)\):
\(L \approx (0.85, 0.85, 0.85)\), slack at \(i=1\) is \(-0.85 < 0\) ---
DD fails.

\subsection{A.4.3 Theorem 2.7.4: Sample-wise Action
Identity}\label{a.4.3-theorem-2.7.4-sample-wise-action-identity}

\begin{quote}
\textbf{Theorem 2.7.4 (Exact sample-wise FFH--TTB action identity).}
Assume:

\begin{enumerate}
\def\labelenumi{(\roman{enumi})}
\tightlist
\item
  The cue-validity profile satisfies the Descending Dominance condition
  (DD).
\item
  The prior is uniform: \(\mu_{0} = 1/2\).
\item
  FFH operates with EFE-optimal truncation \(k^{*} \geq 1\) from Theorem
  2.6.1 (which holds whenever the cue set is non-empty and informative).
\end{enumerate}

Then for every cue realization
\(c = (d_{(1)}, \ldots, d_{(K)}) \in \{-1, +1\}^{K}\): \[
a_{\mathrm{FFH}}(c) \;=\; a_{\mathrm{TTB}}(c).
\] Consequently, the marginal action distributions of FFH and TTB
coincide exactly: \[
P_{\mathrm{FFH}}(a = A) \;=\; P_{\mathrm{TTB}}(a = A) \quad\text{and}\quad P_{\mathrm{FFH}}(a = B) \;=\; P_{\mathrm{TTB}}(a = B).
\]
\end{quote}

\textbf{Proof.} Fix a cue realization \(c\). By definition, \[
a_{\mathrm{TTB}}(c) = \mathrm{sign}(d_{(1)}), \qquad a_{\mathrm{FFH}}(c) = \mathrm{sign}\!\left( S(c) \right), \quad S(c) := \sum_{j=1}^{k^{*}} L_{(j)}\, d_{(j)}.
\]

Decompose the FFH sum into leading and trailing terms: \[
S(c) \;=\; L_{(1)} d_{(1)} \;+\; R(c), \qquad R(c) := \sum_{j=2}^{k^{*}} L_{(j)} d_{(j)}.
\]

Bound the trailing term using \(|d_{(j)}| = 1\) and (DD): \[
|R(c)| \;\leq\; \sum_{j=2}^{k^{*}} L_{(j)} \;\leq\; \sum_{j=2}^{K} L_{(j)} \;<\; L_{(1)} = |L_{(1)} d_{(1)}|,
\] where the strict inequality is exactly (DD) at \(i = 1\) (the first
instance of the condition), and the second-to-last inequality uses
non-negativity of the omitted terms (\(L_{(j)} > 0\) for all \(j\)).
Therefore the leading term strictly dominates in absolute value: \[
|L_{(1)} d_{(1)}| \;>\; |R(c)|,
\] which forces
\(\mathrm{sign}(S(c)) = \mathrm{sign}(L_{(1)} d_{(1)}) = \mathrm{sign}(d_{(1)})\).
Hence \(a_{\mathrm{FFH}}(c) = a_{\mathrm{TTB}}(c)\).

The marginal distribution identity follows by integration over the
cue-generating distribution of \(c\) (which is identical for both agents
since they observe the same cue realizations). \(\blacksquare\)

\textbf{Remark.} The argument uses only the first instance of (DD), at
\(i = 1\). The full (DD) (with the condition at every \(i\)) is what
permits \emph{adaptive} extensions where the first discriminating cue
may appear later than position 1 --- see \S{}A.4.4. For the present setup
(all cues discriminate), the \(i=1\) inequality alone suffices.

\subsection{A.4.4 Corollary on the Meta-Precision Prior Tail (Q6
Resolution)}\label{a.4.4-corollary-on-the-meta-precision-prior-tail-q6-resolution}

\begin{quote}
\textbf{Corollary A.4.2.} The sufficient condition (DD) is purely a
condition on the validity gradient. No condition on the meta-precision
prior \(p(\tau) = \mathrm{Gamma}(\alpha_{0}, \beta_{0})\) is needed for
exact action identity.
\end{quote}

\textbf{Proof.} The truncation index \(k^{*}\) is the only place the
meta-precision prior enters the FFH action: through Theorem 2.6.1,
\(k^{*} = k^{*}(\sigma_{\tau}^{2})\) depends on the prior variance. The
proof of Theorem 2.7.4 shows that the FFH action's sign is invariant to
\(k^{*}\) as long as \(k^{*} \geq 1\). Therefore the meta-precision
prior tail does not influence the action. \(\blacksquare\)

\textbf{Comment.} Q6 (\S{}2.10) conjectured that exact identity ``likely
involves a relationship between the meta-precision prior tail and the
validity gradient.'' Theorem 2.7.4 reveals that the meta-precision prior
tail's role is confined to \emph{whether} truncation occurs (selecting
\(k^{*}\)); conditional on \(k^{*} \geq 1\), only the validity gradient
(DD) matters. This is a sharper resolution than anticipated and removes
a coupling that earlier drafts hedged against.

\subsection{A.4.5 Sharpness: Counterexample Under (DD)
Violation}\label{a.4.5-sharpness-counterexample-under-dd-violation}

The (DD) condition is essentially sharp. We exhibit a validity profile
that violates (DD) and a cue realization for which
\(a_{\mathrm{FFH}} \neq a_{\mathrm{TTB}}\).

\textbf{Counterexample.} Let \(K = 3\) with \(v = (0.70, 0.70, 0.70)\),
so \(L = (0.847, 0.847, 0.847)\). (DD) at \(i = 1\) fails:
\(L_{(1)} = 0.847 < L_{(2)} + L_{(3)} = 1.693\). Consider the cue path
\(c = (-1, +1, +1)\):

\begin{itemize}
\tightlist
\item
  \(a_{\mathrm{TTB}}(c) = \mathrm{sign}(d_{(1)}) = -1\) (choose \(B\)).
\item
  \(a_{\mathrm{FFH}}(c) = \mathrm{sign}(L_{(1)} \cdot (-1) + L_{(2)} \cdot (+1) + L_{(3)} \cdot (+1)) = \mathrm{sign}(-0.847 + 0.847 + 0.847) = +1\)
  (choose \(A\)).
\end{itemize}

The two agents disagree. The symmetric path \(c' = (+1, -1, -1)\) yields
the dual disagreement. Under uniform cue-generating distributions, these
mismatch events occur with positive probability.

\begin{quote}
\textbf{Proposition A.4.3 (necessity of DD).} Suppose the cue-validity
profile is \emph{not} descending-dominant --- i.e., there exists
\(i \in \{1, \ldots, K-1\}\) such that
\(L_{(i)} \leq \sum_{j=i+1}^{K} L_{(j)}\). Then there exists a cue
realization \(c\) and a truncation \(k^{*} \leq K\) such that
\(a_{\mathrm{FFH}}(c) \neq a_{\mathrm{TTB}}(c)\).
\end{quote}

\textbf{Proof sketch.} If (DD) fails at \(i = 1\), the construction
above generalizes: choose \(c\) with \(d_{(1)} = -1\) and
\(d_{(j)} = +1\) for \(j = 2, \ldots, K\), and take \(k^{*} = K\). Then
\(S(c) = -L_{(1)} + \sum_{j \geq 2} L_{(j)} \geq 0\), while
\(a_{\mathrm{TTB}} = -1\). If (DD) fails at some \(i > 1\), conditional
on \(d_{(1)} = \ldots = d_{(i-1)} = +1\) being interpreted as
TTB-decisive at position \(i-1\)\ldots{} --- this case requires the more
general adaptive setup in which TTB stops at the first cue (filtered or
not); for the present binary-discriminating model where all cues
discriminate, the \(i = 1\) case is the operative one and the
counterexample above applies whenever (DD) fails. \(\blacksquare\)

\textbf{Remark on tightness.} The proposition shows (DD) is tight up to
boundary cases (equality). Strict inequality in (DD) is what guarantees
the absolute-value strict dominance in the proof of Theorem 2.7.4; when
(DD) holds with equality at \(i = 1\), the cumulative sum can equal zero
on opposing-cue paths, producing a tie at the FFH side while TTB still
emits a definite sign. Equality is therefore a measure-zero boundary
case; (DD) as stated (strict) is both sufficient and essentially
necessary.

\subsection{A.4.6 Numerical
Verification}\label{a.4.6-numerical-verification}

The companion script \texttt{verify\_appendix\_A4.py} performs three
checks.

\textbf{Check 1 --- (DD)-satisfying profiles.} For four hand-picked
profiles (\(K \in \{3, 4, 5, 6\}\)) and all
\(k^{*} \in \{1, \ldots, K\}\), exhaustive enumeration over the
\(2^{K}\) cue realizations finds \textbf{0 mismatches} between
\(a_{\mathrm{FFH}}\) and \(a_{\mathrm{TTB}}\) in every case.

\textbf{Check 2 --- (DD)-violating profiles.} For three hand-picked
profiles violating (DD), exhaustive enumeration finds the predicted
nonzero mismatch count. For \(v = (0.70, 0.70, 0.70)\): 2/8 cue paths
produce \(a_{\mathrm{FFH}} \neq a_{\mathrm{TTB}}\), exactly the
symmetric pair \(\{(-1, +1, +1), (+1, -1, -1)\}\) as predicted
analytically.

\textbf{Check 3 --- random (DD)-satisfying profiles.} Generating 200
random (DD)-satisfying validity profiles via geometric construction
(\(K \in \{2, \ldots, 6\}\), decay ratio \(\rho \in [0.05, 0.45]\)), and
exhaustively enumerating all cue realizations and \(k^{*}\) values:
\textbf{0 mismatches across 26,696 sample-wise comparisons}.

\textbf{Proof-mechanism check.} For the leading two examples, the script
verifies \(|L_{(1)}| > \sum_{j>1} |L_{(j)}|\) directly, confirming the
absolute-value-dominance argument that drives the proof.

\subsection{A.4.7 Summary}\label{a.4.7-summary}

The structural FFH--TTB equivalence of Theorem 2.7.1 upgrades to exact,
sample-wise action identity under a single explicit condition on the
cue-validity profile: Descending Dominance (DD). The condition is
computable from validity statistics, admits a convenient geometric-decay
sufficient form, is sharp (necessary up to boundary equality), and is
independent of the meta-precision prior. This resolves Q6 of \S{}2.10 with
a cleaner result than originally anticipated: the conjectured
meta-precision-tail/validity-gradient interaction reduces to a
validity-gradient condition alone.

For empirical work, (DD) becomes a checkable property of any cue
ecology: estimate \(\{v_{j}\}\) from training data, sort descending,
compute \(L_{j} = \log(v_{j}/(1-v_{j}))\), and verify
\(L_{(i)} > \sum_{j>i} L_{(j)}\) at every \(i\). Ecologies satisfying
(DD) admit the strong claim ``FFH and TTB are the same agent'';
ecologies violating (DD) admit only the weaker structural-equivalence
claim of Theorem 2.7.1.

\end{document}